\documentclass[sigconf,manuscript,nonacm]{acmart}

\AtBeginDocument{%
	}

\setcopyright{acmlicensed}
\copyrightyear{2018}
\acmYear{2018}
\acmDOI{XXXXXXX.XXXXXXX}

\acmJournal{JACM}
\acmVolume{37}
\acmNumber{4}
\acmArticle{111}
\acmMonth{8}

\usepackage{amsmath}
\usepackage{mathtools}
\usepackage{amsthm}

\usepackage{algorithm}
\usepackage{algorithmic}

\theoremstyle{plain}
\newtheorem{theorem}{Theorem}[section]

\theoremstyle{definition}

\newtheorem{assumption}[theorem]{Assumption}
\theoremstyle{remark}

\usepackage[colorinlistoftodos,bordercolor=orange,backgroundcolor=orange!20,linecolor=orange,textsize=scriptsize]{todonotes}

\DeclareMathOperator{\diag}{diag}       

\newcommand{\norm}[1]{\left\| #1 \right\|}

\newcommand{\mS}{\mathbf{S}}
\newcommand{\mI}{\mathbf{I}}

\newcommand{\mH}{\mathbf{H}}

\newcommand{\cC}{\mathcal{C}}

\newcommand{\bM}{\mathbf{M}}

\newcommand{\del}[1]{}

\newcommand{\eqdef}{\stackrel{\text{def}}{=}}

\newcommand{\R}{\mathbb{R}}

\newcommand{\Exp}[1]{\mathbb{E}\left[#1\right]}

\newcommand{\squeeze}{\textstyle}

\newcommand{\RD}{\mathbb{R}^d}

\definecolor{linenumbercolor}{rgb}{0.98, 0.81, 0.69}

\definecolor{classcolor}{RGB}{0,0,255}

\definecolor{apicolor}{RGB}{255,0,0}

\definecolor{linenumbercolor}{rgb}{0.1, 0.1, 0.1}

\definecolor{modelcolor}{RGB}{108,57,0}
\definecolor{datacolor}{RGB}{108,57,0}
\definecolor{abrcolor}{RGB}{108,57,0}
\definecolor{algcolor}{RGB}{108,57,0}
\definecolor{compcolor}{RGB}{108,57,0}
\definecolor{libcolor}{RGB}{108,57,0}

\newcommand{\dataname}[1]{{\color{datacolor}\sf \small {#1}}}   
\newcommand{\abr}[1]{#1}        

\newcommand{\algname}[1]{{\color{algcolor}\sf \small {#1}}}     
\newcommand{\compname}[1]{{\color{compcolor}\sf \small {#1}}}   

\newcommand{\HS}{L_{*}}      
\newcommand{\HF}{L_{\rm F}}  
\newcommand{\HM}{L_{\infty}} 

\usepackage{csquotes}
\usepackage{longtable}
\usepackage[flushleft]{threeparttable}
\usepackage{colortbl}
\usepackage{xcolor}

\usepackage{makecell}
\usepackage{nicefrac}
\usepackage{enumitem}
\definecolor{bgcolorwe}{rgb}{0.8,1,0.8}
\usepackage{microtype}
\usepackage{graphicx}
\usepackage{subfigure}

\definecolor{myblue}{rgb}{0.1, 0.3, 0.7}

\newcommand{\clr}[1]{#1}
\newcommand{\clrshort}[1]{#1}

\begin{document}
	
	\title{Unlocking FedNL: Self-Contained Compute-Optimized Implementation}
	
	\author{Konstantin Burlachenko}
	\email{konstantin.burlachenko@kaust.edu.sa}
	\orcid{0000-0001-5986-0855}
	\affiliation{%
		\institution{KAUST}
		\city{Thuwal}
		\country{KSA}
	}
	
	\author{Peter Richt\'{a}rik}
	\email{peter.richtarik@kaust.edu.sa}
	\orcid{0000-0003-4380-5848}
	\affiliation{%
		\institution{KAUST}
		\city{Thuwal}
		\country{KSA}
	}

	\renewcommand{\shortauthors}{Burlachenko et al.}
	
	\begin{abstract}
		\label{abs}
		
		Federated Learning (\abr{FL}) is an emerging paradigm that enables intelligent agents to collaboratively train Machine Learning (\abr{ML}) models in a distributed manner, eliminating the need for sharing their local data. The recent work \cite{safaryan2021fednl} introduces a family of Federated Newton Learn (\algname{FedNL}) algorithms, marking a significant step towards applying second-order methods to \abr{FL} and large-scale optimization. However, the reference \algname{FedNL} prototype exhibits three serious practical drawbacks: (i) It requires $4.8$ hours to launch a single experiment in a sever-grade workstation; (ii) The prototype only simulates multi-node setting; (iii) Prototype integration into resource-constrained applications is challenging. To bridge the gap between theory and practice, we present a self-contained implementation of \algname{FedNL}, \algname{FedNL-LS},  \algname{FedNL-PP} for single-node and multi-node settings. Our work resolves the aforementioned issues and reduces the wall clock time by $\times 1000$. With this \algname{FedNL} outperforms alternatives for training logistic regression in a single-node -- {CVXPY} \cite{diamond2016cvxpy}, and in a multi-node -- {Apache Spark} \cite{meng2016mllib}, {Ray/Scikit-Learn} \cite{moritz2018ray}. Finally, we propose two practical-orientated compressors for \algname{FedNL} - adaptive \compname{TopLEK} and cache-aware \compname{RandSeqK}, which fulfill the theory of \algname{FedNL}.
		
	\end{abstract}

	\ccsdesc[500]{Mathematics of computing~Mathematical software performance}
	\ccsdesc[500]{Mathematics of computing~Solvers}
	\ccsdesc[500]{Theory of computation~Convex optimization}
	\ccsdesc[300]{Theory of computation~Distributed algorithms}
	\ccsdesc[300]{Computing methodologies~Distributed simulation}
	\ccsdesc[300]{Computing methodologies~Machine learning algorithms}
	
	\keywords{federated learning, optimization software, large-scale optimization, second-order optimization, distributed optimization, high-performance implementation, C++}


\maketitle

\section{Introduction}
\label{intro}

Convex optimization finds applications in science and engineering and additionally serves as a tool for tackling non-tractable global optimization and combinatorial optimization problems. Examples can be found in  \citet{bertsekas2003convex}, \citet{boyd2004convex}. The second-order methods represent a category of continuous optimization techniques that go beyond utilizing gradient and function value information by incorporating details about the Hessians of the minimization objective. These methods offer advantages, such as invariance to affine changes in optimization variable coordinates and rapid local convergence. For instance, the pure Newton method exhibits a quadratic convergence rate near the solution. Existing ready-to-use systems within Mathematical Optimization and Machine Learning lack good support for large-scale distributed second-order optimization. \citet{wytock2016new} attribute this issue to the following factors: (a) Distributing optimization using Newton and Quasi-Newton methods demands high bandwidth to transmit Hessian-like quantities across the communication network; (b) The memory requirement for forming and storing second-order information is substantial; (c) Practical low-level linear algebra libraries, have limited interfaces. However, these concerns are highly correlated with questions under study in Federated Learning.

Federated Learning (\abr{FL}), introduced by \citet{konevcny2016federated}, is a specialized multidisciplinary subfield within \abr{ML}. It facilitates intelligent clients to collaboratively train \abr{ML} models in a distributed way, eliminating the need for centralized data collection. The 	\abr{FL} optimization algorithms have internal mechanisms to balance memory transfers and computation. Importantly in \abr{FL}, they are an integral part of the optimization algorithms. One of them is \textit{communication compression}. Examples of first-order optimization algorithms that was co-designed with {communication compression} includes  \algname{MARINA} \cite{gorbunov2021marina},\algname{COFIG/FRECON} \cite{zhao2021faster}, \algname{EF-21} \cite{richtarik2021ef21, richtarik2023error}. These algorithms and also recently started to appear algorithms for second-order optimization methods with mechanisms for \textit{communication compression} are still not spread enough, despite the robust theory. The first reason is that despite the rapid growth, \abr{FL} is still a highly specialized subfield within \abr{ML}. The second reason is that effectively applying methods in a multidisciplinary field demands proficiency in several disciplines. Ready-to-use implementations play a crucial role in simplifying this challenge.

The recent work by \citet{safaryan2021fednl} introduces the family of Federated Newton Learn (\algname{FedNL}) algorithms, marking a significant step in incorporating second-order methods into \abr{FL} and large-scale optimization. The optimization problem addressed by \algname{FedNL} \cite{safaryan2021fednl} has a finite sum structure:
\begin{equation}
	\label{eq:main}
	\squeeze \min \limits_{x\in \R^d} \left\{f(x)\eqdef \frac{1}{n}\sum \limits_{i=1}^n f_i(x) \right\}.
\end{equation}

In Eq.\eqref{eq:main}, functions $f(x), f_1(x),\dots, f_n(x) \in \mathbb{R}^d \to \mathbb{R}$ satisfies to the following assumptions:
\begin{assumption}\label{asm:1}
	The $f(x)$ is $\mu_f$ strongly convex. If $f(x)$ is twice continuously differentiable it is equivalent to condition for $\nabla^2 f(x)  - \mu I$ be positive semi-definite $\forall x \in \RD$.
\end{assumption}

\begin{assumption}\label{asm:2}
	There exist Lipschitz constants $\HS$, $\HF$, and $\HM$ such that $\forall i\in[n]$, $\forall x,y\in\R^d$:
	\begin{eqnarray*}
		\|\nabla^2 f_i(x) - \nabla^2 f_i(y)\|_2 &\leq & \HS \|x-y\|, \\
		\|\nabla^2 f_i(x) - \nabla^2 f_i(y)\|_{\rm F} & \leq & \HF \|x-y\|, \\
		\max_{j,l}| (\nabla^2 f_i(x) - \nabla^2 f_i(y))_{jl}| & \leq & \HM \|x-y\|.
	\end{eqnarray*}	
\end{assumption}

\clr{
	To the best of our knowledge, \algname{FedNL} is the state-of-the-art second-order optimization method in terms of \textit{theoretical} convergence guarantees in a class of algorithms striving to solve Eq.\ref{eq:main} under Assumptions \ref{asm:1}, \ref{asm:2}. We will not reiterate the comparisons of \algname{FedNL} with prior methods, as this has been comprehensively covered in Appendix A \cite{safaryan2021fednl}.
}

\algname{FedNL} supports communication compression for transferring Hessian information from clients. Its extensions include \algname{FedNL-PP}, designed for partial participation among clients, and \algname{FedNL-LS}, which ensures global convergence. The algorithm exhibits local superlinear convergence rates in terms of the squared Euclidean distance to the solution, independent of the Hessian's condition number. Practically, this is indistinguishable from the local quadratic rate achieved by the Newton method. Notably, executing \algname{FedNL} does not require knowledge of any problem-specific constants. \clr{These factors make the \algname{FedNL} algorithm family a promising choice for a range of practical applications in which Machine Learning problems fulfill Eq.\ref{eq:main} under Assumptions \ref{asm:1},\ref{asm:2}.}
	


\subsection{Contributions}	
\label{contributions}

Inspired by \citet{safaryan2021fednl}, we have explored the landscape of enhancements to make \algname{FedNL}  more practical. One line of possible research involves modifying optimization algorithms where step size computation explicitly avoids dependence on problem-specific constants. Notably, \algname{FedNL} already exhibits this property. Next, during our experimentation with a reference implementation of \algname{FedNL}, a significant challenge arose in launching numerical experiments. Launching \algname{FedNL} experiments using provided prototypes took $4.8$ hours for a single optimization process. Next, we noticed that the referenced prototype only simulates a distributed environment.  The \algname{FedNL} has a super-linear local convergence rate. With this level of theory development, the gain from further theoretical improvements might not be as substantial as those derived from a highly optimized implementation. These aspects motivated us to create a more well-developed \algname{FedNL} implementation on top of the original work. We acknowledge the considerable challenges of simultaneously developing both a comprehensive theoretical framework and a proficient implementation. Therefore in the present work, we focused our efforts on enhancing its \textit{practical implementation} and \textit{practical applicability} of \algname{FedNL} algorithm family. \clr{The practical implementation also serves other needs. First, the theoretical cost model may not be entirely accurate and needs to be adjusted (for examples of nuances in modern compute systems see Section~$  $\ref{sec:int-div}, Appendix~\ref{app:memory-hierachy-latencies}). Second, without real implementation, scientific methods often remain confined to theory. As S. Boyd noted in \cite{interviewboyd}, this is one reason why Control Theory has experienced limited adoption in the broad sense.
}


Summary of our contributions:

\begin{enumerate}
	\item \textbf{Compute Effectiveness.} We addressed the challenge of executing computationally intensive multi-node simulations for \algname{FedNL} on a single workstation, achieving a remarkable $\times 1000$ improvement in wall-clock time. 
	
	\item \textbf{Self-Contained Design.} We introduced a self-contained single-node and multi-node implementation of \algname{FedNL} (Algorithm \ref{alg:FedNL}), \algname{FedNL-PP} (Algorithm \ref{alg:FedNL-PP}), \algname{FedNL-LS} (Algorithm \ref{alg:FedNL-LS}). Our design facilitates seamless integration into resource-constrained systems and eliminates the need for library dependencies management. Our solution relies only on \abr{OS} interfaces. It is compatible with a number of OS, Compilers (Appendix \ref{app:supported-os-and-compilers}),
	CPUs (Appendix~\ref{app:cpus}). We provide native OS executable applications, dynamic libraries, and static libraries. In addition, we provide an easy way to generate extension modules for other programming languages (Appendix \ref{app:usability}).

	
	\item \textbf{Adaptive TopLEK Compressor.} We introduced an extension of the \compname{TopK} compression mechanism, termed \compname{Top-LEK}. The core idea is to perform compression for \compname{TopK} adaptively and to compress as much as theory allows, but not more (for details see Appendix~\ref{app:toplek}).
	
	\item \textbf{Cache-aware RandSeqK Compressor.} We proposed a cache-aware version of \compname{RandK} compressor, named as \compname{RandSeqK}. The theory of \compname{RandK} provides the need \textit{''null space''}, which we have exploited to make the algorithm \textit{cache-aware} (for details see Appendix~\ref{app:seqk}).
	
	\item \textbf{Logistic Regression with FedNL Outperforms  Best-Practice Solutions.} In a single-node setup, our implementation practically outperforms solvers encapsulated within CVXPY \cite{diamond2016cvxpy}, including the commercial MOSEK \cite{aps2022mosek} for solving logistic regression. In a multi-node setup, our implementation surpasses the performance of Apache Spark MLlib \cite{meng2016mllib} and Ray/Scikit-Learn \cite{moritz2018ray}. Our implementation has an initialization time smaller by $\times 25 - \times 50$, and exhibits faster solving by factor $\times 7 - \times 82$ in wall clock time (see Section~\ref{sec:experiments}).
	\item \clr{\textbf{First Robust Practical Implementation.} To the best of our knowledge, our implementation is the first \textit{robust}, \textit{practical} implementation for training (strongly) convex objectives Eq.\ref{eq:main} in FL settings (see Section~ \ref{sec:existing-fl-sota-system}).
	}
	
\end{enumerate}

Contributions \clrshort{(3)}-(6) are more focused toward \abr{FL}. The principles from contributions (1)-(\clrshort{2}) are valuable in scenarios when a theoretical  compelling\footnote{A robust theory is necessary. The good practical implementation, especially in C++, demands a significant amount of time.} ML algorithm requires a strong realization. We believe some of our findings and improvements raise interesting questions for designing training algorithms because the achievement of improvements on the order of $\times 1000$ (if the underlying compute and storage hardware is fixed) serves as an indicator of underlying fundamental issues. The broader impact of our work is elaborated in Appendix~\ref{app:discussions}.

\subsection{The Importance of Considering Practical Aspects to Avoid Catastrophic Outcomes}
\label{sec:extra-motivation}

First, we want to provide valuable insights from ML-related fields into why neglecting practical implementations can have disastrous consequences. In 1943, Kurt Lewin, a pioneer in social science, stated that \textit{"There is nothing so practical as a good theory" \footnote{The phrase is commonly linked to Kurt Lewin rather than his specific publication.}}, which sets one connection to how theory influences practice. It is well recognized that practical applications validate theories, and prompt theorists to refine their models. However, there are \textit{two less-known facts} why there is a need to balance theory and practice discovered in communities close to Machine Learning.

The first comes from the realm of Data-intensive Computing and Operating Systems. The constructive criticism by \citet{mcsherry2015scalability}, argues that an excessive focus on scalability in big-data systems hides an essential evaluation of absolute performance. The concern of the authors of this work was about whether (a) state-of-the-art distributed processing systems genuinely enhance overall performance or (b) the overheads introduced by them lead in fact to performance degradation. The authors demonstrated that in experiments with popular graph algorithms, unfortunately, the latter is true. This phenomenon has to be considered in the context of any multidisciplinary field.

The second is from the Computer Architecture. Before $2004$, there was no motivation to focus on focused performance optimization of algorithms. Up to this year, the transistor density experienced exponential growth \cite{10.1145/2133806.2133822}, and \textit{``just wait''} for new hardware was a valid strategy to improve wall clock time. From $2004$, the clock rate plateaued at $1.7-3.4$ GHz, and microprocessor vendors began incorporating multiple cores, cache hierarchies, and specialized units. After the paradigm shift, the compute architecture research  discovered that there are two ways to progress:

\paragraph{Design of domain-specific compute architectures} The key lies in adapting algorithms to the architecture's features. This process involves closely intertwining algorithms with the presented functional compute units and the intricate rules governing the utilization.

\paragraph{Refining software implementation} Modern scripting languages offer the advantage of democratizing algorithm implementations. But their eco-system clashes too much with the principles of the real hardware and systems (\clrshort{see arguments in Appendix~ \ref{app:nopython}}). \citet{leiserson2020there} underscore this fundamental conflict by demonstrating a staggering $\times 62\,806$ speedup for dense matrix multiply.

\section{Background on FedNL}
\label{preliminaries}

{
	\begin{algorithm}[H]
		\caption{\algname{FedNL} ({\color{blue}Baseline}. Federated Newton Learn from \citet{safaryan2021fednl}) \clrshort{\textbf{[EXISTENT]}}}
		\label{alg:FedNL}
		\begin{algorithmic}[1]
			\STATE \textbf{Parameters:} Hessian learning rate $\alpha\ge0$; compression operators $\{\cC_1^k, \dots,\cC_n^k\}$
			\STATE \textbf{Initialization:} $x^0\in\R^d$; $\mH_1^0, \dots, \mH_n^0 \in \R^{d\times d}$ and $\mH^0 \eqdef \frac{1}{n}\sum_{i=1}^n \mH_i^0$
			\FOR{each device $i = 1, \dots, n$ in parallel} 
			\STATE Get $x^k$ from the server and compute local gradient $\nabla f_i(x^k)$ and local Hessian $\nabla^2 f_i(x^k)$
			\STATE Send $\nabla f_i(x^k)$,\; $\mS_i^k \eqdef \cC_i^k(\nabla^2 f_i(x^k) - \mH_i^k)$ and $l_i^k \eqdef \|\mH_i^k - \nabla^2 f_i(x^k)\|_{\rm F}$ to the server
			\STATE Update local Hessian shift to $\mH_i^{k+1} = \mH_i^k + \alpha\mS_i^k$
			\ENDFOR
			\STATE On Server:
			\STATE \quad Get $\nabla f_i(x^k),\; \mS_i^k$ and $l_i^k$ from each node $i\in [n]$
			\STATE \quad $\mS^k = \frac{1}{n}\sum\limits_{i=1}^n \mS_i^k,\; l^k = \frac{1}{n}\sum\limits_{i=1}^n l_i^k,\; \mH^{k+1} = \mH^k + \alpha\mS^k$
			\STATE \quad {\textbf{Option 1 (a)}:} $x^{k+1} = x^k - [\mH^{k}]_{\mu}^{-1} \nabla f(x^k)$ \quad {\textbf{Option 2 (b)}:} $x^{k+1} = x^k - [\mH^{k} + l^k\mI]^{-1} \nabla f(x^k)$
		\end{algorithmic}
	\end{algorithm}
}

The optimization problem which \algname{FedNL} solves has a finite sum structure described by Eq.~\eqref{eq:main}. Here, $n \in \mathbb{N}$ is the number of clients in a distributed system, $d \in \mathbb{N}$ denotes the dimension of the problem, and $x^k \in \mathbb{R}^d$ represents the model parameters at iteration $k$. When using \algname{FedNL} for \abr{FL} problems in \abr{ML}, $f_i(x)$ provides the score criteria for using the model $x$ on client $i$ data. The optimization formulation in Eq.~\eqref{eq:main} encodes the final goal of the training process as a selection of a function from a parameterized function class $\mathcal{F}$ indexed by $x \in \mathbb{R}^d$. In this context, $f_i(x)$ often takes the form $f_i(x) \eqdef \frac{w_i}{n_{i}} \sum_{1 \le j \le n_i} (\mathcal{L}_{ij}(b_{ij}, \hat{F}(a_{ij};x)) + R_{i}(x)).$ Here $f_i(x)$ assesses a predictive model $\hat{F}(\cdot;x) \in \mathcal{X}\to \mathcal{Y}$ common to all $n$ clients, $n_i \in \mathbb{N}$ denotes the number of input-output pairs at client $i \in [n]$, and $(a_{ij}, b_{ij}) \in \mathcal{A} \times \mathcal{B}$ represent the single input-output pair with the number $j$ at client $i$. The function $\mathcal{L}_{ij}(y_{\mathrm{real}}, y_{\mathrm{pred}} ): \mathcal{B} \times \mathcal{B} \to \mathbb{R}$ is a loss function that scores prediction, and $R_{i}(x): \mathbb{R}^d \to \mathbb{R}$ is the regularization function. The weight $w_i \ge 0$ encodes the role of client $i$.

The convergence guarantees of \algname{FedNL} hold if $f(x)$ satisfies Assumption \ref{asm:1} and $f_i(x)$ satisfies Assumption \ref{asm:2}. While \algname{FedNL} does not require strong convexity of functions $f_i$, it is required for $f(x)$. In \algname{FedNL} (Algorithm~\ref{alg:FedNL} from Section 3.4  \citet{safaryan2021fednl}), the only quantity not evaluated in runtime is $\alpha \in \mathbb{R}$. It is derived from the global characteristics of the used compressor.

\clr{

To integrate user-defined optimization problems, users must explicitly define oracles $\nabla^2 f_i(x)$, $\nabla f_i(x)$, and $f_i(x)$. To facilitate this process, we provide a comprehensive collection of mathematical primitives for computations on both CPU and NVIDIA GPUs, as well as system primitives for efficient input/output management and memory utilization. Also, we offer tools for numerically verifying the correctness of the $\nabla^2 f_i(x)$ and $\nabla f_i(x)$ oracles (see Appendix~ \ref{app:usability}, \ref{app:futresearch}). Our implementation follows a modular design. Due to the principles of the selected language (C++ ISO/IEC 14882:2020), modern compilation tools, and static linkage, there is no performance penalty from the modularity. 
}

\section{Compute and Storage Demands of a Single Worker}
\label{problem_single_node}

	Algorithm \ref{alg:FedNL} encapsulates the essential procedures employed when multiple \textit{clients} collaboratively solve the optimization problem Eq.\eqref{eq:main}. The algorithm's versatility is independent of specific computing hardware, but in realistic scenarios computing device should be fixed. In our implementation, we target modern general-purpose central processing units. 
	
	\clr{This type of computing device is ubiquitous, spanning from server-grade machines to portable devices. Even when electronic components are integrated into a single System-on-Chip, the Central Processing Unit (CPU) remains a fundamental element. The architectural details at the assembly language level, have evolved sustainably over the decades, facilitating effective decoupling between Electrical Engineering and Computer Science and Engineering. In contrast, the landscape for Graphics Processing Units (GPUs) is the opposite. The Application Programming Interfaces\footnote{OpenCL for ARM Mali GPU, CUDA for NVIDIA GPU, Metal API for Apple GPU, and ROCm for AMD GPU}, memory management rules, and computational organization principles evolve rapidly, even within a single vendor.
	}
	
	\clrshort{To evaluate our implementation of the \algname{FedNL}} a specific class of optimization problems must be chosen and we chose $L_2$ regularized logistic regression. This selection is guided by the existence of experiments with this objective in the original \abr{FedNL} paper, and the fact that it ensures strong convexity. 
	
	The logistic regression can be obtained from Eq. \eqref{eq:main} through the following specialization:

	\begin{equation}
		\label{eq:fi_log_reg_structure}
		f_i(x) = \dfrac{1}{n_i} \sum_{j=1}^{n_i} \log (1 + \exp(-b_{ij} \cdot a_{ij}^\top x)) + \dfrac{\lambda}{2} \|x\|_2^2
	\end{equation}
	From Eq.\ref{eq:fi_log_reg_structure} we can analytically compute $\nabla f_i(x)$, $\nabla^2 f_i(x)$ by:
	
	\begin{eqnarray}
		\label{eq:grad_fi_logreg_structure}
		\nabla f_i(x) = A \times \begin{bmatrix} \dfrac{-1/n_i}{\exp (x^\top (b_{i,1} a_{i,1})) + 1} \\ \dots \\ \dfrac{-1/n_i}{\exp (x^\top (b_{i,n_i} a_{i,n_i})) + 1} \end{bmatrix} + \lambda \cdot x,
	\end{eqnarray}
	
	\begin{eqnarray}
		\label{eq:hessian_fi_logreg_structure}
		\nabla^2 f_i(x) = A_i \times \diag(h_1,\dots, h_{n_i}) \times A^{\top} + \lambda \cdot I_{d \times d}.
	\end{eqnarray}
	Where $I_{d,d} \in \mathbb{R}^{d \times d}$ is the identity matrix, and
	\begin{eqnarray}
		\label{eq:hessian_structure_h_k}
		h_k &=& \dfrac{1}{n_i} \dfrac{\exp(x^\top \cdot b_{ij} a_{ij})}{ \left( 1 + \exp(x^\top \cdot b_{ij} a_{ij}) \right)^2 }, \\
		A_i &\eqdef& \begin{bmatrix}	b_{i,1} \cdot a_{i,1}  b_{i,2}\cdot a_{i,1}  \dots b_{i,n_i} \cdot a_{i,n_i} \end{bmatrix}_{d \times n_i} \notag.
	\end{eqnarray}
	
	These analytical formulas have been verified numerically. Regarding memory, our implementation allocates virtual memory in the executing processes without utilizing specific mechanisms in the operating system to force dedicated physical memory (via \textit{page-locking}, also known as \textit{pinning} mechanisms). 
	
	The virtual memory is used to store $x \in \mathbb{R}^d$, $\nabla f_i(x) \in \mathbb{R}^d$, $\nabla^2 f_i(x) \in \mathbb{R}^{d \times d}$, design matrix $A_i \in \mathbb{R}^{d \times n_i}$ and other intermediate buffers. The selected dataset in LIBSVM \cite{chang2011libsvm} format is read from disk storage twice, following a schema similar to Scikit-Learn \cite{pedregosa2011scikit} and Apache Spark MLlib \cite{meng2016mllib}.
	
	\section{Compute Issues in Original Implementation}
	\label{sec:issues}
	
	The reference implementation of \algname{FedNL} was developed in Python \cite{van1995python} with NumPy \cite{van2011numpy} employed as the computational backbone. A single execution of the provided implementation for logistic regression utilized parameters $d=301$, $n=142$, derived by splitting the LIBSVM \dataname{W8A} dataset into chunks $n_i=348$, takes $19,770$ seconds for \compname{TopK[$K=8d$]} and $17,510$ seconds for \compname{RandK[$K=8d$]}, with $r=1000$ rounds of repeating Lines 3-11 of Algorithm~\ref{alg:FedNL}. This measurements were taken on a machine equipped with an \textit{Intel(R) Xeon(R) Gold 6246 CPU} \footnote{\href{https://ark.intel.com/content/www/us/en/ark/products/193969/intel-xeon-gold-6246-processor-24-75m-cache-3-30-ghz.html}{Intel Xeon Gold 6246 Processor Technical Specification}.} with 12 physical computation cores. The CPU clock frequency was set to $\mu=3.3$ GHz (For preparation details see Appendix~\ref{app:careful-reproduce}).

	\paragraph{Back-of-the-Envelope calculation} 
	To estimate the lower bound on the execution time of the simulation for a fixed algorithm, we need to examine the internal organization of the \textit{{Intel(R) Xeon(R) Gold 6246 CPU}}. This organization is determined by the micro-architecture of \textit{{Cascade Lake}} processors, which encompasses information about the components inside the \abr{CPU}. The adders and multipliers are electrical circuits that perform corresponding operations at the rising edge of the clock that synchronize the operation of inside \abr{CPU}. In the \textit{Cascade Lake} micro-architecture, the number of Float Functional Units $\rm{FPU}=3$ \footnote{\href{https://en.wikichip.org/wiki/intel/microarchitectures/cascade_lake}{Reference information about Cascade Lake Microarchitecture}. Microarchitecture details are typically under NDA.}. These functional units, responsible for float add, subtract, and multiply have a throughput of $1$ operation per clock cycle in modern \abr{CPU} (See  \citet{fog_instruction_tables}). Each \abr{FPU} is a pipelined device. If the pipeline needs to be restarted, additional $4$ clocks are incurred per operation. The computation demands of Algorithm \ref{alg:FedNL} require each of $n$ client compute hessian with  $\mathcal{O}(d^2\cdot n_i)$ arithmetic operations, full gradient with $\mathcal{O}(d\cdot n_i)$ arithmetic operations, function values $\mathcal{O}(d\cdot n_i)$ per round. Hessian compression takes $\mathcal{O}(d^2)$, update the hessian shift $\mathcal{O}(d^2)$ arithmetic operations. Clients' compute logic happened during $r$ rounds requires: $$\mathcal{O}\left( (d^2\cdot n_i + d n_i + 2d^2) \cdot r\right)/(\mu \cdot \rm{cores} \cdot \rm{fpu}) \propto 0.26\,\rm{sec.}$$
	
	The master has to perform $n$ additions of hessian with $d \cdot k$ elements, and $n$ additions of gradients with dimension $d$. The master's compute time for processing:
	$$\mathcal{O}\left( (d\cdot k + d) \cdot r \cdot n\right)/(\mu \cdot \rm{cores} \cdot \rm{fpu}) \propto 0.0032 \,\rm{sec.}$$
	
	The master also needs to solve a linear system in each round. In the reference implementation, it was done using Gaussian elimination, which demands $(2/3)d^3$ arithmetic operations. Time for it:
	$$\mathcal{O} (
	{3}/{2} \cdot d^3 \cdot r 
	) / (\mu \cdot \rm{fpu}) \propto 4.1316 \,\rm{sec.}$$
	
	With optimal CPU core utilization, the time for \textit{float arithmetic} should be a small multiple of $4.394$ sec. However, \textit{float operations} is also required to get the actual operands. For doing this the \abr{CPU} control unit sends requests to the Load and Store Units (LS), which is also in charge of memory access to \abr{CPU} registers and caches. The number of LS units in \textit{{Cascade Lake}} is $3$. If assume that LS units work mostly with \texttt{L1} cache then extra penalty $\times 3$ (See Table~\ref{tbl:latencies-for-memory}) should be paid per single access. If assume that each float-operations requires $3$ memory accesses, then the memory access penalty for \textit{float operations} will be $(4.394 \cdot \rm{fpu}) / \rm{ls} \cdot 3 = 13.182$ sec.
	
	\paragraph{Insights from rough estimates} Rough lower bound of the estimated target execution time is $17.576$ sec. This estimate does not include memory cost for transfers in client-master communication,  controlling logic, and \textit{possible improvements} from utilizing Single Instruction Multiple Data (SIMD) instructions available in this \abr{CPU}. The observed time for launching the \algname{FedNL} baseline is $19770\,\rm{sec.}$ ($5.5$ hours), indicating a notable discrepancy:
	
	$$5.5\, \rm{hours} \ggg 17.576\,\rm{sec.}$$ 
	
	We addressed the challenges related to creating a practical single-node simulation of \algname{FedNL}, ensuring its compatibility with diverse \abr{OS}/\abr{CPU} configurations (see Appendix~\ref{app:supported-os-and-compilers}). After providing an optimized single-node simulation we created a practical multi-node implementation.
	
	\subsection{Computation Problems of Reference Implementation}
	
	After examining Python/NumPy reference implementations, we found a deeper issue. Major ML frameworks, burdened by extensive auxiliary management code, are suboptimal for complex system and algorithm creation with high-performance requirements. To tackle this, we've shifted away from general-purpose ML and FL middleware and the prevalent Python-centric design philosophy. \clrshort{For additional discussions on the computational problems see Appendix~\ref{app:nopython}}.
		
	\section{Structure of x1000 Time Improvement}
	
	We present more notable improvements from the \algname{FedNL} simulation in one machine, focusing on the wall clock time improvement for training logistic regression Eq.\eqref{eq:main}, \eqref{eq:fi_log_reg_structure} with \dataname{LIBSVM} \dataname{W8A} dataset. For finer granularity of improvements see Appendix~\ref{app:history-of-improvements}. We augmented each sample in a dataset with an artificial feature equal to $1$ to have an intercept term. After augmentation, \dataname{W8A} dataset contains $d=301$ features. Number of rounds $r=1000$. The dataset is reshuffled u.a.r and was split across $n=142$ clients with $n_i=350$. The $x^0=0$ and regularization coefficient $\lambda=0.001$, for this problem $\lambda\left(\nabla^2 f\right) \in [0.001, 0.0058]$. The overall time encompasses: (1) loading and parsing the dataset; (2) distributing the dataset and preparing runtime; (3) saving the outcome of the experiment to the disk; (4) training with Algorithm \ref{alg:FedNL}. The (1)-(3) takes $4.7\%$, and (4) $95.3\%$ of the  simulation time.
	
	\subsection{Naive C++ Implementation: x20}
	\label{sec:depart-from-python}
	
	Targeting the \abr{CPU} requires the \algname{FedNL} ecosystem to be implemented in a formal programming language. Defining a \textit{programming language} lacks universal consensus. The definition advocated by authors of compiler-based languages suggested that a language must directly translate logic into computation devices. Scripting languages lack this feature due to the requirement of an interpreter. Studies \cite{pereira2017energy}, \cite{leiserson2020there} underscore the superior speed and energy efficiency of compiled languages. Transitioning from Python/NumPy to C++ \cite{cpp} yielded a $\times 22$ time improvement for \algname{FedNL/TopK} and $\times 16$ for \algname{FedNL/RandK}. Each client is represented as a user-space thread in the simulation.
	

	\subsection{Data Processing Optimization: x1.077}
	
	We optimized the input data handling in \dataname{LIBSVM} format by moving from sequential I/O to memory-mapped files. This step, coupled with custom string to FP64 parsing, leads to a speedup $\times 1.077$. \clrshort{For more details on memory-mapped files ee Appendix~\ref{app:mmap}.}
	
	\subsection{Eliminating Some Integer Division: x1.225}
	\label{sec:int-div}
	
	Combination circuits handling integer and floating-point operations exhibit latency from $1$ to $4$ clocks. The division poses a challenge due to microarchitectural complexities, often requiring approximately $56$ clocks \cite{fog_instruction_tables}. In our dense linear algebra implementation, we optimized indexing operations by eliminating a division during indexing. These optimizations yielded a $\times 1.2$ gain. In our dense matrix/vector implementation, we leverage SIMD CPU instructions ({AVX-512}). To chunk dense arrays into 512 bits packs and the rest we utilized bits tricks. Our implementation requires $7$ x86\_64 instructions, and GCC 11.4 emits $9$ (from C++ code). It gave $\times 1.0212$ speedup.
	
	\subsection{Utilizing AVX512 CPU Extension: x1.379}
	
	
	Data alignment and utilization of the AVX-512 instruction set resulted in an additional gain $\times 1.379$.
	
	\subsection{Compiler and Linker Optimization: x1.128}
	
	By explicitly disabling exceptions and runtime type information in C++ compilers, we achieved a gain of $\times 1.07$. Leveraging whole program optimization, where code-emitting decisions are deferred to the final stages, adds an extra gain of $\times 1.047$. Employing force inlining at specific locations results in an improvement of $\times 1.007$. The combined effect of compile-time optimization is $\times 1.128$.
	
	\subsection{Use Sparsity from FedNL Compressors: x1.44}
	
	The \algname{FedNL} Lines 5,6,10 contain sparse updates. Exploiting this for the dense matrices has a drawback in that SIMD cannot be used, but the gain from not performing useless arithmetics outweighs it.

	\subsection{Reuse Computation from Oracles: x1.50}
	
	
	There exists a significant redundancy in the structure of $f_i(x)$, $\nabla f_i(x)$, $\nabla^2 f_i(x)$ as presented in Eq.\eqref{eq:fi_log_reg_structure}, \eqref{eq:grad_fi_logreg_structure}, \eqref{eq:hessian_fi_logreg_structure}, \eqref{eq:hessian_structure_h_k}. The classification margins $b_{ij} \langle x^\top, a_{ij} \rangle$ are reused within a single client three times. The computational cost in clock cycles is approximately $\mathcal{O}(d)$ for elementary add/multiply operations. Not only is the classification margin reused in all oracles but also the sigmoid function values $g(z)=(1 + \exp(-z))^{-1}$ and its derivatives $g(-z)=1-g(z)$ and $g(-z) \cdot g(z)=\exp(z)/(1+\exp(z))^2$ are implicitly reused in all three oracles. We eliminated this by reusing common values, incurring the cost of storing two vectors of dimension $\mathcal{O}(n_i)$ for classification margins and sigmoid values.
	
	\subsection{Basic Linear Algebra Improvements: x1.338}
	
	A careful implementation of adding the same scalar to the diagonal in Line 11 resulted in a gain of $\times 1.06$. Exploiting the symmetry of the input matrix when computing the Frobenius norm in Line 5 provided a gain of $\times 1.00751$. Explicitly storing information about the number of columns in a dense matrix yielded a $\times 1.0421$ gain. Implementing an efficient memory \textit{move} and \textit{copy} for matrix gave $\times 1.019$. Improving the initialization of matrices gave $\times 1.057$. To enhance instruction-level parallelism, the manual loop unrolling for vector-vector and vector-scalar operations gave $1.034$. Eliminate the aliasing effect problem in C++ code gain $\times 1.08$. Cumulative gain is $\times 1.3386$.

	\subsection{Linear System Solve Improvements: x1.31}
	
	We transitioned from dense Gaussian elimination to the more numerically stable Cholesky-Banachiewicz decomposition for dense matrices \cite{golub2013matrix}. We didn't explore the connections between \abr{FedNL} and \textit{indirect solvers} like Krylov methods  \cite{kelley1999iterative} and Multi-grid methods \cite{hackbusch2013multi}.\textit{Dense direct} methods are nonheuristic, independent of problem instances.  The switch to Cholesky, with optimized forward-backward substitution, yielded a $\times 1.196$ gain. Further enhancements, with in-place vector arithmetic and cache-aware computation organization, contributed to an addition  $\times 1.096$ gain.

	\subsection{Hessian and Gradients Oracles: x3.072}
	\label{sec:better-hessians-for-logreg}
	
	The $\nabla^2 f_i(x)$ computation used a naive implementation employing straightforward matrix multiplication with $3$ nested loops. We progressed to cache-aware matrix tiled multiplication, incorporating $9$ nested loops. The optimal tile size can be estimated based on the CPU's L1 and L2 data cache sizes, which are $32$ and $1024$ KBytes in our \abr{CPU}. The optimal tiles size to operate on three matrices (two inputs and output) are $\sqrt{L1/8}/3$, $ \sqrt{L2/8}/3$ \cite{jia1981complexity} for FP64 matrix items. Theory recommends tile sizes of $21$ and $120$, but practical experimentation favored the smaller sizes as $4$ and $32$. The gap arises from the difference between cache-aware and cache-oblivious (\abr{CO}) algorithms. The \abr{CO} algorithms adapt to the multilevel structure of caches, available cache size in a specific moment implicitly \cite{demaine2002cache} and do not require explicit knowledge of caches in advance. We tried standard \abr{CO} schema \footnote{\href{https://math.mit.edu/~stevenj/18.335/oblivious-matmul-handout.pdf}{Cache-Oblivious Matrix Multiplication} by \href{https://math.mit.edu/\~stevenj/}{Steven G. Johnson}: \href{https://math.mit.edu/\~stevenj/18.335/oblivious-matmul-handout.pdf}{https://math.mit.edu/\~{}stevenj/18.335/oblivious-matmul-handout.pdf}} for matrix multiplication with 8-way splitting which is to best of our knowledge is standard schema for \abr{CO} matrix multiplication to date. It resulted in a $\times 3$ speedup compared to the tuned tiled version. The final possible gain is $\times 1.2$.

	\paragraph{Better strategy} We can compute $\nabla^2 f_i(x^k)$ from Eq. \eqref{eq:hessian_fi_logreg_structure} as the sum of symmetric Rank-1 matrices. We can focus on the upper diagonal part and symmetrize the result matrix afterward (gain $\times 1.85$ overall). We did switch to processing $4$ samples with instruction-level parallelism inside the hessian oracle (gain $\times 1.6303$). We fused operations for matrix-vector operations and added multiples of vectors required from Eq.~\eqref{eq:grad_fi_logreg_structure} (gain $\times 1.0203$). The cumulative gain is $\times 3.077$.
	
	\subsection{Better Compressors Implementation: x1.14}
	
	Optimized \abr{SIMD} generation of integer sequences $(s, s+1, \dots)$, coupled with a known at compile-time $s$ value, yielded a $\times 1.006$ gain because this operation used in data shuffling and in \compname{RandK}. Next, we computed and stored indices for the upper triangular part of $\nabla^2 f_i$ once without recomputing (gain $\times 1.0165$). 
	
	For sparsification compressors, spending an additional $\mathcal{O}(k\log_2(k))$ time per client on sorting by index improved memory access in master, resulting in a $\times 1.0182$ gain.
	
	In implementing \compname{TopK}, diverse approaches were explored, including quicksort, merge sort, radix sort, randomized order statistics, \abr{CO} Multi-way merge sort \cite{cormen2022introduction}, \abr{CO} Funnelsort \cite{frigo1999cache}. The most efficient implementation involved a 4-way Min-Heap \cite{tarjan1983data} for supporting $K$ smallest items found so far, resulting in a $\times 1.0412$ gain. In the \compname{RandK} compressor, adopting in-place memory shuffling (instead of using two separate arrays of size $\mathcal{O}(d^2)$) gained $\times 1.073$. Finally, during waiting for $x^{k+1}$ (Line 4, Algorithm~\ref{alg:FedNL}), clients start precomputation of indices for the next round (gain is $\times 1.0211$).
	
	\subsection{Subtleties with Multi-Threading: \clrshort{x}1.412}
	
	To minimize client wait times we utilized busy loops on atomic variables instead of semaphore \abr{OS} primitives, resulting in a gain of $\times 1.0057$. To minimize system contention we established a thread pool with a size matching the number of physical cores. Clients were statically dispatched to this pool, avoiding unnecessary congestion for compute cores. Next, we processed messages from clients as they became available. 
	
	The last two techniques result in a combined gain of $\times 1.404$. This is a common technique in parallel and concurrent programming. These optimizations aim to minimize contention and enhance overall system efficiency.

	\subsection{Subtle Memory Optimization: \clrshort{x}1.278}
	
	In C++ runtime, threads allocate memory from a global heap with \abr{OS} locking mechanism. To enhance efficiency, we introduced thread-custom memory pools for vectors and matrices, resulting in a gain of $\times 1.0109$. Additionally, using aligned load/store for dense vector/matrix operations in $\nabla^2 f_i(x)$, $\nabla f(x)$ oracles provided a gain of $\times 1.11$. The client-master communication was organized via a memory buffer. Next, we observed that labels $b_{ij}$ is not needed in Eqs. \eqref{eq:fi_log_reg_structure}, \eqref{eq:grad_fi_logreg_structure}, \eqref{eq:hessian_fi_logreg_structure} explicitly and can obserbed into $A_i$ explicitly. It gave a gain of $\times 1.009$. Next, in the naive implementation, clients stored both $A_i$ and ${A_i}^\top$. By eliminating the storage of $A_i$ and adding the functionality for matrix operation with transposed argument, we gained $\times 1.129$. Cumulative gain is $\times 1.278$.
	
	\section{Conclusion About Single-Node Simulation}
	
	
	
	
	The total speedup gained from the previous steps is $\times 1000$. Extra improvements are more technical and obtained with profiling tools such as Valgrind \cite{nethercote2007valgrind}, and LLVM remarks \cite{lattner2004llvm} (See Appendix~	\ref{app:history-of-improvements}).
	
	\section{Network Improvements: x1.29}
	
	\paragraph{Baseline} We focused on a setting where communication happens via \abr{TCP/IP} for the reasons in Appendix~\ref{app:networks-why-tcp}. We took the setting $n=10$, \compname{TopK[k=8d]}, $r=300$, \dataname{W8A} dataset. Each client$\to$master stream of information was organized as a separate \abr{TCP} connection. The baseline took $14.92$ seconds. 
	
	\paragraph{Improvements} We observed that it's more effective to have a single communication channel from client to master. Next, we interleaved connection establishment with the master and dataset loading in the clients. For \compname{RandK} and \compname{RandSeqK} compressors, we enable index reconstruction using a pseudo-random generator. For \compname{TopK} and \compname{TopLEK}, it is necessary to transfer indices. We found that employing a fixed-width $32$-bit integer format surpassed the performance of a strategy involving varying sizes. Next, we intentionally disabled the Nagle algorithm \cite{nagle1984congestion} due to the explicit formulation of small buffers, as detailed in Appendix~\ref{app:networks-background}.
	
	\clearpage
	\section{Introduced Compressors}
	
	In our pursuit of bridging the gap between \textit{theory} and \textit{practice} we asked ourselves:
	\begin{center}
		\textit{"Is any nullspace of the introduced compressors that can be exploited for practical purposes?"}
	\end{center}
	
	We observed that the \compname{TopK} compressor employed in the original paper through contractive definition \citet[p.11]{safaryan2021fednl}. This absence of an unbiasedness requirement allows for the construction of "any algorithm" satisfying this property. We introduced \compname{TopLEK[k']} described in Appendix~\ref{app:toplek}. It adapts the \compname{TopK[$k$]} compressor, transforming it into a search process for \compname{TopK[$k' \le k$]}. The selection of $k' \in [0, k]$ ensures that contractive inequality becomes a tight equality.
	
	Next, the \compname{RandK} compressor exhibits an intriguing degree of freedom. While \compname{RandK} selects a subset of coordinates of cardinality $k$ u.a.r. from a total of $d$ coordinates, zeroing out the rest and scaling the output to preserve unbiasedness, an alternative realization of item selection is possible. In \compname{RandSeqK} schema, the group of coordinates is sequentially chosen from the start index $s\sim_{u.a.r.} [d]$, but next $k-1$ are sequential deterministic indices $\mathbb{Z}_d$. This \compname{RandSeqK} schema is unbiased and satisfies \citet[p.11]{safaryan2021fednl} and has the same variance as \compname{RandK}. But it is more appealing for practice because (a) it is cache-aware; (b) the number of invocations of the used pseudo-random generator is $1$. For a more elaborate presentation of \compname{RandSeqK} see Appendix~\ref{app:seqk}.

	\section{Experiments}
	\label{sec:experiments}
	
	We trained $L_2$ regularized logistic regression as defined in Eq.\eqref{eq:fi_log_reg_structure} with $\lambda=0.001$. The initial iterate $x^0=0$. When utilizing \compname{RandK} and \compname{RandSeqK}, we leveraged our implementation's ability to (optionally) reconstruct indices of sparsified information. For \compname{TopK} and \compname{TopLEK}, the indices are transferred as 32-bit integers. See Appendix~\ref{app:software-env},~\ref{app:hardware-env} for details on the software and hardware environment. Employed steps for reliable time measurements are described in Appendix~\ref{app:careful-reproduce}. We augmented implementation with \compname{Natural Compressor} \cite{horvoth2022natural}, which behaves remarkably well for \algname{FedNL}. The selected datasets are example datasets used in the original FedNL paper by \citet{safaryan2021fednl}.

	

	\subsection{Single-node: Baseline Vanilla FedNL Improvement}
	\label{sec:single-node-fednl}
	
	In a single-node simulation, we tackled logistic Rregression with a dimensionality of $d=301$. The experiment involved a total of $n=142$ simulated clients and the number of rounds $r=1000$. The \dataname{W8A} dataset reshuffled u.a.r. and was partitioned into equal $n_i$ chunks. We utilized \algname{FedNL} without Line Search, Option-B, $\alpha$ using option-2. Results are presented in Table \ref{tbl:compare-in-singlenode}. It can be observed that our \algname{FedNL/RandK[K=8d]} implementation achieves a significant total speedup of $\times 929.4$ in a single-node setup compared to the baseline. Similarly, for \algname{FedNL/TopK[K=8d]}, the total speedup from our implementation is $\times 1053.9$. The master aggregates the following data from all clients:
	(i) $2\,937.0$ MBytes for \compname{RandK} and \compname{SeqK} compressors;
	(ii) $49\,568.7$ MBytes for \compname{Identical} mapping $\mathbb{C}(x)\eqdef x$ compressor;
	(iii) $4\,241.4$ MBytes for \compname{TopK};
	(iv) $358.8$ MBytes for \compname{TopLEK} compressor.
	
	We employed a two-level schema for $\nabla f_i(x)$ aggregation in our implementation. After obtaining $\nabla f_i(x)$, the master dispatches updates to one of the $4$ helper threads (configurable) in a round-robin fashion, incurring an extra $\mathcal{O}(d)$ memory storage per helper. Once all workers finish work, the master performs the final aggregation. For Hessian updates, the master utilizes another pool of configurable $4$ (configurable) helpers responsible for decompression and atomic updates to $H^k$.
	
	{
	\begin{table}[h!]
		\centering
		\begin{threeparttable}
		\caption{Single-node simulation, $n=142$, \algname{FedNL}(B), $r=1000$, $\lambda=0.001$, $\alpha$ - option 2, FP64, $24$ cores at $3.3$ GHz.}
		\begin{tabular}{|p{0.55\textwidth}|c|l|}
			\hline
			Client Compression & $\| \nabla f(x^{last}) \|$ & Total Time (seconds) \\
			\hline
			\hline
			1. \cellcolor{bgcolorwe}RandK[K=8d] (We) & $3 \cdot 10^{-18}$ & $18.84$ \\
			\hline
			2. RandK[k=8d] (Base) & $3 \cdot 10^{-18}$  & $17\,510.00$ \\
			\hline
			3. \cellcolor{bgcolorwe}TopK[K=8d] (We) & $2.80 \cdot 10^{-18}$ &  $18.72$    \\
			\hline
			4. TopK[k=8d] (Base) & $2.80 \cdot 10^{-18}$ & $19\,770.00$   \\
			\hline
			5. \cellcolor{bgcolorwe}RandSeqK[K=8d] (We) & $3.19 \cdot 10^{-18}$ & $16.70$      \\
			\hline
			6. \cellcolor{bgcolorwe}TopLEK[K=8d] (We) & $3.45 \cdot 10^{-18}$ & $18.48$   \\
			\hline
			7. \cellcolor{bgcolorwe}Natural (We) & $3.10 \cdot 10^{-18}$ & $27.02$   \\
			\hline
			8. \cellcolor{bgcolorwe}Ident (We) & $2.46 \cdot 10^{-18}$ & $24.12$   \\
			\hline
		\end{tabular}
		\label{tbl:compare-in-singlenode}
	\end{threeparttable}
	\end{table}
	}

	\begin{table}[h!]
		\centering
		\begin{threeparttable}
			\caption{Single-node simulation, $n=142$, \algname{FedNL-LS} (B), $\|\nabla f(x^{last}) \| \approx 9 \cdot 10^{-10}$, FP64, $24$ cores at $3.3$ GHz.}
			\begin{tabular}{|l|c|c|c|c|c|}
				\hline
				\makecell{\parbox{5cm}{Solver}}
				 & \makecell{W8A \\ $d=301$, $n_i=350$} & \makecell{A9A \\ $d=124$, $n_i=229$} & \makecell{Phishing\\ $d=69$, $n_i=77$} \\
				 \hline
				\hline
				\multicolumn{4}{|c|}{\textbf{Initialization Time (seconds)}} \\
				\hline
				CVXPY & +2.54 & +2.33 & +2.28      \\
				\hline
				\algname{FedNL}  & +0.939 & +0.196 & +0.081 \\
				\hline
				\multicolumn{4}{|c|}{\textbf{Solving Time (seconds)}} \\				
				\hline
				CLARABEL & 19.24 & 10.83 & 2.50 \\
				\hline
				ECOS & 22.22 & 8.02  & 2.55 \\
				\hline
				ECOS-BB & 22.00 & 8.00 &  2.12    \\
				\hline
				SCS & 31.14 & 19.36 & 4.57   \\
				\hline
				MOSEK & 16.90 & 9.59 &  3.55      \\
				\hline 
				\makecell[l]{\algname{FedNL-LS}/RandK[k=8d]} & \cellcolor{bgcolorwe} 4.35 & \cellcolor{bgcolorwe} 0.34 & \cellcolor{bgcolorwe} 0.12   \\
				\hline
				\makecell[l]{\algname{FedNL-LS}/RandSeqK[k=8d]} & \cellcolor{bgcolorwe} 3.34s & \cellcolor{bgcolorwe} 0.29 & \cellcolor{bgcolorwe} 0.06 \\
				\hline 
				\makecell[l]{\algname{FedNL-LS}/TopK[k=8d]} & \cellcolor{bgcolorwe} 4.49 & \cellcolor{bgcolorwe} 0.46 & \cellcolor{bgcolorwe} 0.10  \\
				\hline 
				\makecell[l]{\algname{FedNL-LS}/TopLEK[k=8d]} & \cellcolor{bgcolorwe} 4.79 & \cellcolor{bgcolorwe} 0.34 & \cellcolor{bgcolorwe} 0.61  \\
				\hline
				\makecell[l]{\algname{FedNL-LS}/Natural} & \cellcolor{bgcolorwe} 3.13 & \cellcolor{bgcolorwe} 0.17 & \cellcolor{bgcolorwe} 0.08   \\
				\hline
				\makecell[l]{\algname{FedNL-LS}/Identical} & \cellcolor{bgcolorwe} 0.63 & \cellcolor{bgcolorwe} 0.09 & \cellcolor{bgcolorwe} 0.06 \\
				\hline	
			\end{tabular}
			\label{tbl:compare-vs-cvxpy}
		\end{threeparttable}
	
		\centering
		\begin{threeparttable}
			\caption{Multi-node setting, $n=50$ clients, $1$ master, $|\nabla f(x^{last})|\approx 10^{-9}$, FP64, $1$ {CPU} core/node.
			}
			\begin{tabular}{|l|c|c|c|c|c|}
				\hline	
				\makecell{\parbox{5cm}{Solution}} & \makecell{W8A \\ $d=301$, $n_i=994$} & \makecell{A9A \\ $d=124$, $n_i=651$} & \makecell{Phishing\\ $d=69$,  $n_i=221$} \\
				\hline
				\hline
  			    \multicolumn{4}{|c|}{\textbf{Initialization Time (seconds)}} \\
				\hline
				Ray & \multicolumn{3}{c|}{+52.0}    \\
				\hline
				Spark &  \multicolumn{3}{c|}{+25.82}      \\
				\hline
				\algname{FedNL} &  \multicolumn{3}{c|}{+1.1} \\
				\hline
				\multicolumn{4}{|c|}{\textbf{Solving Time (seconds)}} \\
				\hline				
				Ray & 116.17 & 28.13 & 11.54 \\
				\hline
				Spark & 36.65 & 33.59 & 33.14   \\
				\hline	
				\makecell[l]{ \algname{FedNL}/RandK[k=8d]} & \cellcolor{bgcolorwe} 12.6 & \cellcolor{bgcolorwe} 4.52 & \cellcolor{bgcolorwe} 0.21   \\
				\hline
				\makecell[l]{\algname{FedNL}/RandSeqK[k=8d]} & \cellcolor{bgcolorwe} 12.56 & \cellcolor{bgcolorwe} 5.10  & \cellcolor{bgcolorwe} 0.14 \\
				\hline 
				\makecell[l]{\algname{FedNL}/TopK[k=8d]} & \cellcolor{bgcolorwe} 12.20  & \cellcolor{bgcolorwe} 5.79  & \cellcolor{bgcolorwe} 5.23  \\
				\hline 
				\makecell[l]{\algname{FedNL}/TopLEK[k=8d]} & \cellcolor{bgcolorwe} 15.11  & \cellcolor{bgcolorwe} 3.26 & \cellcolor{bgcolorwe} 0.82 \\
				\hline
				\makecell[l]{\algname{FedNL}/Natural} & \cellcolor{bgcolorwe} 5.75 & \cellcolor{bgcolorwe} 1.56 & \cellcolor{bgcolorwe} 0.14 \\
				\hline
			\end{tabular}
			\label{tbl:compare-in-multinode}
		\end{threeparttable}
\end{table}
	
	
	
	\subsection{Single-node: Extension FedNL-LS and CVXPY}
	\label{sec:single-node-cmp-vs-industry}
	
	We implemented \algname{FedNL-LS} which represents the \algname{FedNL} with global convergence guarantees (See pseudocode in Appendix \ref{app:fednl-ls-descr} or \citet{safaryan2021fednl}) both for simulation and practical distributed usage. In experiments, the line search procedure requires almost always a $1$ step. We have compared \algname{FedNL-LS} against solvers from CVXPY \citet{diamond2016cvxpy} which can solve logistic regression. The summary is presented in Table~\ref{tbl:compare-vs-cvxpy}. Plots are available in  Appendix~\ref{app:single-node-cmp-vs-industry-extra}. 
	
	We have fine-tuned tolerance for CVXPY solvers so that $\norm{\nabla f(x^k)}$ is the same. Experiment shows that our implementation already outperforms solvers from CVXPY  \citet{diamond2016cvxpy} CLARABEL \citet{clarabel}, MOSEK\citet{aps2022mosek}, SCS \citet{ocpb:16}, ECOS \citet{domahidi2013ecos} by $\times 20$. The solving times in Table~\ref{tbl:compare-vs-cvxpy} were obtained from CVXPY’s internal functions. These times reflect the actual duration used by the corresponding solver, excluding any overhead introduced by CVXPY. The initialization time refers to the period required to load all necessary Python extensions into the interpreter and to parse the data using \texttt{sklearn.datasets}. As shown for the \texttt{A9A} and \texttt{Phishing} datasets, the {initialization time alone} is $\times 7$ longer than both the initialization and solving times combined for \algname{FedNL-LS} with any compressor. We did not perform a comparison with GUROBI \cite{gurobi}. Despite its strength in solving linear and mixed-integer programming problems and accessibility to be used from CVXPY and ability to obtain an academic license, the GUROBI solver has inherent limitations when applied to logistic regression or more complex convex models (that includes intermediate transformation into exponential cone).

	Additionally, our experiments indicate that the benefits introduced by \compname{RandSeqK} are already noticeable without fine-grained measurements when  $d$ and $n_i$  are small. In this low-dimensional regime, the internal logic of the compressor and its memory access patterns become significant.
		
	\subsection{Multi Node: FedNL, Apache Spark and Rays}
	\label{sec:multi-node-cmp-vs-spark}
	
	To the best of our knowledge, the best-known open-source ready-to-use solutions that can be used for distributing training logistic regression are Apache Spark \cite{meng2016mllib} and Rays \cite{moritz2018ray}. We provide the time required for \abr{FedNL}, Rays, and Apache Spark to achieve a tolerance of $|\nabla f(x^k)|\approx 10^{-9}$ via configuring final tolerance for solvers. For details about the cluster configuration see Appendix~\ref{app:software-env}.

	From Table~\ref{tbl:compare-in-multinode}, it is evident that Rays and Apache Spark necessitate more initialization time. \abr{FedNL} surpasses alternative approaches in this benchmark not only in initialization time but in solve time as well. For extra multi-node experiments with \algname{FedNL}, \algname{FedNL-PP}, \algname{FedNL-LS} see Appendix~\ref{app:experiments-multi-node}.
	
	\clr
	{\subsection{Practical Limitations of Existing Federated Learning Systems} 
		\label{sec:existing-fl-sota-system}
		
		In cross-device Federated Learning (FL), clients typically consist of edge devices such as IoT and mobile devices. The simplest IoT devices may lack an Operating System(OS), operating in a bare-metal environment where feasible implementations can be carried out only in Assembly, C, or C++. Mobile applications, on the other hand, function in a highly competitive and performance-sensitive market. 
		
		Given the time and energy constraints of many mobile applications, Python is often the least favorable choice for mobile development (for competitive language analysis, see \cite{pereira2021ranking}). Consequently, FL systems that rely solely on Python runtimes, such as \cite{reina2021openfl} and \cite{garcia2022flute}, can face significant limitations in real-world applications. Some frameworks, like \cite{burlachenko2021fl_pytorch} and \cite{granqvist2024pfl}, explicitly state that their focus is on simulation rather than practical implementation. Interestingly, in response to real-world challenges, frameworks like \texttt{FedML} \cite{he2020fedml} and \texttt{FLOWER} \cite{beutel2020flower}, originally designed in Python, are transitioning to Java and C++ to enhance their applicability and performance. Unfortunately, existing FL frameworks \cite{he2020fedml}, \cite{beutel2020flower}, \cite{granqvist2024pfl}, \cite{roth2022nvidia}, \cite{garcia2022flute}, and \cite{ro2021fedjax} lack mechanisms for training logistic regression models robustly and autonomously without human intervention during training. As a result, our work does not compare with these frameworks. For baselines comparison, we have selected to compete against solvers accessible via CVXPY \cite{diamond2016cvxpy}, and industrial solutions such as Apache Spark \cite{meng2016mllib}, and  Ray/Scikit-Learn \cite{moritz2018ray}. Although these frameworks provide effective and robust industrial solutions, they do not fully adhere to FL principles, particularly regarding partial client participation or communication compression. Despite this limitation, we have conducted comparisons with them as they represent well-developed tools.

}

	\clearpage
	\bibliographystyle{ACM-Reference-Format}
	\bibliography{fednl-in-practice}
	\clearpage
	\tableofcontents

	
	\clearpage
	\appendix
	\onecolumn
	
	\section{Practically Implemented FedNL Extensions}
	\label{app:extra-pesudo-code-for-fednl}

	{
		\begin{algorithm}[H]
			\caption{\algname{FedNL-LS} ({\color{blue}Baseline}. Federated Newton Learn with Line Search \cite{safaryan2021fednl} \clrshort{\textbf{[EXISTENT]}})}
			\label{alg:FedNL-LS}
			\begin{algorithmic}[1]
				\STATE \textbf{Parameters:} Hessian learning rate $\alpha\ge0$; compression operators $\{\cC_1^k, \dots,\cC_n^k\}$; {line search parameters $c \in (0,\nicefrac{1}{2}]$ and $\gamma \in (0,1)$}
				\STATE \textbf{Initialization:} $x^0\in\R^d$; $\mH_1^0, \dots, \mH_n^0 \in \R^{d\times d}$ and $\mH^0 \eqdef \frac{1}{n}\sum_{i=1}^n \mH_i^0$
				\FOR{each device $i = 1, \dots, n$ in parallel} 
				\STATE Get $x^k$ from the server; compute $f_i(x^k)$,\; $\nabla f_i(x^k)$ and $\nabla^2 f_i(x^k)$
				\STATE Send $f_i(x^k)$,\; $\nabla f_i(x^k)$ and $\mS_i^k \eqdef \cC_i^k(\nabla^2 f_i(x^k) - \mH_i^k)$ to the server
				\STATE Update local Hessian shifts $\mH_i^{k+1} = \mH_i^k + \alpha\mS_i^k$
				\ENDFOR
				\STATE On Server
				\STATE \quad Get $f_i(x^k)$,\; $\nabla f_i(x^k)$ and $\mS_i^k$ from all devices $i\in[n]$
				\STATE \quad $f(x^k) = \frac{1}{n}\sum_{i=1}^n f_i(x^k), \; \nabla f(x^k) = \frac{1}{n}\sum_{i=1}^n \nabla f_i(x^k), \; \mS^k = \frac{1}{n}\sum_{i=1}^n \mS_i^k$
				\STATE \quad {Compute search direction $d^k = [\mH^k]_\mu^{-1}\nabla f(x^k)$}
				\STATE \quad {Find the smallest integer $s\ge0$ satisfying $f(x^k+\gamma^sd^k) \le f(x^k) + c\gamma^s \langle \nabla f(x^k), d^k\rangle$}
				\STATE \quad Update global model to $x^{k+1} = x^k + \gamma^sd^k$
				\STATE \quad Update global Hessian shift to $\mH^{k+1} = \mH^k + \alpha\mS^k$
			\end{algorithmic}
		\end{algorithm}
	}

	\subsection{FedNL-LS Extension: Globalization via Line Search}
	\label{app:fednl-ls-descr}

	The first extension of \algname{FedNL} that we implemented is \algname{FedNL-LS}. We selected an algorithm in which the globalization technique remains independent of any problem-specific parameters. The pseudocode for this algorithm, presented as Algorithm \ref{alg:FedNL-LS}, has been sourced from \cite{safaryan2021fednl}.

	The \algname{FedNL} family includes an extension incorporating a globalization strategy through cubic regularization \algname{FedNL-CR}. However, its practical implementation requires knowledge of problem-dependent $L_*$.
	
	\clearpage
	
		{
		\begin{algorithm}[h!]
			\caption{\algname{FedNL-PP} ({\color{blue}Baseline}. Federated Newton Learn with Partial Participation \cite{safaryan2021fednl} \clrshort{\textbf{[EXISTENT]}})}
			\label{alg:FedNL-PP}
			\begin{algorithmic}[1]
				\STATE {\bfseries Parameters:} Hessian learning rate $\alpha>0$; compression operators $\{\cC_1^k, \dots,\cC_n^k\}$; {number of participating devices $\tau \in \{1,2,\dots,n\}$}
				\STATE {\bfseries Initialization:}
				For all $i\in [n]$: $w^0_i = x^0 \in \R^d$; $\mH_i^0 \in \R^{d\times d}$; $l_i^0 = \|\mH_i^{0} - \nabla^2 f_i(w_i^{0})\|_{\rm F}$; $g_i^0 = (\mH_i^{0} + l_i^{0} \mI)w_i^{0} - \nabla f_i(w_i^{0})$; Moreover: $\mH^0 = \frac{1}{n} \sum_{i=1}^n \mH_i^0$; $l^0 = \frac{1}{n} \sum_{i=1}^n l_i^0$; $g^0 = \frac{1}{n} \sum_{i=1}^n g_i^0$
				\STATE \textbf{on} server
				\STATE ~~~ $x^{k+1} = \left(  \mH^k + l^k\mI  \right)^{-1} g^k$ \hfill { \scriptsize Main step: Update the global model}
				\STATE ~~~ {Choose a subset $S^{k} \subseteq \{1,\dots, n\}$ of devices of cardinality $\tau$, uniformly at random}
				\STATE ~~~ Send $x^{k+1}$ to {the selected devices $i\in S^k$} \hfill { \scriptsize Communicate to selected clients}
				\FOR{each device $i = 1, \dots, n$ in parallel}
				\STATE {{\bf for participating devices} $i \in S^k$ {\bf do} }
				\STATE $w_i^{k+1} = x^{k+1}$ \hfill { \scriptsize Update local model}
				\STATE $\mH_i^{k+1} = \mH_i^k + \alpha \cC_i^k(\nabla^2 f_i(w_i^{k+1}) - \mH_i^k)$ \hfill { \scriptsize Update local Hessian estimate}
				\STATE $l_i^{k+1} = \|\mH_i^{k+1} - \nabla^2 f_i(w_i^{k+1})\|_{\rm F}$ \hfill { \scriptsize Compute local Hessian error}
				\STATE $g_i^{k+1} = (\mH_i^{k+1} + l_i^{k+1} \mI)w_i^{k+1} - \nabla f_i(w_i^{k+1})$ \hfill { \scriptsize Compute Hessian-corrected local gradient}
				\STATE Send $\cC_i^k(\nabla^2 f_i(w_i^{k+1}) - \mH_i^k)$,\; $l_i^{k+1} - l_i^k$ and $g_i^{k+1} - g_i^k$ to server \hfill { \scriptsize Communicate to server}
				\STATE { {\bf for non-participating devices} $i \notin S^k$ {\bf do} }
				\STATE $w_i^{k+1} = w_i^k$, $\mH_i^{k+1} = \mH_i^k$, $l_i^{k+1} = l_i^k$, $g_i^{k+1} = g_i^k$  \hfill { \scriptsize Do nothing}
				\ENDFOR
				
				\STATE On Server
				\STATE ~~~ $g^{k+1} = g^k + \frac{1}{n}\sum_{i\in S^k} \left(  g_i^{k+1} - g_i^k  \right)$  \hfill { \scriptsize Maintain the relationship $g^k = \frac{1}{n} \sum_{i=1}^n g_i^k$}
				\STATE ~~~ $\mH^{k+1} = \mH^k + \frac{\alpha}{n}\sum_{i\in S^k} \cC_i^k(\nabla^2 f_i(w_i^{k+1}) - \mH_i^k)$   \hfill { \scriptsize Update  the Hessian estimate on the server}
				
				\STATE ~~~ $l^{k+1} = l^k + \frac{1}{n}\sum_{i\in S^k} \left(  l_i^{k+1} - l_i^k  \right)$ \hfill { \scriptsize Maintain the relationship $l^k = \frac{1}{n} \sum_{i=1}^n l_i^k$}
			\end{algorithmic}
		\end{algorithm}
	}

	\subsection{FedNL-PP Extension: FedNL with Clients Partial Participation}
	\label{app:fednl-pp-descr}

	The subsequent extension to \algname{FedNL}, which we incorporate into our practical implementation, is \algname{FedNL-PP} from the original \algname{FedNL} paper by \cite{safaryan2021fednl} (Algorithm \ref{alg:FedNL-PP}). This extension enables the handling situation, where only randomly selected clients participate in each iteration.

	
	


\clearpage
\section{Progression of Practical Enhancements: Chronological Overview}
\label{app:history-of-improvements}

There is an inherent complexity involved in concurrently addressing three cornerstones:

\begin{enumerate}
	\item \textit{{Theoretical Algorithm:}} serving as a foundational framework.
	\item \textit{{Modern Computation Systems:}} serving as efficient and reliable computing backbone.
	\item \textit{{Best Engineering Practices:}} also serves as a bridge between the previous two aspects.
\end{enumerate}

In our work, we transition a well-developed theoretical algorithm analyzed with worst-case guarantees, exploring possible practical improvement gains. We adhered to a set of decisions in creating a practical implementation, taking steps that do not violate theory at all on one hand, but from another side, they target to make \algname{FedNL} practical.

In refining \algname{FedNL}, our focus was on addressing $L_2$ regularized logistic regression as a proxy which is used for measure improvements. The LIBSVM \dataname{W8A} dataset was chosen for experimentation. Each sample in the dataset was augmented with an artificial feature set to $1$ to introduce an intercept term. Post-augmentation, our dataset \dataname{W8A} has $d=301$ features, and encompasses $49\,749$ samples. Uniform shuffling and equitable distribution across $n=142$ clients were ensured, with each receiving a share of $350$ sample data points, while the remaining $49$ samples were excluded. The initial value of the optimization variable $x$ is initialized at zero. The regularization coefficient is $\lambda=0.001$. In our instance, the strong convexity parameter is $\mu_f \ge 0.001$ and smoothness contact is  $L_f \le 0.0058$. This lead to condition number $\frac{L_f}{\mu_f} \le 5.8$. We executed $1000$ rounds.

\paragraph{Experiments setup} All our experiments have been carried out under the Assumption that the preliminary step from Appendix~\ref{app:careful-reproduce} was done, and next, the launching has been done $4$ times with minimum time from $4$ launches. The employed hardware for single-node multi-core simulation is described in Appendix~\ref{app:hardware-env-single-node}. The start iterate is $x^0=0$.  Our \algname{FedNL} implementation contains function value evaluation $f(x)$ that is optionally tracked and computed via obtaining this information from clients. Norm of a full gradient in last iterate $\| \nabla f(x^k)\| \approx 3 \cdot 10^{-18}$ for \algname{FedNL} with \compname{RandK[k=8d]} compressor, and $\| \nabla f(x^k)\| \approx 3.88 \cdot 10^{-18}$ for \algname{FedNL} with \compname{TopK[k=8d]} compressor. 

\paragraph{Improvements history} The detailed history of improvements presented in the Table \ref{tab:improvements-log} below. The total speedup of our implementation in a single-node multi-core machine \texttt{v63} compared to baseline \texttt{v0} is {$\times 951.88$} for \compname{RandK[k=8d]} and {$\times 1069$} for \compname{TopK[k=8d]}.

{
	\footnotesize
	
	\begin{longtable}{|p{0.46\textwidth}|p{0.07\textwidth}|p{0.13\textwidth}|p{0.07\textwidth}|p{0.13\textwidth}|}
		\caption{Improvemens for \compname{TopK[K=8d]} and  \compname{RandK[K=8d]}, $d=301,n_i=350,n=142$, \dataname{W8A}.}
		\label{tab:improvements-log} \\
		\hline
		\textbf{Improvement Step} & \textbf{TopK. Time(sec.)} & \textbf{TopK. Relative speedup $t_{i+1}/t_{i}$} & \textbf{RandK. Time(sec.)} & \textbf{RandK. Relative speedup $t_{i+1}/t_{i}$}
		\endfirsthead
		\hline		
		\textbf{Improvement Step} & \textbf{TopK. Time(sec.)} & \textbf{TopK. Relative speedup $t_{i+1}/t_{i}$} & \textbf{RandK. Time(sec.)} & \textbf{RandK. Relative speedup $t_{i+1}/t_{i}$} \\
		\endhead
		\hline 
		\endfoot
		\hline
		
		v63. Compiler and Language relative improvements. & 18.488 & 1.002 & 18.395 & 1.002 \\
		\hline		
		v62. Improvements inside pseudo-random generators and Knuth shuffle implementation. & 18.525 & 1.0105 & 18.438 & 1.0218 \\	
		\hline
		v61. Optimize data alignment in dense linear algebra operations by transitioning to AVX-512 vectorization from AVX2(256 bits). & 18.72 & 1.3633 & 18.84 & 1.3305 \\	
		\hline
		v60. Compiler improvements for local FedNL implementation. & 25.521 & 1.0018 & 25.067 & 0.9954 \\
		\hline		
		v59. Bit tricks for computing residual from division by a number which is a power of two. & 25.569 & 1.00406 & 24.953 & 1.021239 \\
		\hline
		v58. Use more aligned load/store in Hessian oracle. & 25.673 & 1.016 & 25.483 & 1.019 \\ 		
		\hline
		v57. Use aligned load/store for dense vector operations. & 26.104 & 1.10423 & 25.787 & 1.0895 \\
		\hline
		v56. Eliminating not need for information in memory buffers, and storing vector of labels explicitly. & 28.825 & 1.000173 & 28.097 & 1.03185 \\		
		\hline
		v55. Precompute indices without rechecking in a hot-loop for RandK and other minor improvements. & 29.098 & 1.014 & 28.992 & 1.0159 \\		
		\hline		
		v54. Eliminate division during memory allocation from constructed memory pools. & 29.542 & 1.0002 & 29.455 & 1.0008 \\		
		\hline		
		v53. Eliminate the storage of design matrix $A^\top$. Add need methods for matrix-vector multiplication with $A^\top$. & 29.549 & 1.13144 & 29.480 & 1.12869 \\
		\hline
		v52. Additional internal vectorization across samples with reducing stores during hessian evaluation. & 33.433 & 1.6303 & 33.274 & 1.6267 \\
		\hline
		v51. Use symmetry during evaluating $\norm{.}_{F}$. & 54.506 & 1.00066 & 54.128 & 1.00751 \\
		\hline
		v50. Unroll the first iteration in Hessian Oracle. & 54.542 & 1.0489 & 54.535 & 1.0351 \\
		\hline
		v49. More space-efficient method for TopK based on min-heaps. & 57.211 & 1.012 & 56.451 & 1 \\
		\hline
		v48. Optimized version for generating sequences with SSE/AVX vectorization. & 57.940 & 1.014 & 56.451 & 1.006 \\
		\hline
		v47. More Systems Optimization. Use memory-mapped files for parsing input data. & 58.788 & 1.023 & 56.802 & 1.022 \\		
		\hline
		v46. Yield client execution while waiting for a signal from the server instead of a complete busy loop. & 60.198 & 1.0057 & 58.102 & 1.0005 \\
		\hline
		v45. Use light vector for gradient aggregations and minor improvement of Cholesky Factorization. & 60.544 & 1.0016 & 58.393 & 1.0029 \\
		\hline
		v44. Compute and send $L_k$ once the difference that defines $L_k$ is available. This quantity is in a critical path. & 60.644 & 1.003 & 58.567 & 1.013 \\
		\hline		
		v43. Compute indices for the RandK while waiting for the master. Add a RandSeqK compressor (which works $58.4$ sec.). & 60.826 & 1.0 & 59.358 & 1.0211 \\
		\hline		
		v42. Remove not needed conditions in matrix-vector and matrix-matrix operations. Create extra conversion methods to obtain a flattened index in the matrix. Fused operation for matrix-vector operation and add multiple of vector. & 60.826 & 1.0203 & 60.618 & 1.0342 \\
		\hline
		v41. Sorting indices for RandK and TopK to make computation more cache-friendly. & 62.063 & 1.0182 & 62.697 & 1.01033 \\
		\hline
		v40. Force inlining for compression mechanisms. & 63.193 & 1.007 & 63.345 & 1.0005 \\
		\hline
		v39. Switch from creating from the number of userspace threads equal to the number of clients. Create a pool of workers equal to the number of physical cores available in the system (in our case it is $12$). Process several clients or work items within one thread. Also, eliminate extra copying of current iterate $x_i$ inside workers. Process messages from the clients once they are available. & 63.647 & 1.404 & 63.379 & 1.819 \\
		\hline
		v38. Improvement in parsing and analyzing input files in LIBSVM dataset format. Elimination of creating temporary strings with low-level primitives from C++17/20. & 89.368 & 1.0835 & 115.340 & 1.0533 \\
		\hline
		v37. Improved TopK implementation. Through various approaches - quicksort, merge sort, Multi-way merge sort, CO Funnelsort, and radix sort the fastest approach is based on a D-way heap to support K smallest items found so far. & 96.833 & 1.0412 & 121.490 & $\approx$ 1.0 \\
		\hline
		v36. Applying it to sparse updates of the dense matrix in the server. Cache-friendly implementation and usage of byte buffers by clients used to prepare information to send to the master. & 100.829 & 1.40216 & 121.499 & 1.44588 \\
		\hline
		v35. Custom memory pools for memory allocations for dense vectors and matrices. The memory pool implementation does not allow moving vectors from one computation CPU thread to another. However, this makes memory allocation very fast. & 141.379 & 1.0419 & 175.673 & 1.0109 \\		
		\hline
		v34. Store information about the number of columns in a dense matrix explicitly without recomputing it every time it is needed. & 150.207 & 1.0052 & 177.601 & 1.0421 \\
		\hline
		v33. Add vectorized implementation for light dense vector (light read-only view of a dense vector) for $L_2$ norm and inner product. Useful in Backward/Forward substitution and Cholesky factorization implementation in Master. & 150.998 & 1.0689 & 185.093 & 1.021 \\
		\hline
		v32. Manually unroll loops for vector and vector scalar operations in dense vector vectorized implementation.  & 161.410 & 1.034 & 189.075 & 1.000 \\
		\hline
		v31. Compute and store indices for the upper triangular part of the matrix with shape [d,d] only once.  Reuse these indices in all rounds without recomputing them for TopK and RandK. & 166.899 & 1.045 & 189.134 & 1.0165 \\
		\hline
		v30. Cache-friendly Cholesky Factorization implementation which produces both L and L transpose factors, that subsequently used cache-friendly forward/backward substitution. & 174.495 & 1.034 & 192.271 & 0.986 \\
		\hline
		v29. Turn on the whole program optimization with the linker. & 180.588 & 1.047 & 189.612 & 1.034 \\
		\hline
		v28. Turn off support of exception and runtime type information from the C++ compiler. & 189.216 & 1.07 & 196.108 & 1.04 \\
		\hline
		v27. Extra vectorization in CPU implementation of Hessian oracle. & 204.315 & 1.02 & 204.219 & 1.001 \\
		\hline 
		v26. An alternative implementation of logistic regression Hessian Oracle. Evaluate as the sum of symmetric Rank-1 matrices. Compute only the upper diagonal part and symmetrize once. & $208.937$ & $1.603$ & $204.440$ & $1.805$ \\
		\hline 
		v25. Improve diagonal matrix multiplication by dense. Improve Matrix-Matrix multiplication - tune tile sizes. & $334.992$ & $1.089$ & $369.037$ & $1.025$ \\
		\hline 
		v24. Removing one division during conversion from flat matrix index into matrix row/column index.  Use uninitialized buffers inside hessian/function/gradient oracles. & $364.886$ & $1.14$ & $378.563$ & $1.26$ \\
		\hline
		v21. Compute the exponent of classification margins only once during computing hessian/gradient oracles. (I.e. compute $\exp(x^\top a_i \cdot b_i)$ only once and reuse them in all oracle types). & $406.550$ & $1.0004$ & $477.912$ & $1.04$ \\
		\hline
		v20. More effective initialization of design matrix (by columns). Remove double default initialization of dense matrix. & $406.730$ & $1.057$ & $497.912$ & $1.02$ \\
		\hline
		v17. Compute classification margin once (i.e. $x^\top \cdot b_{ij} a_{ij}$) and reuse this quantity in gradient and hessian oracle. & $430.058$ & $1.458$ & $511.612$ & $1.44$ \\
		\hline
		v16. Explicit call size() functions in all places of source code out of the loops and use it once without recalling the function again. & $627.172$ & $1.08$ & $737.735$ & $1.08$ \\
		\hline
		v15. Move $l_i$ information into the separate data stream for sending information asynchronously to the master.  & $681.874$ & $1.01$ & $798.195$ & $1.03$ \\
		\hline
		v14. Custom implementation of adding elements to diagonal. & $689.630$ & $1.06$ & $824.973$ & $1.019$ \\
		\hline
		v12. Shuffle the array in place instead of shuffling a separate array. Special copy operations for big large objections of float/doubles. & $734.119$ & $1.005$ & $841.172$ & $1.073$ \\
		\hline
		v10. Cholesky decomposition with forward/backward substitution instead of Gauss Elimination during linear solve. & $737.92$ & $1.146$ & $903.392$ & $1.196$ \\
		\hline	
		v09. Sorting-based implementation for TopK instead of using D-way heaps. Create a separate dedicated stream of sending information with information about gradients. & $845.982$ & $1.008$ & $1080.820$ & $1.012$ \\
		\hline
		v7. Extra constant and no exception qualifiers into various places - make code nicer for compiler optimization. Simplify dataset representation. Remove information about that presented sample from the dataset. Add tile matrix multiplication instead of baseline. & $853.032$ & $1.016$ & $ 1093.900$ & $1.011$ \\
		\hline
		v1. Baseline. Usual C++ implementation of the algorithm. & $867.227$ & $22.8027$ & $1106.52$ & $15.8318$ \\
		\hline
		v0. Baseline implementation in Python/Numpy  \href{https://github.com/Rustem-Islamov/FedNL-Public}{https://github.com/Rustem-Islamov/FedNL-Public}. & $19770.0$ & $1$ & $17510.0$ & $1$ \\	
		\hline
	\end{longtable}	
	}

	\clearpage
	\section{RandSeqK: A Practical  Cache-Aware Improvement of RandK}
	\label{app:seqk}
	
	\subsection{Background about RandK Compressor}
	\label{app:randk}
	
	The sparsification of the input matrix $M \in \mathbb{R}^{d \times d}$ with \compname{RandK} happens by selecting $k$ items from possible $n=d(d+1)/2$ elements from the upper triangular part of the input matrix $M$ and scaling the result in a way to preserve unbiasedness. The distribution employed for selecting this subset is uniform across all possible subsets of cardinality $k$. The specific subset is selected with probability $\left(\frac{k!}{n! (n-k)!}\right)^{-1}$, and the probability that a specific element will be included in the selection equals to:
	
	\begin{eqnarray*}
		p &=& \left(\frac{(k-1)!}{(n-1)!((n-1)-(k-1))!}\right)/
		\left(\frac{k!}{n!(n-k)!}\right) = \left(\frac{(k-1)!}{(n-1)!(n-k)!}\right)/
		\left(\frac{k!}{n!(n-k)!}\right) \\
		&=& \left(\frac{(k-1)!n}{n!(n-k)!}\right)/
		\left(\frac{k!}{n!(n-k)!}\right) = \frac{(k-1)!n}{k!} = \frac{k}{n}
	\end{eqnarray*}
	
	This implies that specific items will be selected during sparsification can be described as indicator random variables which take $1$ with probability $\dfrac{k}{n}$, and $0$ with probability $1-\dfrac{k}{n}$.
	
	\paragraph{RandK as a collection of Bernoulli Random Variables} The \compname{RandK}($\bM$) compressor applied for symmetric matrices from $\mathbb{R}^{d \times d}$ can be observed as an operator that selects elements from the upper triangular part and scales the results by scalar constant $C$. Let's use notation whereby with $e_{ij} \in \mathbb{R}^{d \times d}$ we denoted the matrix from $\mathbb{R}^{d \times d}$ with the only non-zero element in positions $(i,j)$ and $(j,i)$ equal to $1$. If $\delta \eqdef \Exp{Z_{mn}}, \forall m, n \in [d]$ we proceed as follows:
	\begin{eqnarray*}	
		\Exp{RandK(\bM)} &=& \Exp{C \sum_{i=1}^d \sum_{j=i}^d  e_{ij} Z_{ij} m_{ij}}=\sum_{i=1}^d \sum_{j=i}^d C \Exp{Z_{ij}} e_{ij} m_{ij} \\
		&=& C \cdot \delta \sum_{i=1}^d \sum_{j=i}^d e_{ij} m_{ij} = C \cdot \delta \cdot \bM.
	\end{eqnarray*}
	
	To ensure unbiased compression, there is only one choice for the constant $C=\dfrac{1}{\delta}$.
	
	\textbf{Observation 1:} \textit{Although we utilized the fact that 
		$\Exp{Z_{ij}}$ does not vary across matrix elements, during the derivation, we did not explicitly rely on any specific dependence between $Z_{ij}$ r.v.
	}
	
	Next, the derivations for the theory of \compname{RandK} for bounding variance can be simplified by observing that ${Z_{ij}^2}={Z_{ij}}$ due to $Z_{ij}$ being an indicator variable. We proceed as follows:
	
	\begin{eqnarray*}	
		\Exp{\|\mathrm{RandK}(\bM)-\bM\|_F^2} &=& 
		\Exp{\sum_{i=1}^d \sum_{j=1}^d \left( [\mathrm{RandK}(\bM)]_{ij} - m_{ij} \right)^2} \\
		&=& 
		\Exp{\sum_{i=1}^d \left( C\cdot Z_{ii}m_{ii} - m_{ii} \right)^2} +
		2 \Exp{\sum_{i=1}^d \sum_{j=i+1}^d \left( C\cdot Z_{ij}m_{ij} - m_{ij} \right)^2} \\
		& = & \Exp{\sum_{i=1}^d \left( C\cdot Z_{ii} - 1 \right)^2 m_{ii}^2} + 2 \Exp{\sum_{i=1}^d \sum_{j=i+1}^d \left(C\cdot Z_{ij} - 1\right)^2 m_{ij}^2} \\
		& = & {\sum_{i=1}^d \Exp{(C^2\cdot Z_{ii}^2 - 2C Z_{ii} + 1)} m_{ii}^2} +
		2{\sum_{i=1}^d \sum_{j=1}^d \Exp{(C^2\cdot Z_{ij}^2 - 2C Z_{ij} + 1)} m_{ij}^2}
		\\
		& = & {\sum_{i=1}^d \Exp{(C^2\cdot Z_{ii}^2 - 1)} m_{ii}^2} + 
		2{\sum_{i=1}^d \sum_{j=i+1}^d \Exp{(C^2\cdot Z_{ij}^2 - 1)} m_{ij}^2} \\
		& = & {\sum_{i=1}^d \Exp{(C^2\cdot Z_{ii} - 1)} m_{ii}^2} + 
		2{\sum_{i=1}^d \sum_{j=i+1}^d \Exp{(C^2\cdot Z_{ij} - 1)} m_{ij}^2} \\	 
		&=& (C - 1) \|\bM\|_F^2.	
	\end{eqnarray*}
	
	Therefore for \compname{RandK} compressor $w=\dfrac{1}{\delta}-1=C-1$. 
	
	\textbf{Observation 2:} \textit{During derivation to bound variance 
		we utilized the fact that $\Exp{Z_{ij}}$ does not vary across matrix elements, however, we did not explicitly rely on any specific dependence between $Z_{ij}$ r.v.
	}
	
	\subsection{Looking for a Degrees of Freedom in RandK Compressor}
	\label{app:dof-in-randk}
		
	The theoretical derivation for \compname{RandK} does not prescribe the joint distribution of $Z_{ij}, i \le j$, therefore analysis of \compname{RandK} is more general and can be applied to other compression algorithms. The only requirement to maintain previous mathematical analysis is that $\Exp{Z_{ij}}$ is constant equal to $\delta$. This will imply automatically that $w=1/\delta - 1$ for a compressor that specifies input leavening non-zero (and scale by $C$) items from the upper triangular part of input matrix $\bM$ specified by $Z_{ij}$.
	
	\subsection{Compression Schema: Cache-Aware RandSeqK}
	The proposed modification of the \compname{RandK} compressor introduces a sampling strategy that ensures a cache-aware memory access pattern during the compression of a matrix $M \in \mathbb{R}^{d \times d}$ arranged densely in column-wise order. If we order all elements from the upper triangular part into the sequence $E$, its size becomes $w={d(d+1)}/{2}$. The sampling strategy begins by randomly selecting a start index $s \sim_{u.a.r.} E$. The subsequent set of indices is created deterministically:
	
	$$I=\{s, (s+1)\mod w, \dots, (s+k-1) \mod w \}$$
	
	The \compname{RandSeqK} compressor is then applied and utilizes the same logic of \compname{RandK}, with a difference that the selection of $Z_{ij}$ is modified in the following sense $Z_{ij}=1 \iff (i,j) \in I \subseteq E$.
	
	The number of indices in set $I$ is $k$, implying that exactly $k$ indicator random variables $Z_{ij}$ will be equal to $1$. These random variables are sampled in a dependent manner. The total number of groups from which sampling occurs is $w$, and each group consists of:
	$$\{i \mod w, (i+1) \mod w, \dots, (i+k-1)\mod w\}.$$
	
	For each element in the original sequence $e$, there are $k$ groups with sequential indices that cover this element. The probability of selecting a specific element (i.e., $Z_{ij}=1$) from the sampling schema—where the start index $s$ is initially sampled, and then the next $k-1$ are committed to, is equal to $k/n$. It does not depend on the indices $(i,j)$, allowing to reuse theory from Appendix~\ref{app:randk}.
	
	\subsection{Advantages of RandSeqK over RandK}
	
	The \compname{RandSeqK} requires only $1$ pseudo-random generator (PRG) call, while \compname{RandK} demands $k$ calls.
	
	In scenarios where the input matrix is located in DRAM memory and is primed for compression, both compressors necessitate fetching $k$ items, each comprising $b$ bytes ($b=8$ for FP64). For \compname{RandK}, this results in a worst-case scenario of $k$ DRAM memory transactions. The \compname{RandSeqK} faces a worst-case scenario of ${kb}/{L}+2$ memory transactions, where $L$ denotes the cache line size. In contemporary systems, the cache line size typically stands at $64$ bytes. For FP64, \compname{RandSeqK} markedly amplifies memory access efficiency in both clients (Algorithm~\ref{alg:FedNL}, Lines $5$ and $6$) and the master (Algorithm~\ref{alg:FedNL}, Line $10$) by a factor of $L/b=8$. This improvement may seem negligible, but if considering data fetching from DRAM memory based on Table \ref{tbl:latencies-for-memory} from Appendix \ref	{app:memory-hierachy-latencies}, the scaled improvement becomes $\times 2640$ if make comparison with float point arithmetics.
	
	Contemporary ARM and x86 processors often integrate hardware components for data prefetching, enhancing memory access latency from DRAM. Specific details about the number and types of prefetchers in CPUs are under Non-Disclosure Agreements. While \compname{RandK} selects items for compression uniformly at random, \compname{RandSeqK} exhibits a more structured and predictable access pattern for prefetchers, contributing to improved efficiency in memory operations.
	
	\clearpage
	\section{TopLEK: A Randomized and Adaptive Improvement over TopK}
	\label{app:toplek}
	
	\subsection{Background about TopK Compressor}
	
	To simplify exposition, let's initially focus on the class of contractive compressors to vectors:
	
	\begin{equation}
		\label{def:biased_compressors}
		\{\mathcal{C} \in \mathbb{R}^d \to \mathbb{R}^d : \mathbb{E}\left[ \|\mathcal{C}(x) - x\|^2\right]\} \le (1-\alpha)\|x\|^2, \forall x \in \mathbb{R}^d, \alpha \in (0,1]\}.
	\end{equation}
	
	An important example of a compress operator that satisfies this property is the \compname{TopK} compress operator. This operator preserves the $k$ largest (in absolute value) entries of the input, zeroing out the rest. In this case, the compression algorithm is deterministic (if breaks ties in the same way).
	
	The deterministic \compname{TopK} compressor satisfies condition in Eq. \ref{def:biased_compressors} deterministically: $$\|\mathrm{TopK}(x) - x\|^2\ \le (1-\alpha)\|x\|^2, \forall x \in \mathbb{R}^d.$$
	
	Firstly, we observe that \compname{TopK} is homogeneous in $\beta \in \mathbb{R}, \beta \ne 0$. For $\forall x \in \RD$:
	\begin{eqnarray*}
		\|\mathrm{TopK}(x) - x\|^2\ \le (1-\alpha)\|x\|^2 \iff
		\|\mathrm{TopK}(\beta x) - \beta x\|^2\ \le (1-\alpha)\|\beta x\|^2 \iff \\
		\|\mathrm{TopK}(\beta x) / \beta - x\|^2\ \le (1-\alpha)\|x\|^2,  \iff
		\|\mathrm{TopK}(x) - x\|^2\ \le (1-\alpha)\|x\|^2
	\end{eqnarray*}
	
	\paragraph{Comments} The \compname{TopK} compressor returns the $k$ largest values in absolute value. Scaling by any nonzero number does not change the result of the compress operator (if ties are broken in the same way). After returning the result, $\mathrm{TopK}(\beta x) / \beta$ correctly reconstitutes the sign. Thus, we can conclude that \compname{TopK} is a homogeneous function.
	
	\subsection{Pessimism of TopK Analysis} If $x \in \RD \backslash \{0\}^d$ and selected indices for \compname{TopK} compressor is the set $S(x) \in 2^{[d]}$ then we have:
	
	\begin{eqnarray*}
		\|\mathrm{TopK}(x) - x\|^2 = \|x-\mathrm{TopK}(x)\|^2 \le (1-\alpha)\|x\|^2, \forall x \in \RD &\iff \\
		\sum_{i=1}^{d} x_i^2 - \sum_{j \in S(x)} {x_{j}}^2 \le (1-\alpha) \sum_{i=1}^{d} x_i^2, , \forall x \in \RD &\iff \\
		1 - \dfrac{(\sum_{j \in S(x)} {x_{j}}^2)}{(\sum_{i=}^{d} x_i^2)} \le (1-\alpha), \forall x \in \RD \iff \alpha \le \dfrac{(\sum_{j \in S(x)} {x_{j}}^2)}{(\sum_{i=}^{d} x_i^2)}, \forall x \in \RD
	\end{eqnarray*}
	
	With utilizing the set $S(x)$ and $S'(x) \eqdef [d] \setminus S(x)$ the optimal $\alpha_{\rm{opt}}$ can be expressed as:
	
	\begin{eqnarray*}
		\alpha_{\rm{opt}} &=&{\min}_{x \in \RD} {\left(\sum_{j \in S(x)} {x_{j}}^2\right)}, \mathrm{subject\,to:\,} \|x\|_2^2=1 \iff \\
		\alpha_{\rm{opt}} &=&{\min}_{x \in \RD} {\left(1 - \sum_{j \in S'(x)} {x_{j}}^2\right)}, \mathrm{subject\,to:\,} \|x\|_2^2=1.
	\end{eqnarray*}
	
	One using Lagrangian for fixed sets $S$ and $S'$ can observe that stationary points are only attained when $\forall i,j: x_i=x_j$. Because the problem is bounded there is no need to analyze behavior once $x \to \inf$ in $\RD$. For whole $\RD$ $\alpha_{\rm{opt}}$ attained minimum value $\frac{k}{d}$ only in the diagonal of the $\RD$. 
	
	In most practical computer implementations the set $\mathbb{R}$ is implemented as a set of finite $M$ values. In this case, the number of diagonal elements in such a finite representation is $M$, and the number of non-diagonal elements is $M^d - M$. Diagonal is only a small part of the practically implementable elements of $\mathbb{R}^d$. Even though $\alpha=\frac{k}{d}$ is worst case $\alpha$ for \compname{TopK} compressor, it is too pessimistic.

	\subsection{Compression Schema: Adaptive TopLEK Compressor}
		
	The worst-case contraction factor from Definition \ref{def:biased_compressors} for \compname{TopK} compressor may be too pessimistic. One possibility is to generalize the definition of $\alpha \in \mathbb{R}$ to $\alpha(x)$ where $x$ is input to the compressor. If followed this way, then $\alpha(x)$ will vary in each client in each round. Serious reworking is required to analyze the convergence of the Lyapunov function for \algname{FedNL} in this case. 
	
	We propose an alternative and practical approach to address this situation. Instead of introducing \textit{adaptivity} from $\alpha(x)$ directly into Optimization Algorithms, we incorporate \textit{adaptivity} into the compression mechanism. Clients during employing \algname{FedNL} obtain information about value $k$ at the beginning of the training. Then they compute $1-\alpha=1 - k/d$ also at the beginning of the optimization process to define contraction value $1-\alpha$ from Definition~ \ref{def:biased_compressors}. Next in the compression phase clients return not $k$ values, but $k'$ items $0 \le k' \le k$ with the strategy described in Algorithm~\ref{alg:top_le_k}.

	{
	\begin{algorithm}
		\begin{algorithmic}[1]
			\STATE {\bfseries Input:} Input for compressor $x \in \RD$; Integer value $0 < k \le d$;
			\STATE  Compute $1-\alpha = 1 - \dfrac{k}{d}$
			\STATE  Step $1$: Compute $\dfrac{\|\mathrm{TopK}[K=k](x) - x\|^2}{\|x\|^2}  = (1-\alpha_{k}(x))$
			\STATE $\dots$
			\STATE  Step $K+1$: Compute $\dfrac{\|\mathrm{TopK}[K=0](x) - x\|^2}{\|x\|^2}  = (1-\alpha_{0}(x)) = 1$
			\STATE Find step $i \in [K]$ such that $(1-\alpha_i(x)) \ge 1-\alpha \ge (1-\alpha_{j}(x))$ where $j=i-1$. \\ 
			Condition is equivalent to $\alpha_i(x) \le \alpha \le \alpha_{j}(x)$.
			\STATE Sample a Bernoulli random variable and:
			\begin{enumerate}
				\item Compress $x$ with \compname{TopK} and $K$ defined from step $i$ with probability ${p}$
				\item Compress $x$ with \compname{TopK} and $K$ defined from step $j$ with probability $1-p$			
			\end{enumerate}
		\end{algorithmic}
		\caption{\algname{TopLEK}: Top Less Equal K Compressor \clrshort{\textbf{[NEW]}}}
		\label{alg:top_le_k}
	\end{algorithm}
	}

	In Algorithm \ref{alg:top_le_k} step $i$ and step $j \eqdef i-1$ corresponds to \compname{TopK} with $K_j > K_i$. The running value of $k$ is decreasing during the execution of Algorithm \ref{alg:top_le_k}. As steps progress leads to $1-\alpha_i \to 1$. In particular for step $K+1$ the value $1-\alpha(K=0) = 1$. Therefore we will find steps $i,j$ such that:
	\begin{eqnarray*}
		(1-\alpha_i(x)) \ge 1-\alpha \ge (1-\alpha_{j}(x)), j \eqdef i-1.
	\end{eqnarray*}	
	
	To make Eq.\ref{def:biased_compressors} as tight as possible we compress in a randomized way with \compname{TopK[$K_i$]} compressor employed with probability ${p}$, and with \compname{TopK[$K_j$]} compressor with probability $1-{p}$, we ${p}$ derived as:
	
	\begin{eqnarray*}
		&&\mathbb{E}\left[ \|\mathcal{C}(x) - x\|^2\right] \le (1-\alpha) \|x\|^2 \\
		&\iff& p \left[ \|\mathcal{C}_i(x) - x\|^2\right] + (1-p) \left[ \|\mathcal{C}_j(x) - x\|^2\right] \le (1-\alpha) \|x\|^2 \\
		&\iff& (p (1-\alpha_{i}) + (1-p) (1-\alpha_{j})) \|x\|^2 \le (1-\alpha) \|x\|^2 \\
		&\iff& p -p \alpha_{i} + 1-\alpha_{j} - p(1-\alpha_{j}) = 1-\alpha \\
		&\iff& -p \alpha_{i} -\alpha_{j} + p\alpha_{j} = -\alpha \iff p (\alpha_{j} - \alpha_{i}) = \alpha_{j} - \alpha \iff p = \dfrac{\alpha_{j} - \alpha}{\alpha_{j} - \alpha_{i}}
	\end{eqnarray*}
	
	The last equation which defines ${p}$ has non-negative values 
	in the numerator and denominator. Due to construction of $\alpha_{j},\alpha, \alpha_i$ 
	namely $\alpha_i(x) \le \alpha \le \alpha_{j}(x)$ the value of ${p} \le 1$. Therefore $p \in [0,1]$, represents a valid probability. The constructed \compname{TopLEK[K]} compressor contractive inequality Eq.\ref{def:biased_compressors} attains tight equality duing using \algname{FedNL}, i.e. $$\mathbb{E}\left[ \|\mathcal{C}(x) - x\|^2\right] = (1-\alpha) \|x\|^2.$$
	
	The contraction coefficient $1-\alpha$ for clients during the execution of \algname{FedNL} remains constant and the mathematical framework of \algname{FedNL} can be seamlessly applied without any modifications. A notable advantage of \algname{TopLEK[k]} over \compname{TopK[k]} is that clients can transmit not only $k$ components but at most $k$. In fortuitous scenarios, clients may only need to send $0$ components.
	
	\subsection{TopLEK Applied for Matrices}
	
	The \algname{FedNL} operates on a family of contractive compressors applied for matrices which is defined as:
	\begin{eqnarray*}
		\mathcal{C}(\delta)=\{\mathcal{C} \in \mathbb{R}^{d \times d} \to \mathbb{R}^{d \times d} : &(i)& \mathbb{E}\left[ \|\mathcal{C}(x) - x\|_F ^2\right] \le (1-\delta)\|x\|_F^2, \forall x \in \mathbb{R}^{d \times d}, \delta \in (0,1], \\ &(ii)& \|\mathcal{C}(M)\|_F \le \|M\|_F\}.
	\end{eqnarray*}
	
	Requirements (i) can be seen as a generalization of the class of contractive operators for vectors, defined in Eq.\ref{def:biased_compressors} to matrices. Because both \compname{TopK} and \compname{TopLEK} only replace some matrix elements by $0$ value, and do not perform any scaling, both compressors satisfy the requirement (ii) automatically.
	
	\clearpage
	\section{Missing Plots}
	
	\subsection{Missing Plots for Single-node Comparison FedNL-LS with CVXPY}
	\label{app:single-node-cmp-vs-industry-extra}
	
	\begin{figure}[h]
		\centering
		\includegraphics[width=0.325\textwidth]{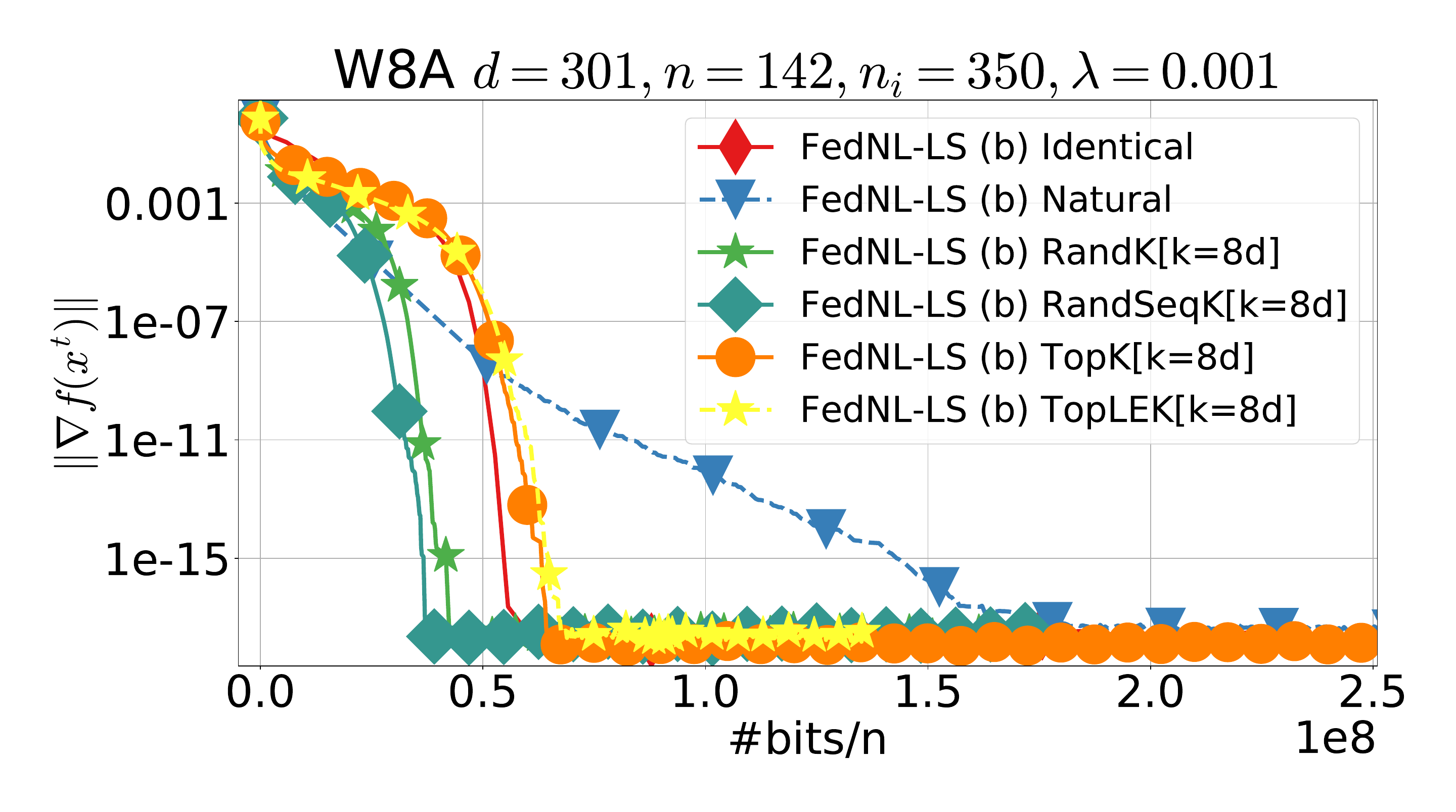}
		\includegraphics[width=0.325\textwidth]{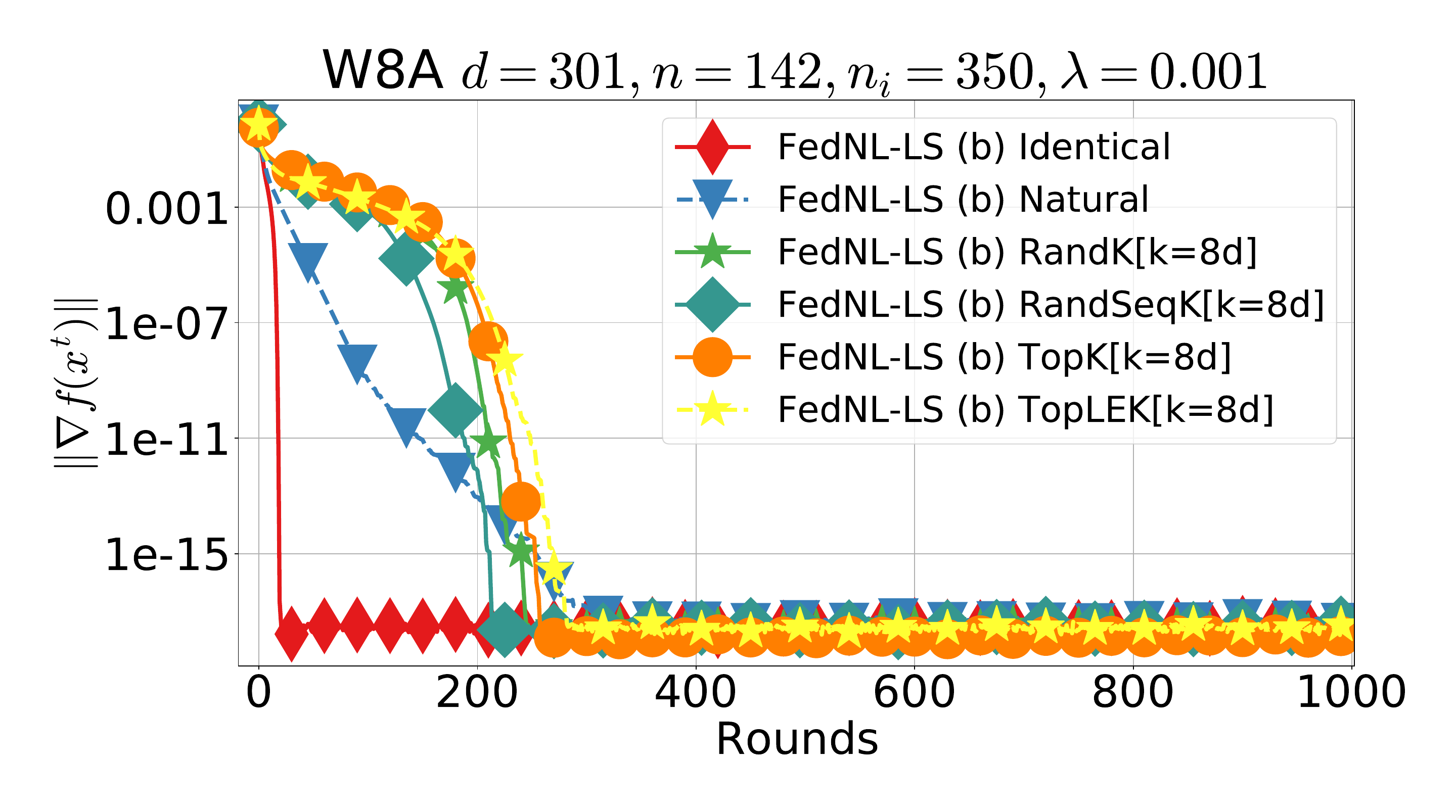}
		\includegraphics[width=0.325\textwidth]{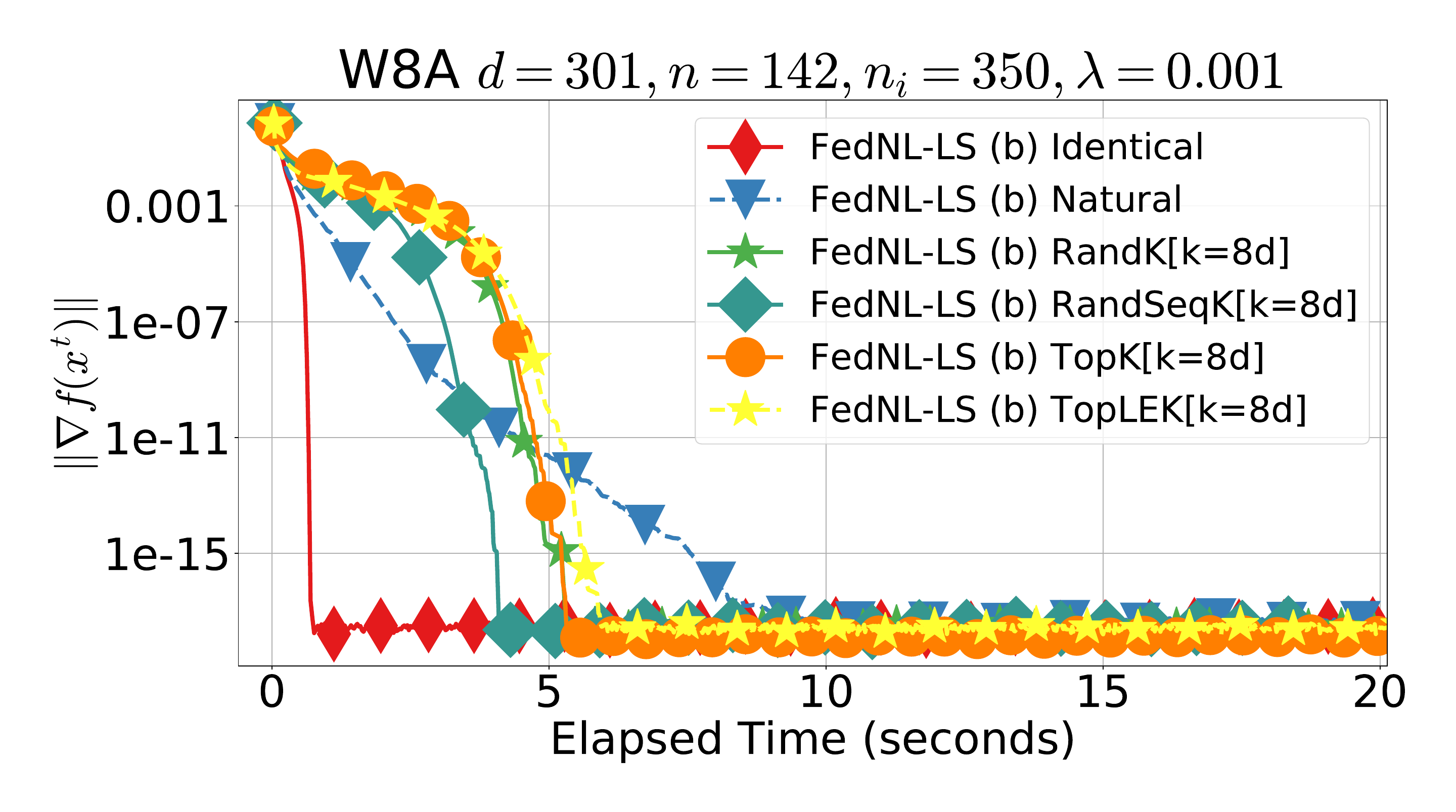}		
		\caption{\algname{FedNL-LS} simulation in a single-node, $1000$ rounds, theoretical step-size, FP64. Line search parameters $c=0.49,\gamma=0.5$. Dataset \dataname{W8A} ($49749$ samples) augmented with intercept	split to $n_i=350$ samples/client.}
		\label{fig:fednl-ls-w8a}
	\end{figure}
	
	\begin{figure}[h]
		\centering
		\includegraphics[width=0.325\textwidth]{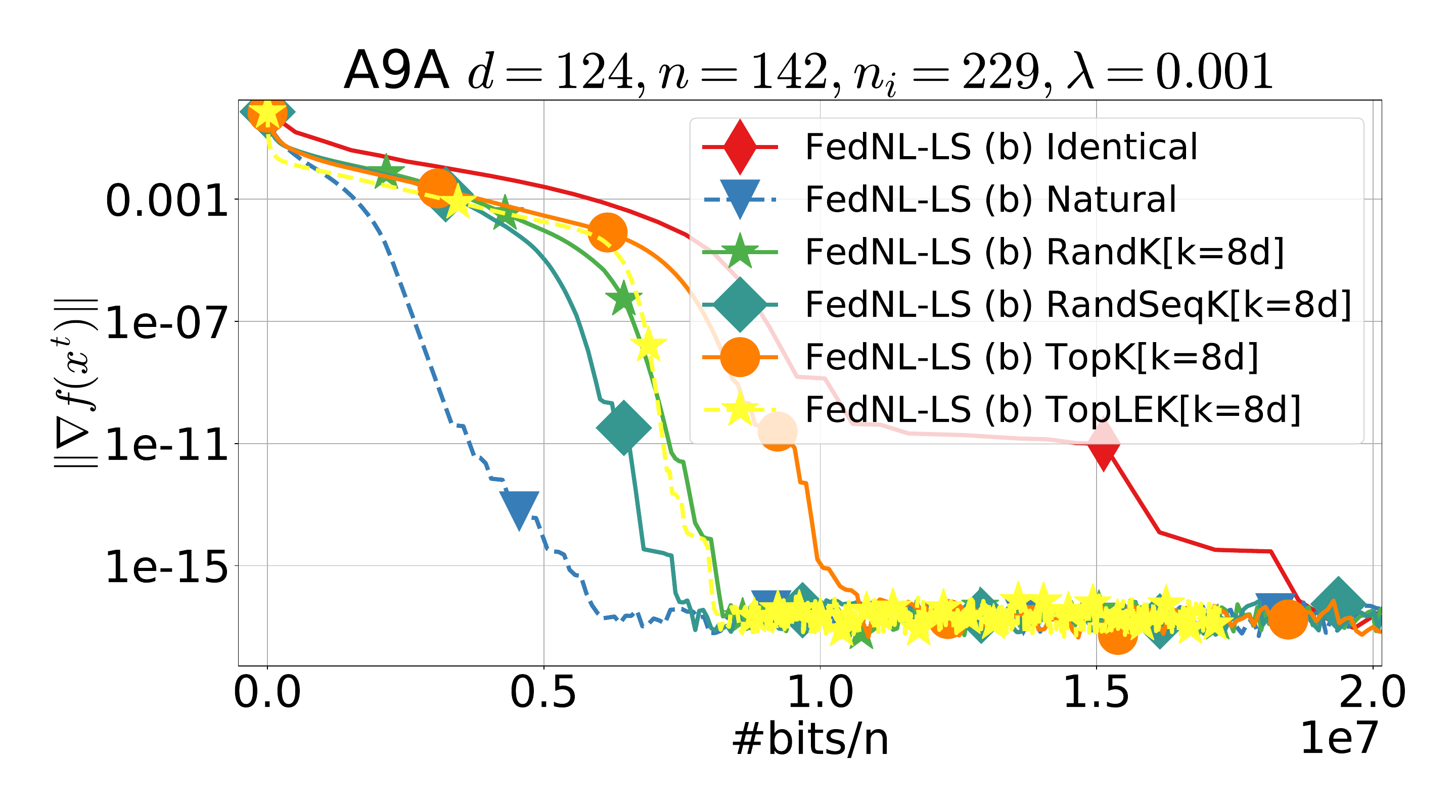}
		\includegraphics[width=0.325\textwidth]{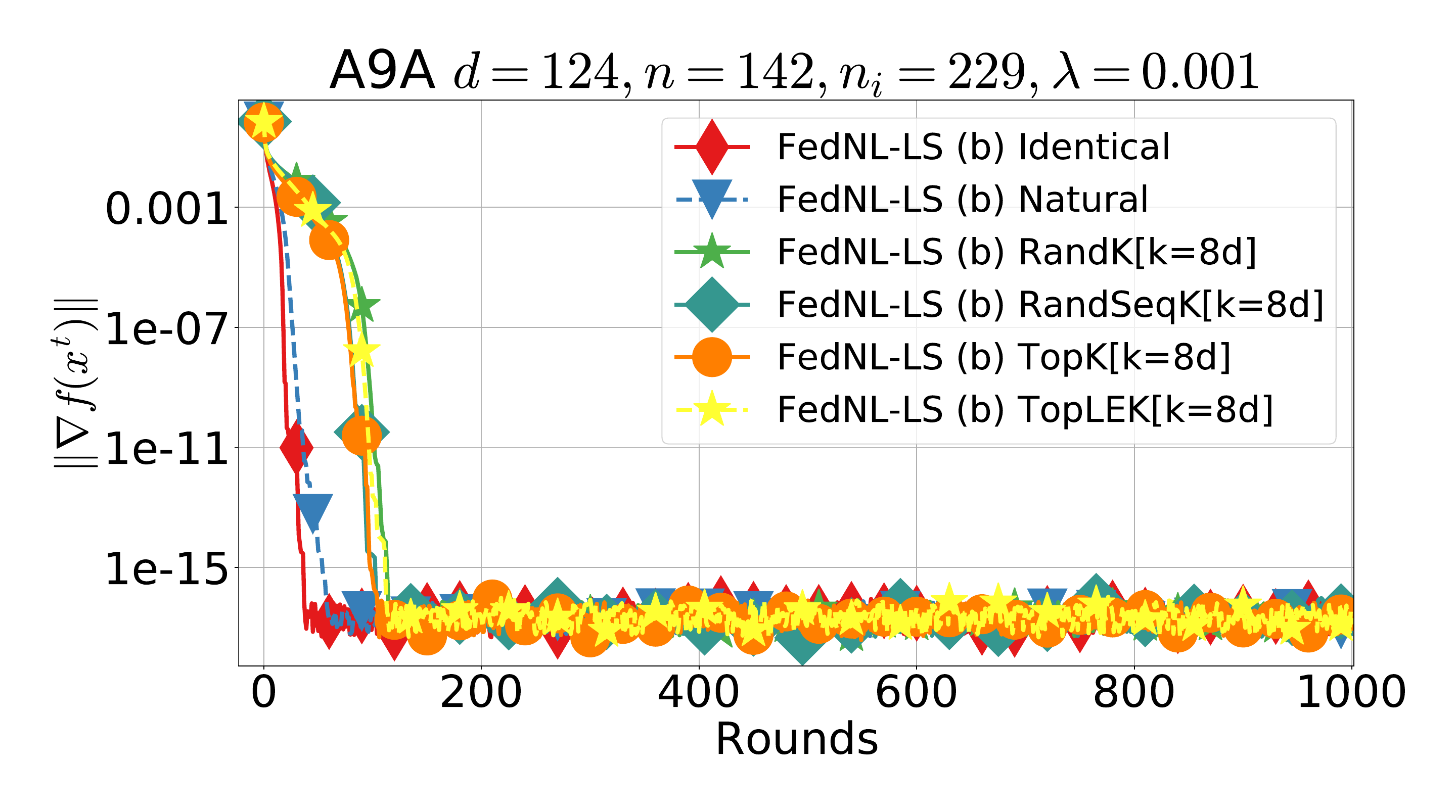}
		\includegraphics[width=0.325\textwidth]{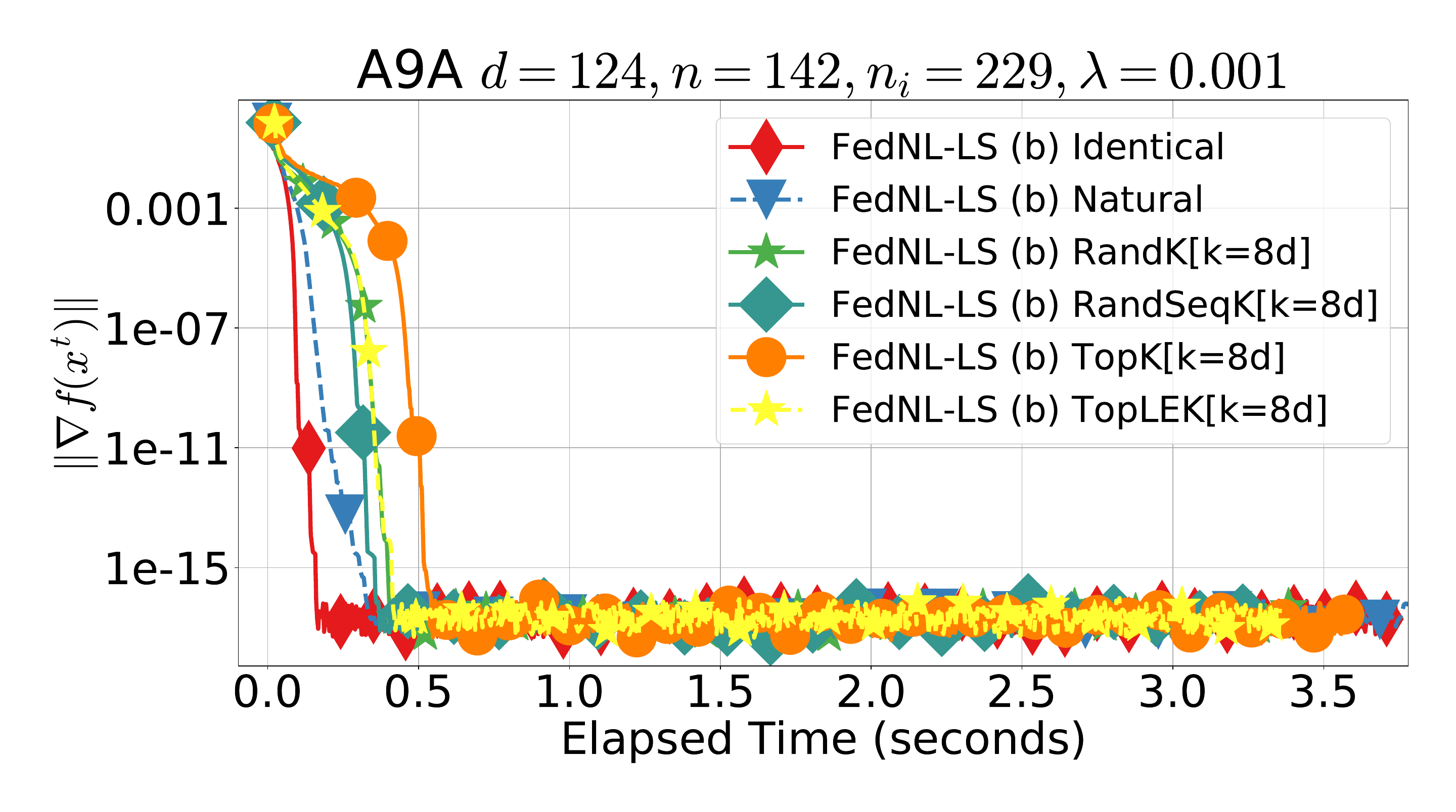}
		
		\caption{\algname{FedNL-LS} simulation in a single-node, $1000$ rounds, theoretical step-size, FP64. Line search parameters $c=0.49,\gamma=0.5$. Dataset \dataname{A9A} ($32561$ samples) augmented with intercept	split to $n_i=229$ samples/client.}
		
		\label{fig:fednl-ls-a9a}
	\end{figure}
	
	\begin{figure}[h]
		\centering
		\includegraphics[width=0.325\textwidth]{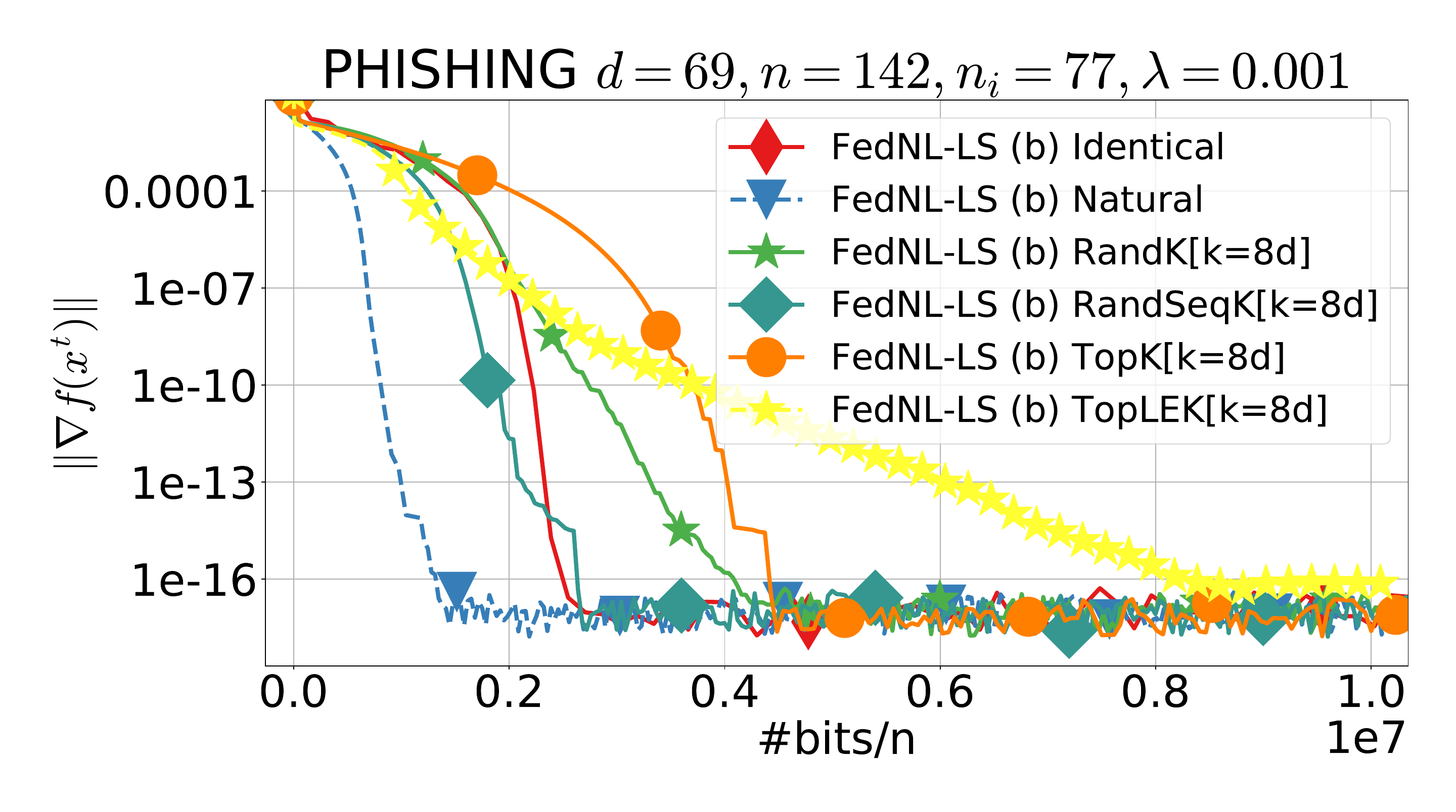}
		\includegraphics[width=0.325\textwidth]{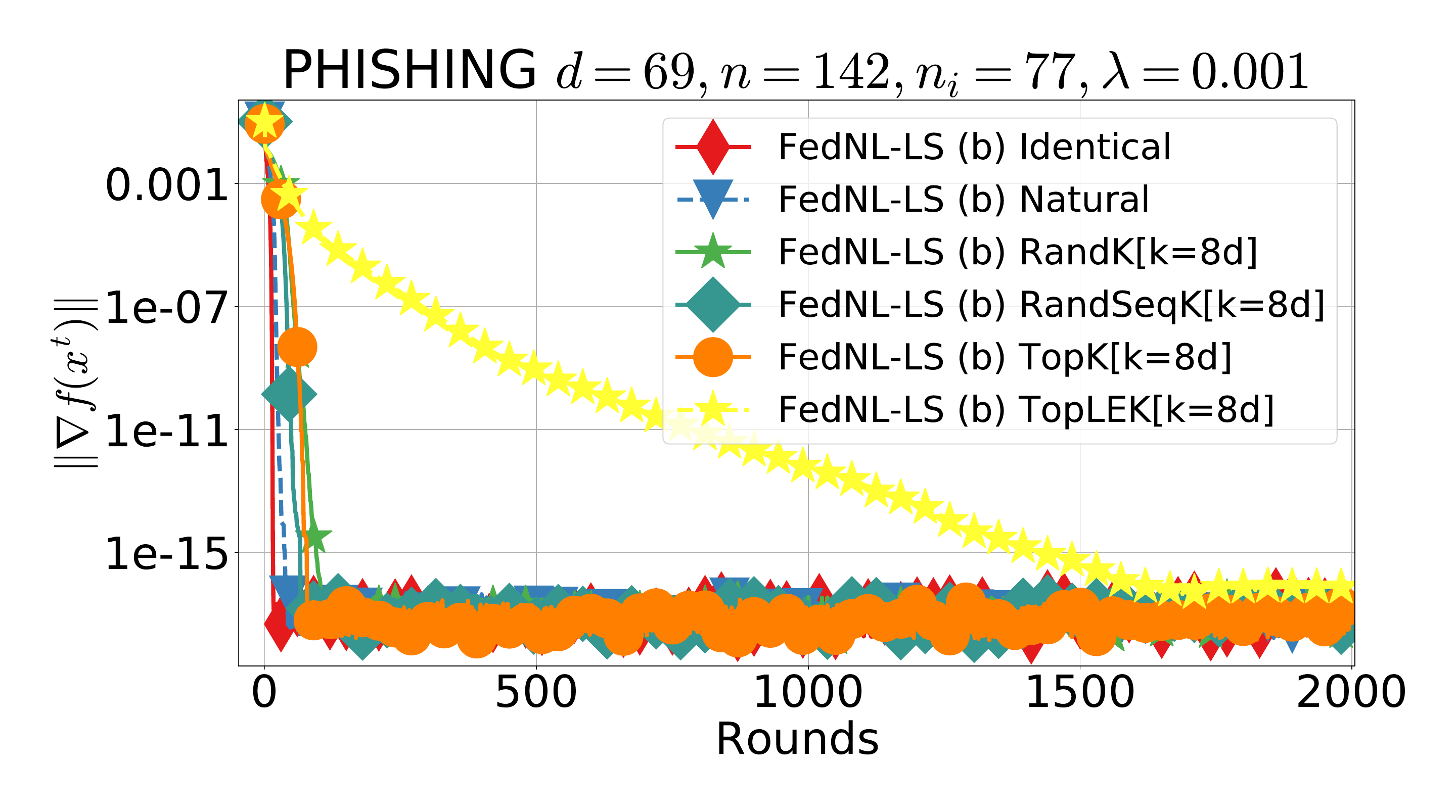}
		\includegraphics[width=0.325\textwidth]{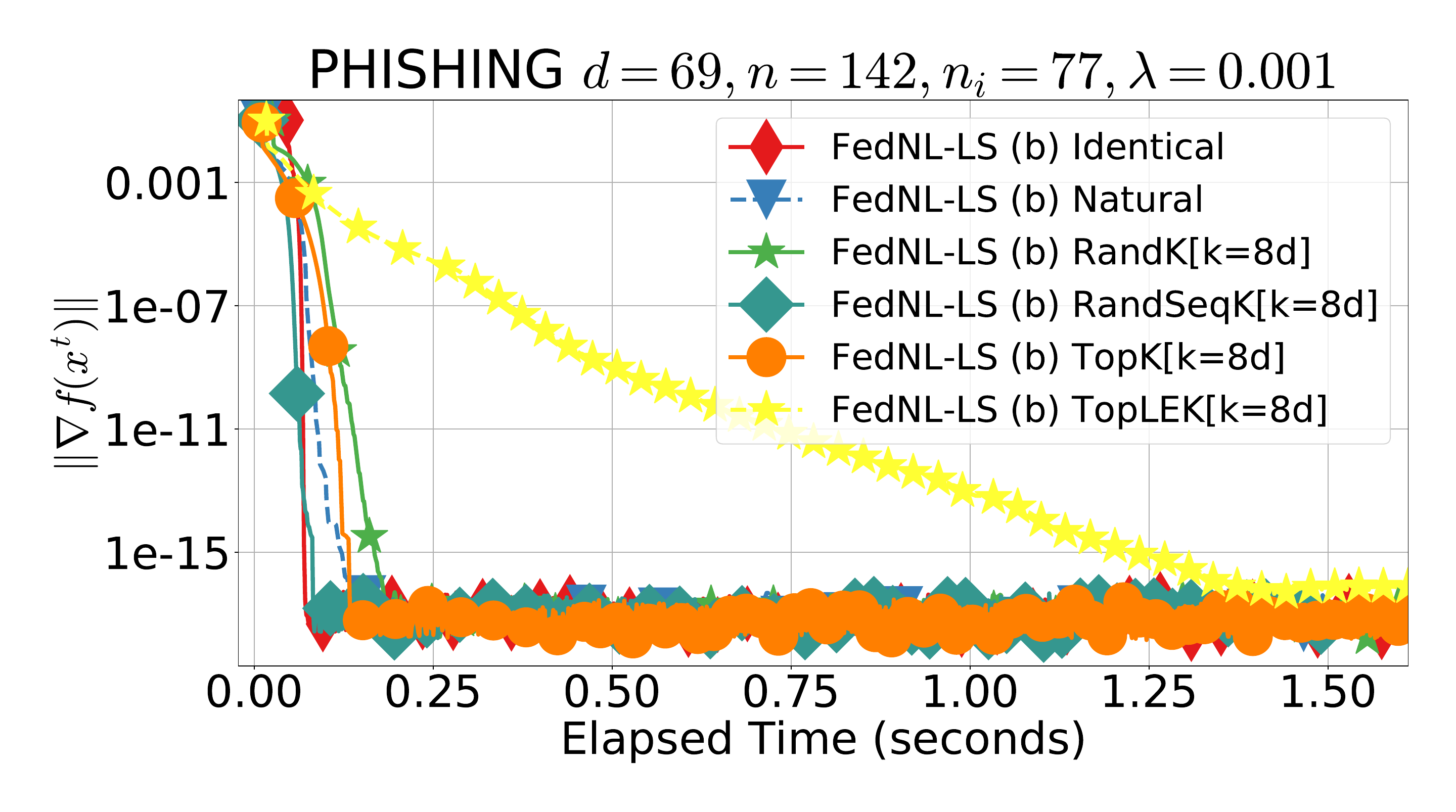}
		
		\caption{\algname{FedNL-LS} simulation in a single-node, $2000$ rounds, theoretical step-size, FP64. Line search parameters $c=0.49,\gamma=0.5$. Dataset \dataname{PHISHING} ( $11055$ samples) augmented with intercept	split to $n_i=77$ samples/client.}
		\label{fig:fednl-ls-phishing}
	\end{figure}
	
	The results of an experiment using \algname{FedNL-LS} which represent a modification of \algname{FedNL} are presented in Figures \ref{fig:fednl-ls-w8a}, \ref{fig:fednl-ls-a9a}, \ref{fig:fednl-ls-phishing}. For a pseudocode of \algname{FedNL-LS} see  Appendix~\ref{app:fednl-ls-descr}. The \textit{communicated bits} include the cost for transferring the scalar values and auxiliary information from the compressor.
	
	For \compname{TopK} and \compname{TopLEK} auxiliary information includes $32$ bits per single index transfer. For \compname{TopLEK} communicated bits include an extra $32$ bits value to encode the number of communicated components. For \compname{RandK} and \compname{RandSeqK} we support two modes: (i) indicies are transferred explicitly; (ii) the master reconstructs them via using known seeds for \abr{PRG}. In all experiments (ii) strategy is used.
	

	\clearpage
	\subsection{Missing Plots for Multi-Node Experiments with FedNL, FedNL-PP, and FedNL-LS}
	
	\label{app:experiments-multi-node}
	
	
	\begin{figure}[h]
		\centering
		\includegraphics[width=0.325\textwidth]{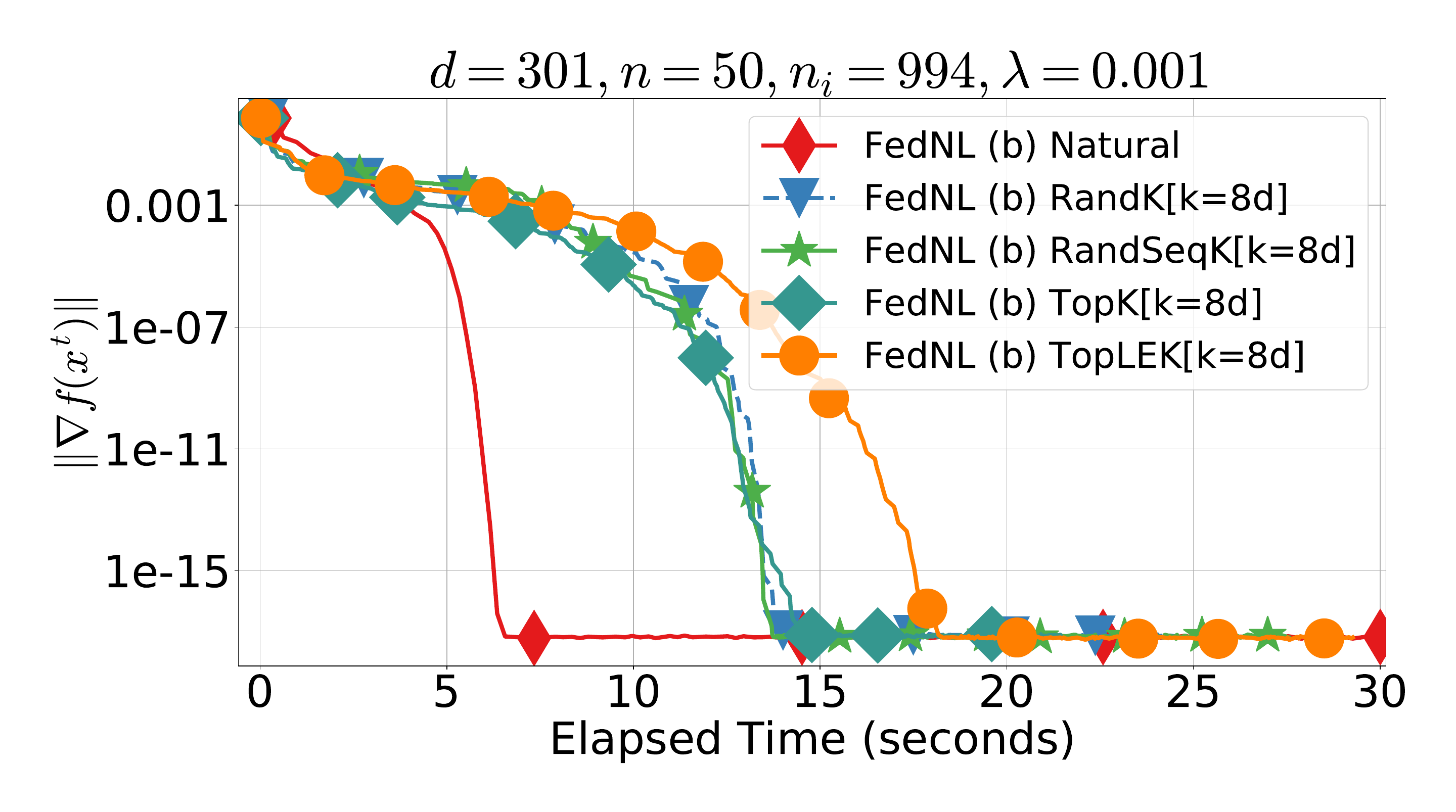}		
		\includegraphics[width=0.325\textwidth]{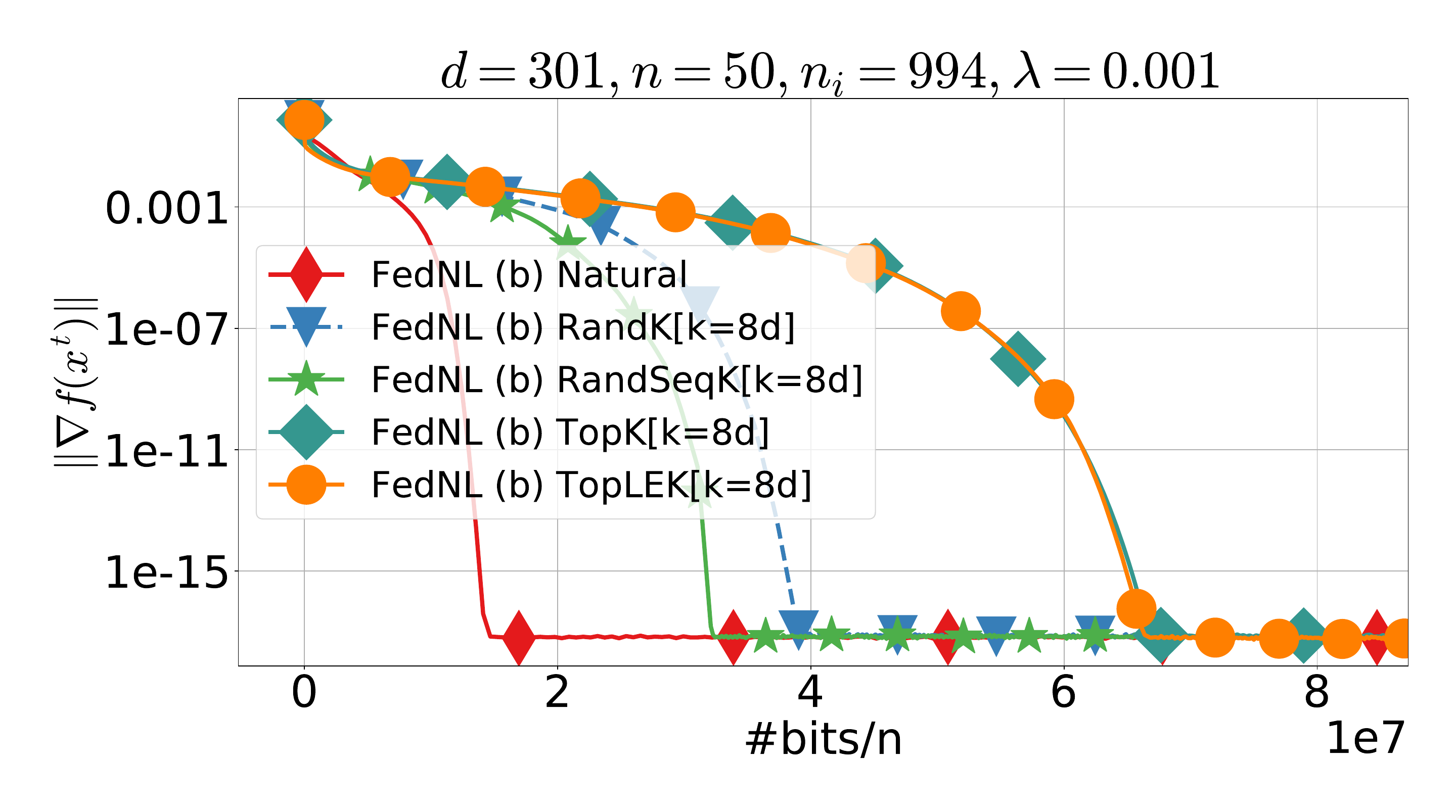}
		\includegraphics[width=0.325\textwidth]{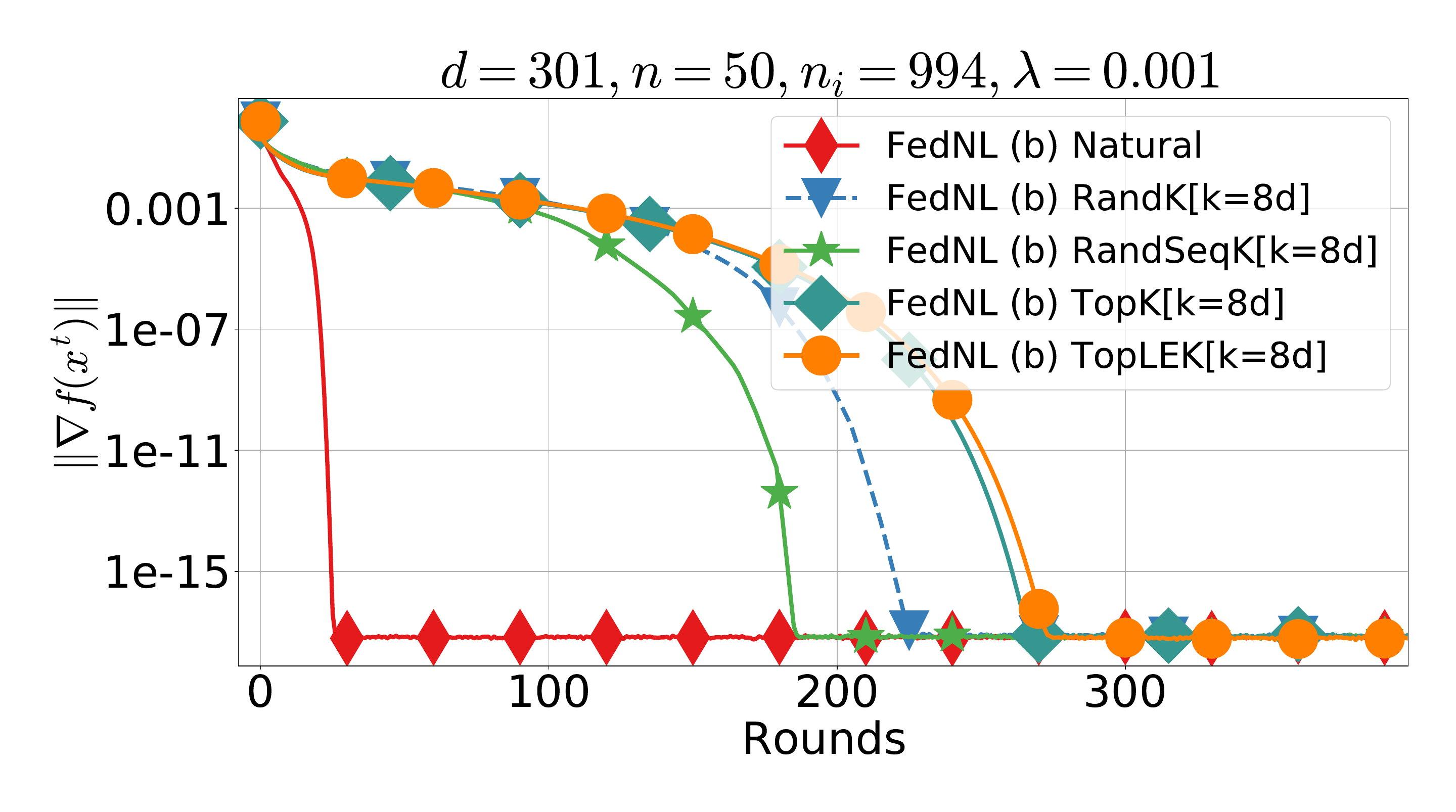}
		
		\caption{\algname{FedNL} in multi-node setting, theoretical step-size, $n=50$, FP64 arithmetic, 1 {CPU} core per node and master, TCP/IPv4, dataset \dataname{W8A} reshuffled u.a.r. and augmented with  intercept.}
		\label{fig:fednl-w8a-app}
	\end{figure}
	
	\begin{figure}[h]
		\centering
		\includegraphics[width=0.325\textwidth]{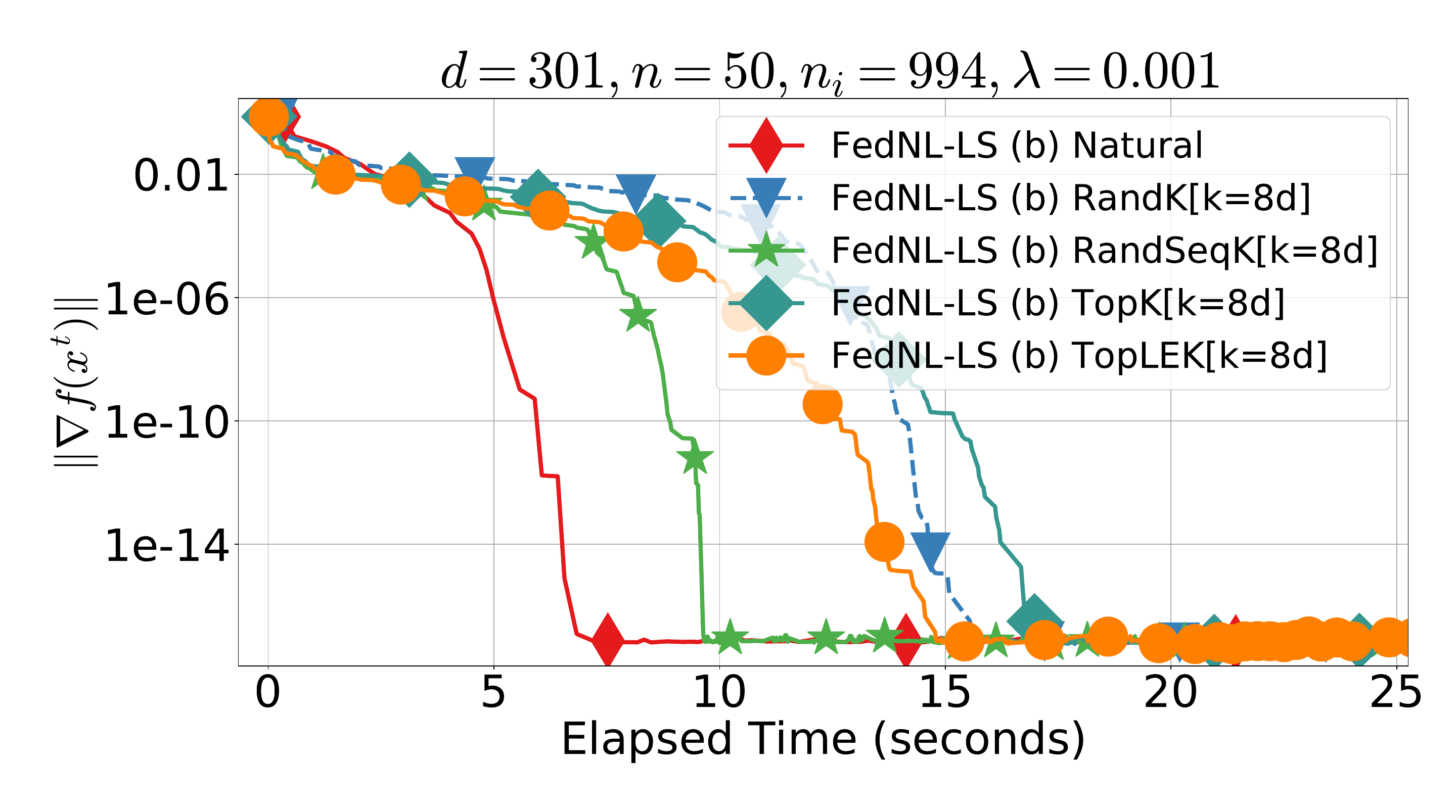}
		\includegraphics[width=0.325\textwidth]{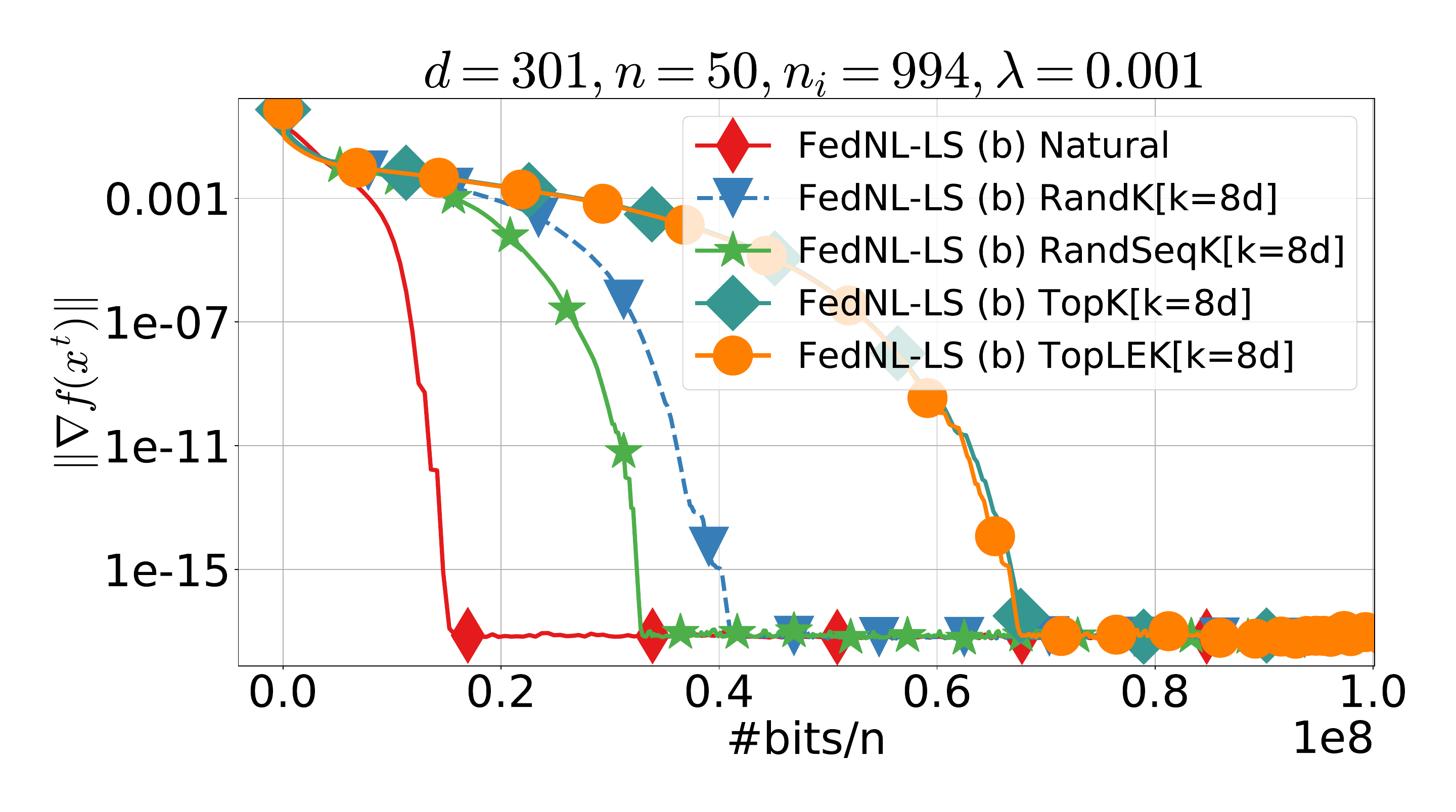}
		\includegraphics[width=0.325\textwidth]{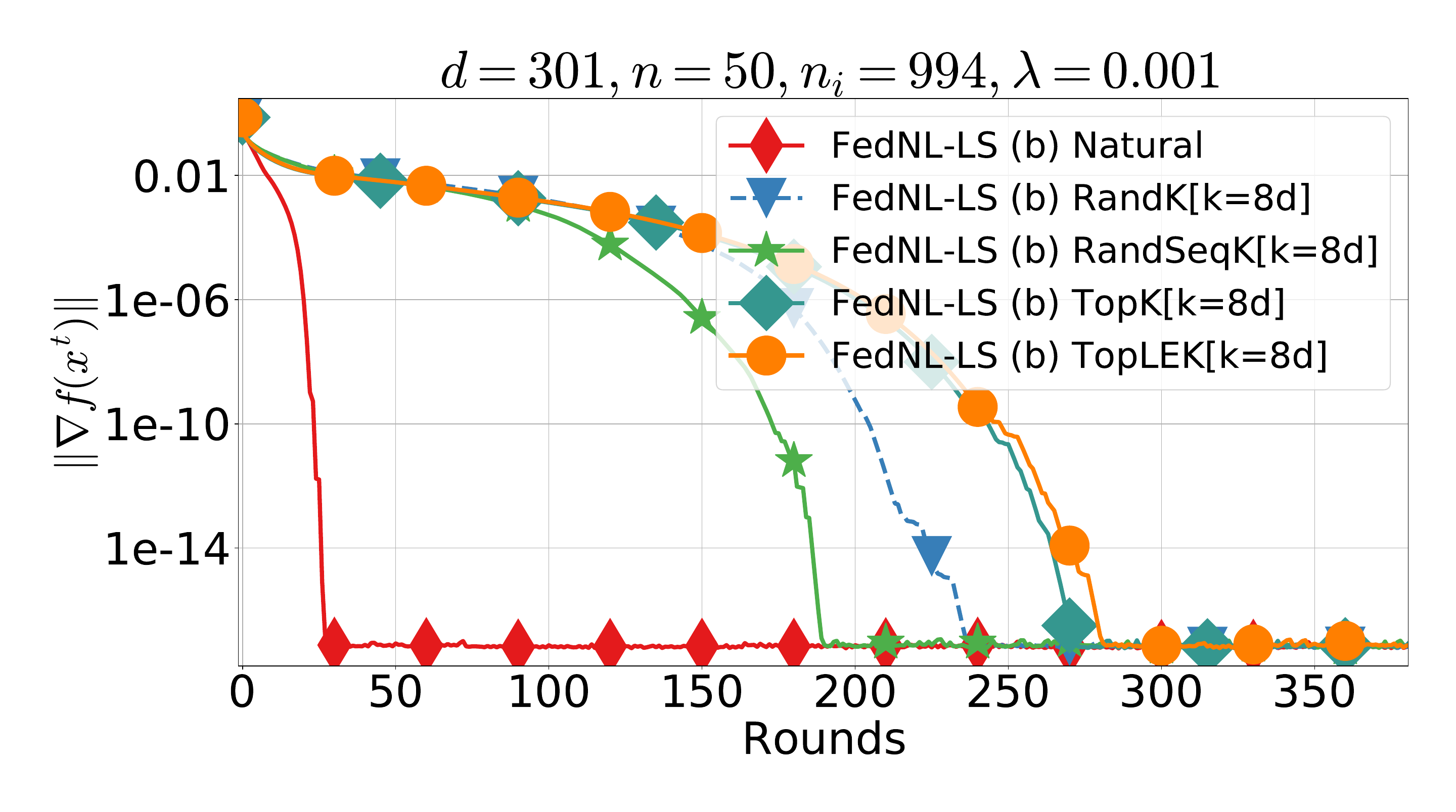}
		
		\caption{ \algname{FedNL-LS} in multi-node setting, $n=50$, FP64 arithmetic, 1 {CPU} core per node and master, TCP/IPv4, dataset \dataname{W8A} reshuffled u.a.r. and augmented with intercept. The line search parameters $c=0.49,\gamma=0.5$.}
		\label{fig:fednl-ls-w8a-app}
	\end{figure}
	
	\begin{figure}[h]
		\centering
		
		\includegraphics[width=0.325\textwidth]{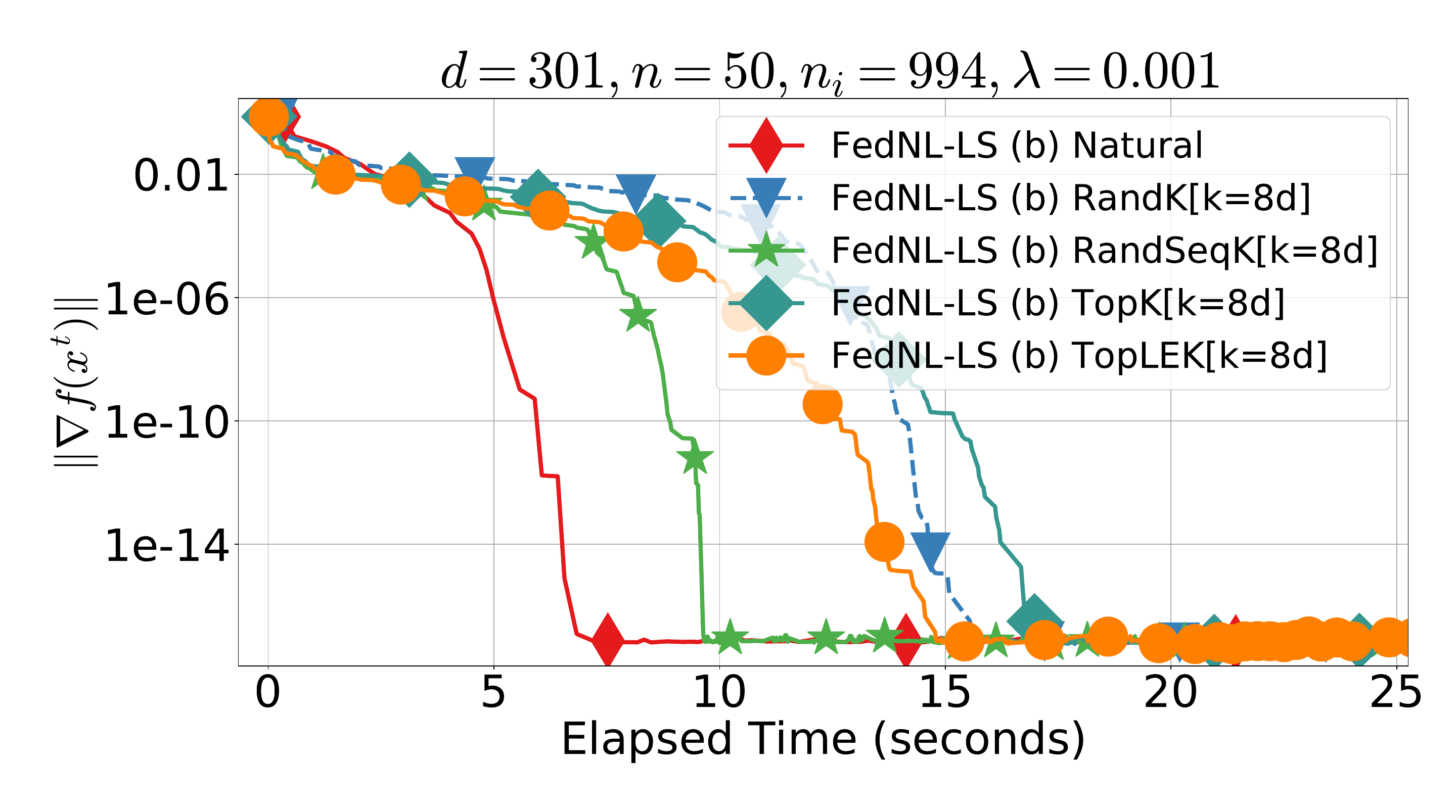}
		\includegraphics[width=0.325\textwidth]{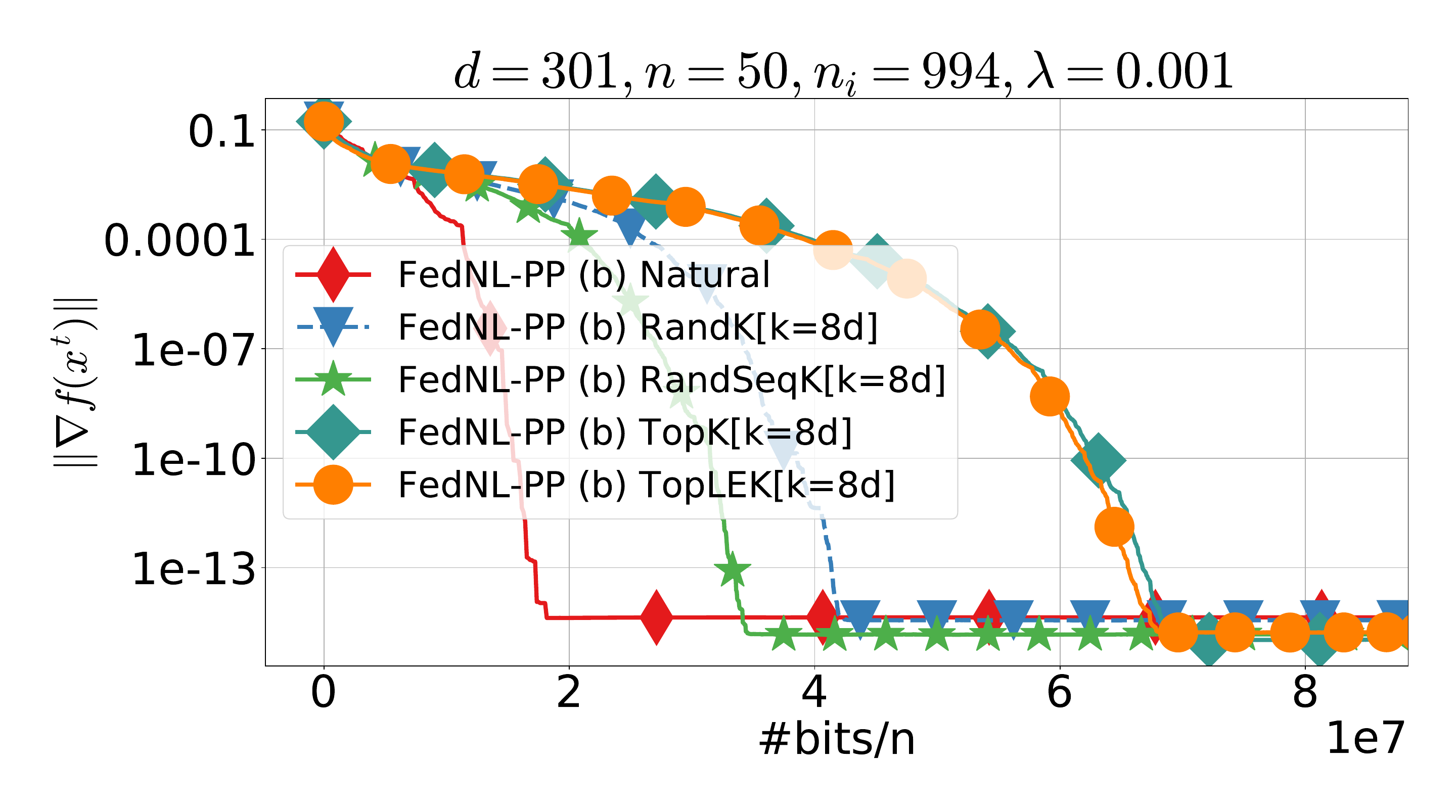}
		\includegraphics[width=0.325\textwidth]{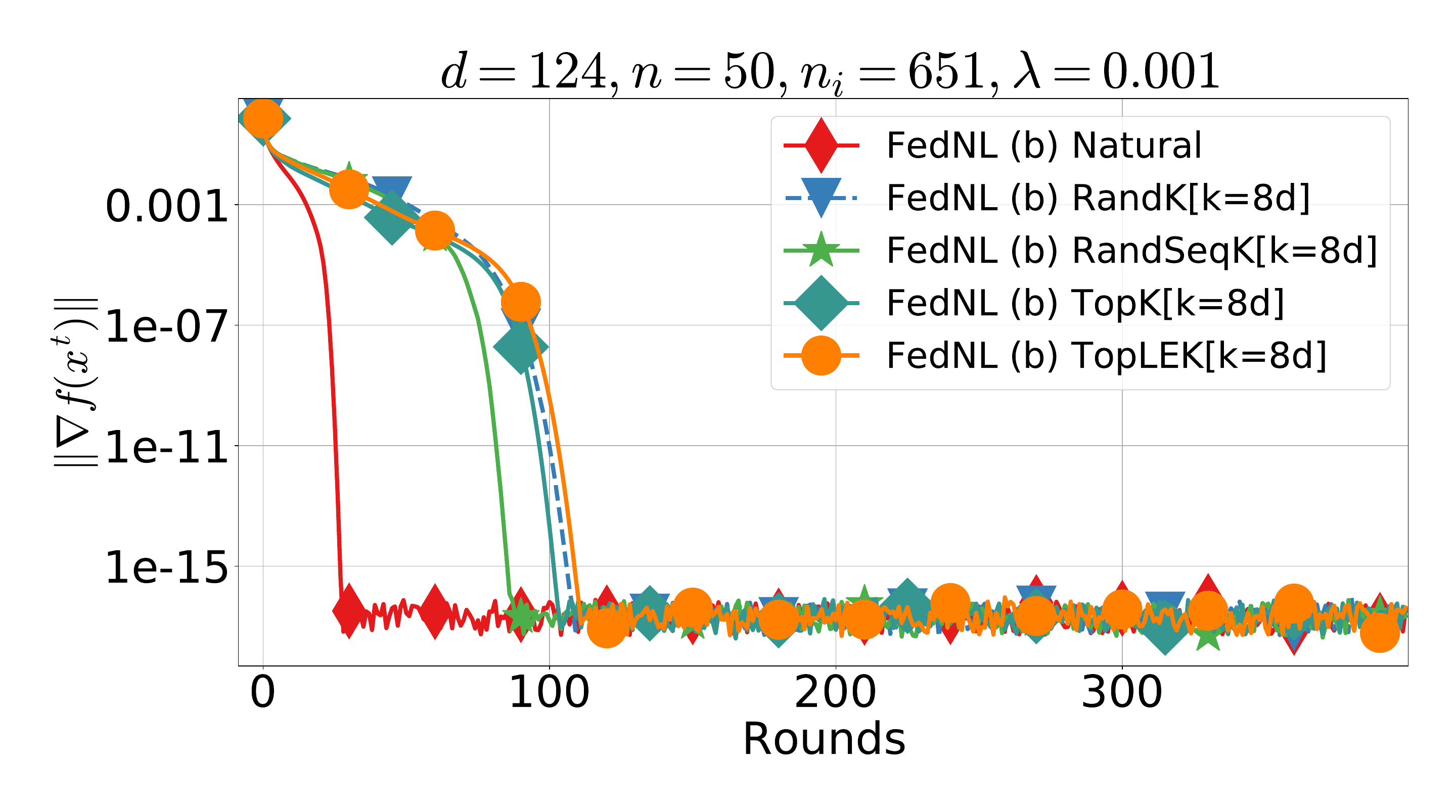}	
		
		\caption{ \algname{FedNL-PP} in  multi-node setting, $n=50$, $|S^k|=12$ clients per round, FP64 arithmetic, 1 {CPU} core per node and master, TCP/IPv4. \dataname{W8A} dataset reshuffled u.a.r. and augmented with intercept.}
		\label{fig:fednl-pp-w8a-app}
	\end{figure}
	
	
	\begin{figure}[h]
		\centering
		\includegraphics[width=0.325\textwidth]{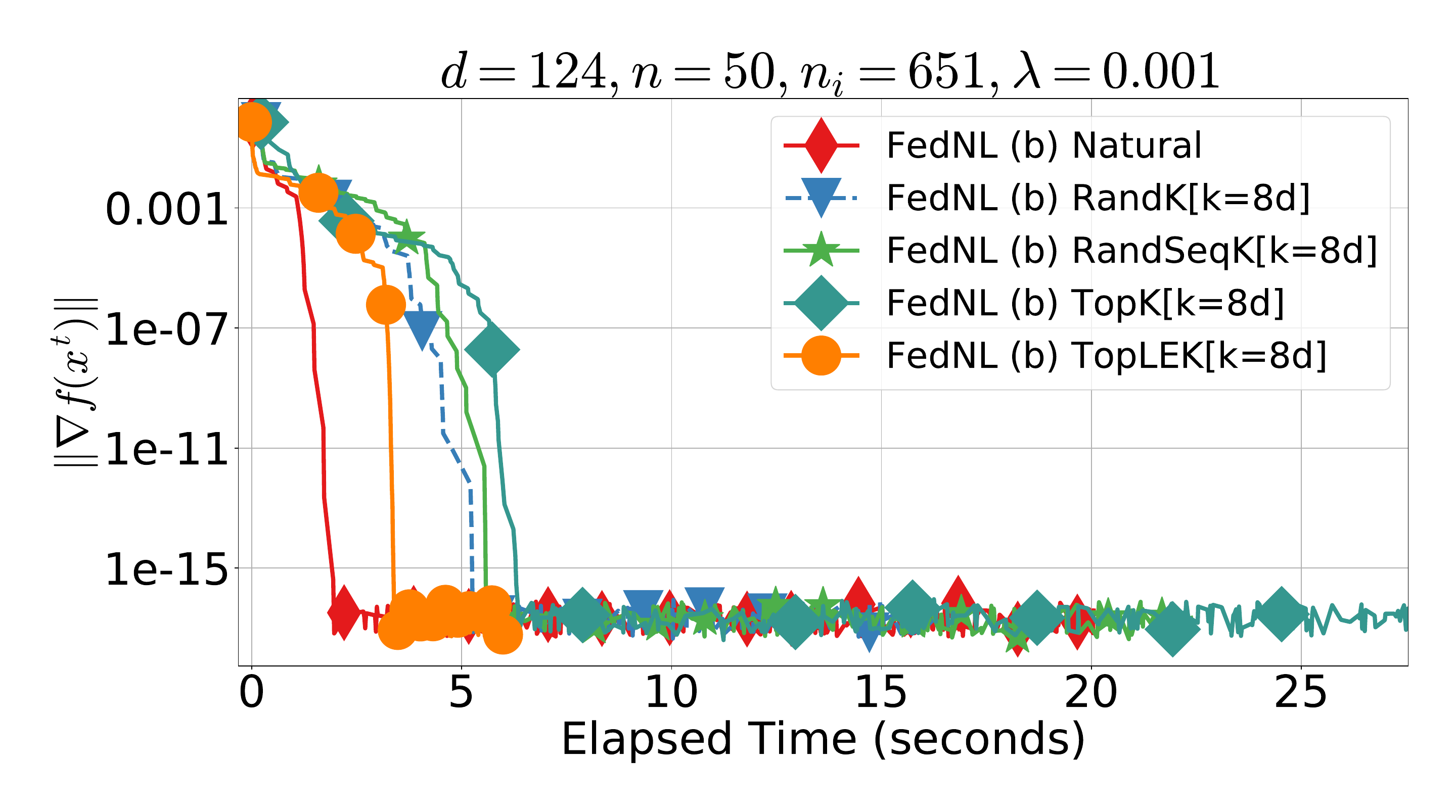}
		\includegraphics[width=0.325\textwidth]{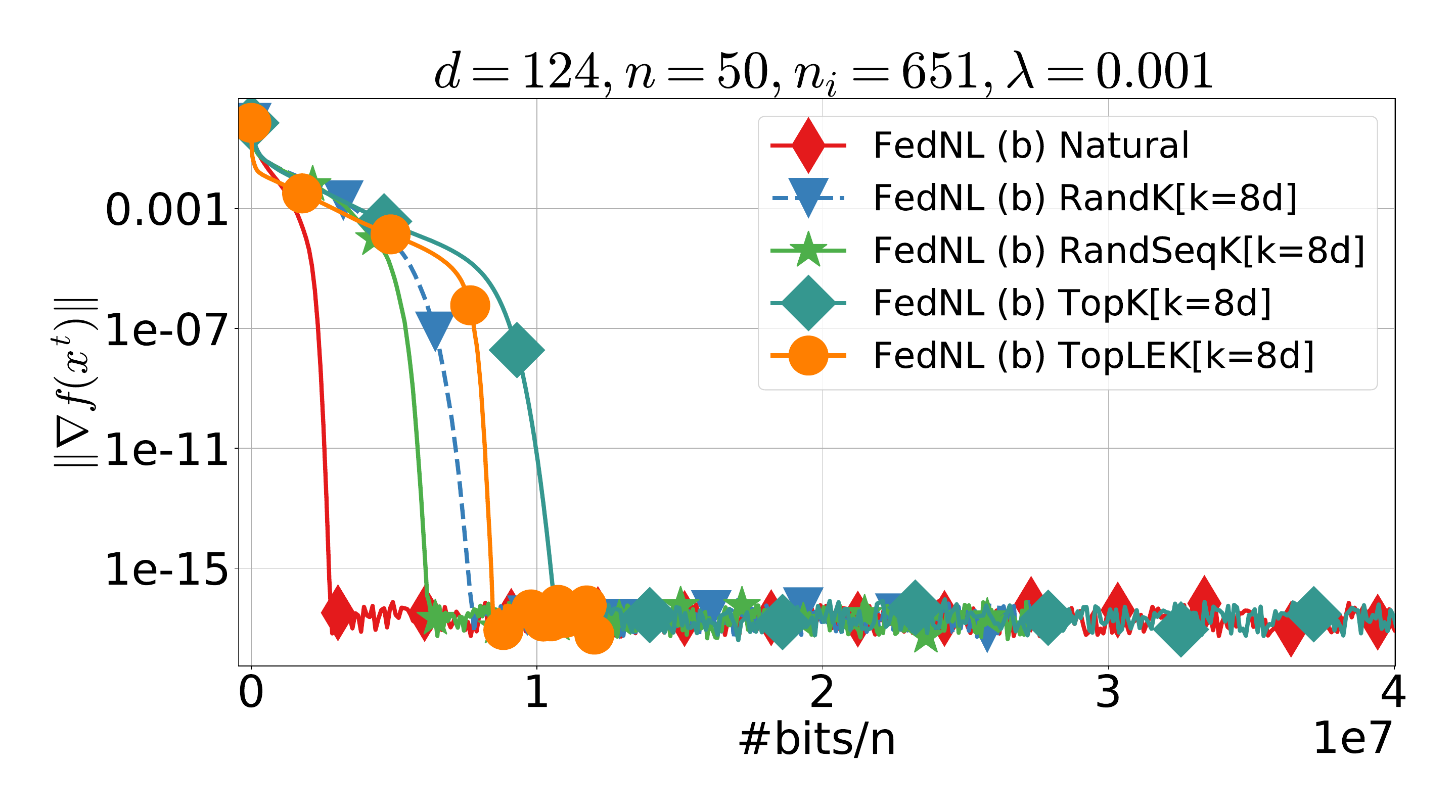}
		\includegraphics[width=0.325\textwidth]{./figs/fednl-appendix-figs/a9a/fednl/grad-vs-rounds.pdf}
		
		\caption{\algname{FedNL} in multi-node setting, theoretical step-size, $n=50$, FP64 arithmetic, 1 {CPU} core per node and master, TCP/IPv4, dataset \dataname{A9A} reshuffled u.a.r. and augmented with  intercept.}
				
		\label{fig:fednl-a9a-app}
	\end{figure}
	
	\begin{figure}[h]
		\centering
		\includegraphics[width=0.325\textwidth]{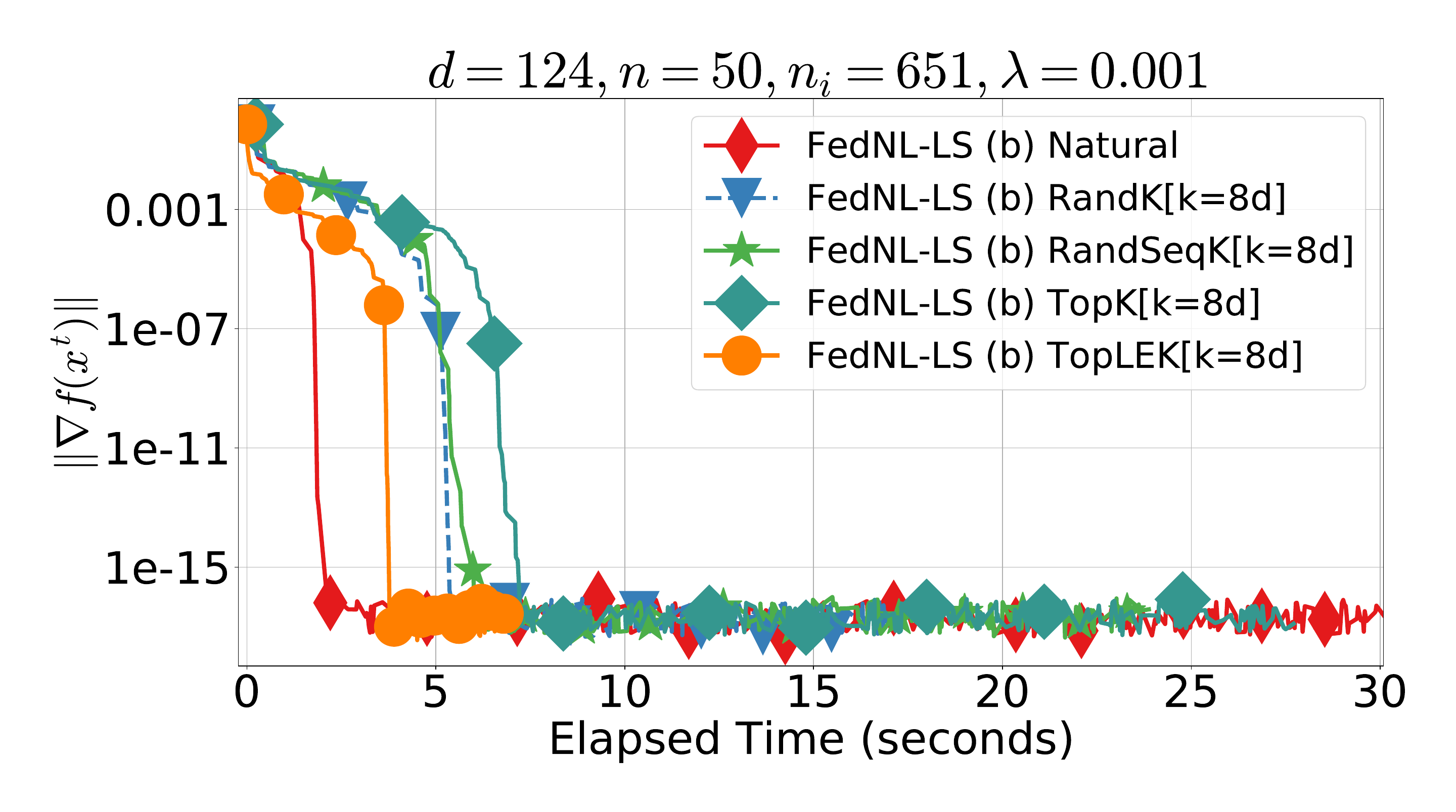}	
		\includegraphics[width=0.325\textwidth]{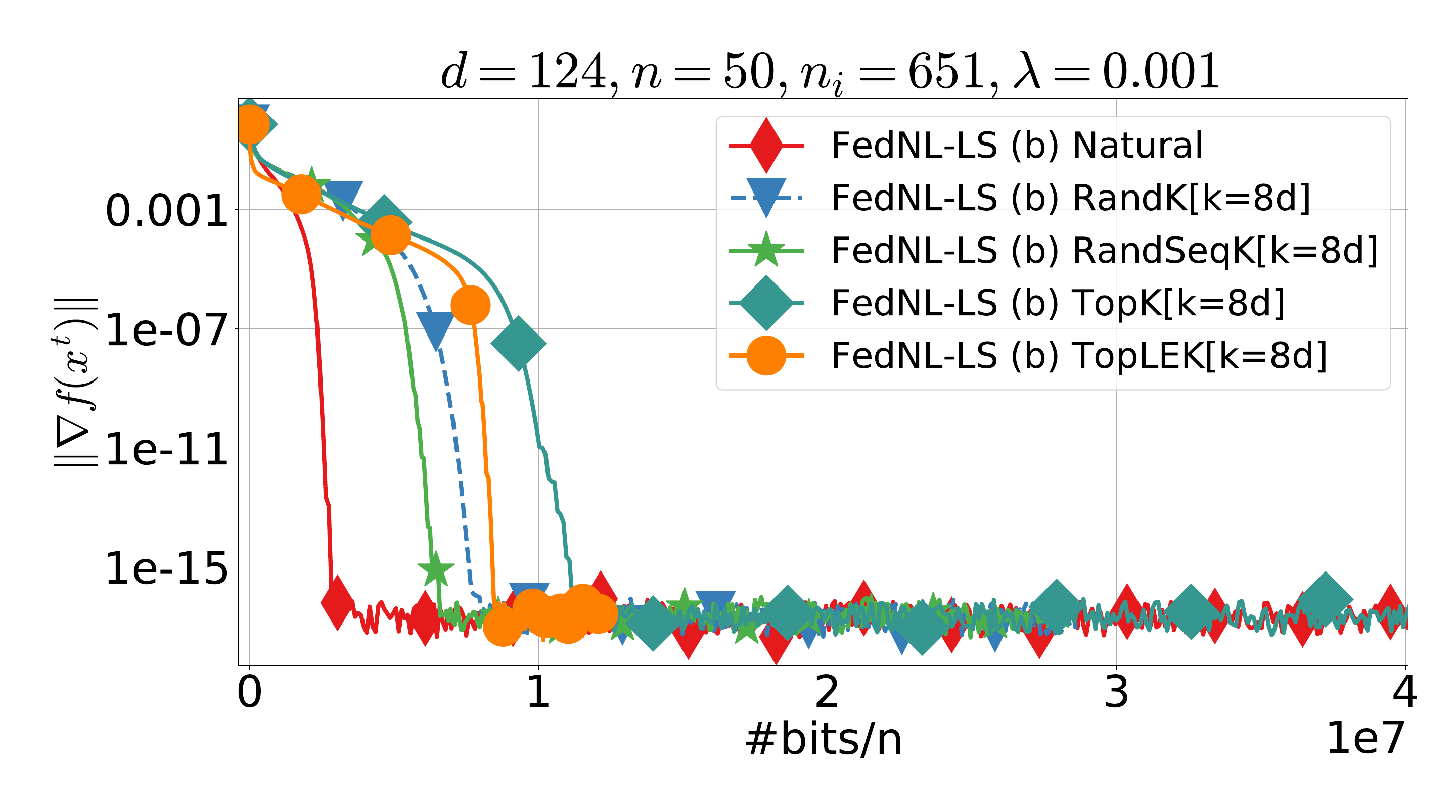}
		\includegraphics[width=0.325\textwidth]{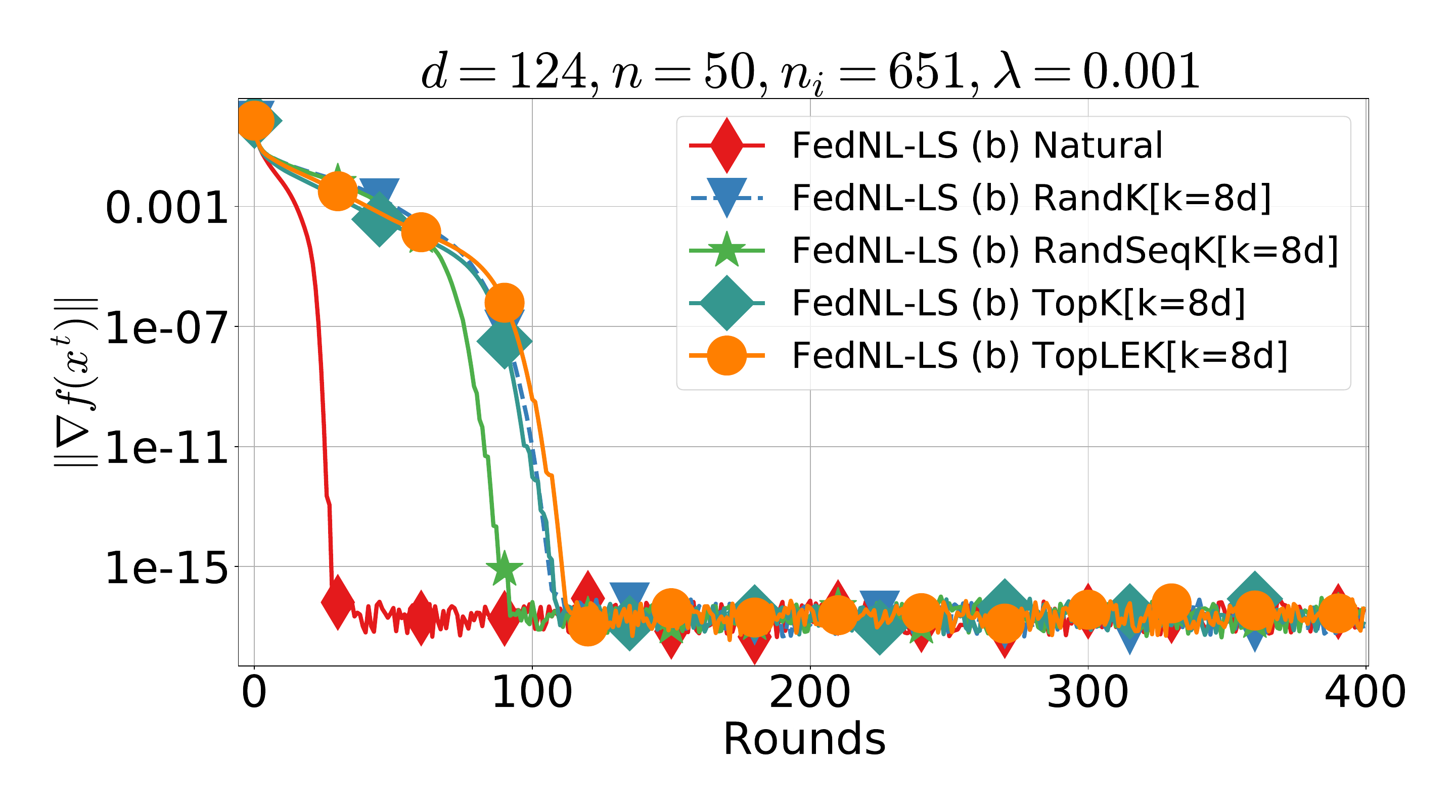}
		
		\caption{ \algname{FedNL-LS} in multi-node setting, $n=50$, FP64 arithmetic, 1 {CPU} core per node and master, TCP/IPv4, dataset \dataname{A9A} reshuffled u.a.r. and augmented with intercept. The line search parameters $c=0.49,\gamma=0.5$.}
		\label{fig:fednl-ls-a9a-app}
	\end{figure}
	
	\begin{figure}[h]
		\centering
		\includegraphics[width=0.325\textwidth]{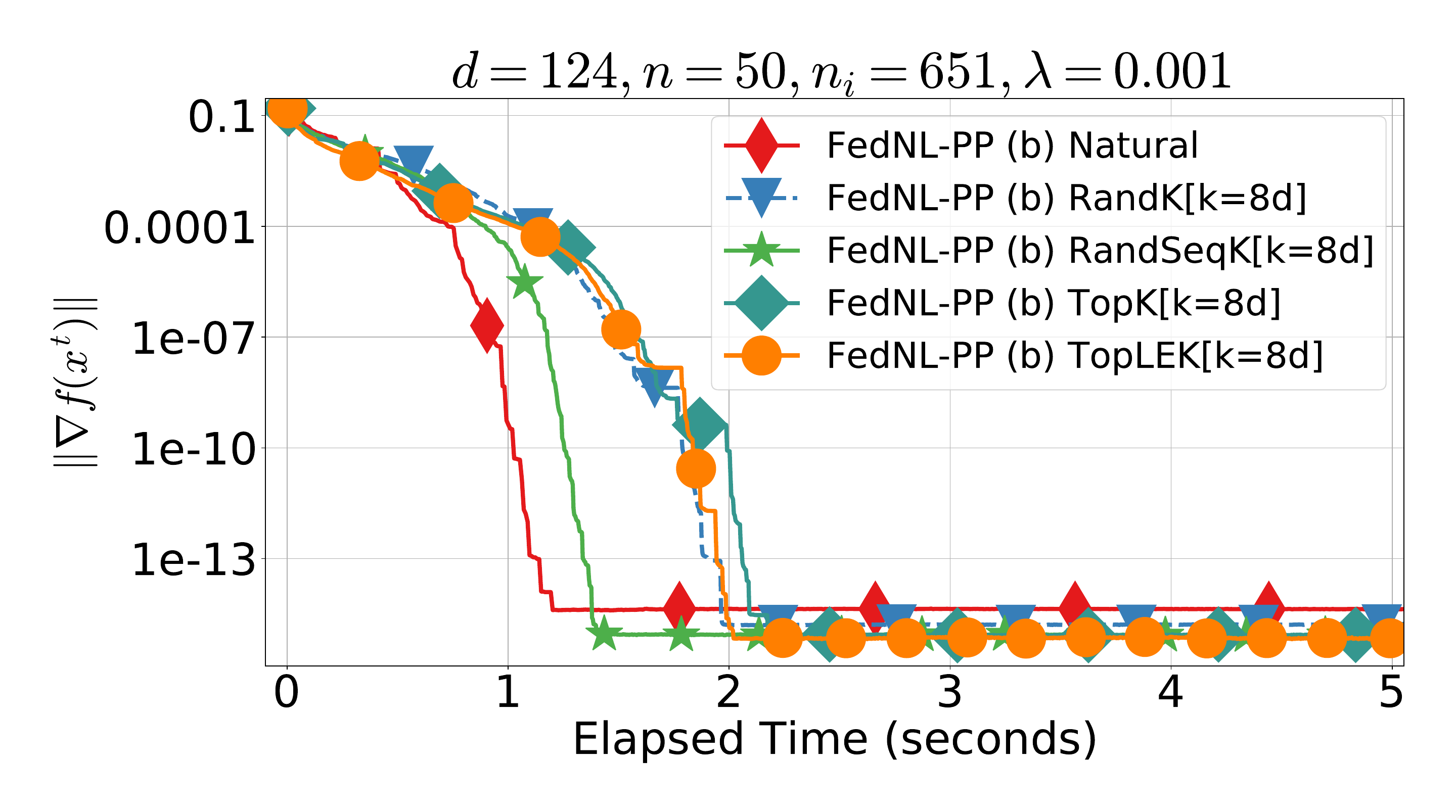}
		\includegraphics[width=0.325\textwidth]{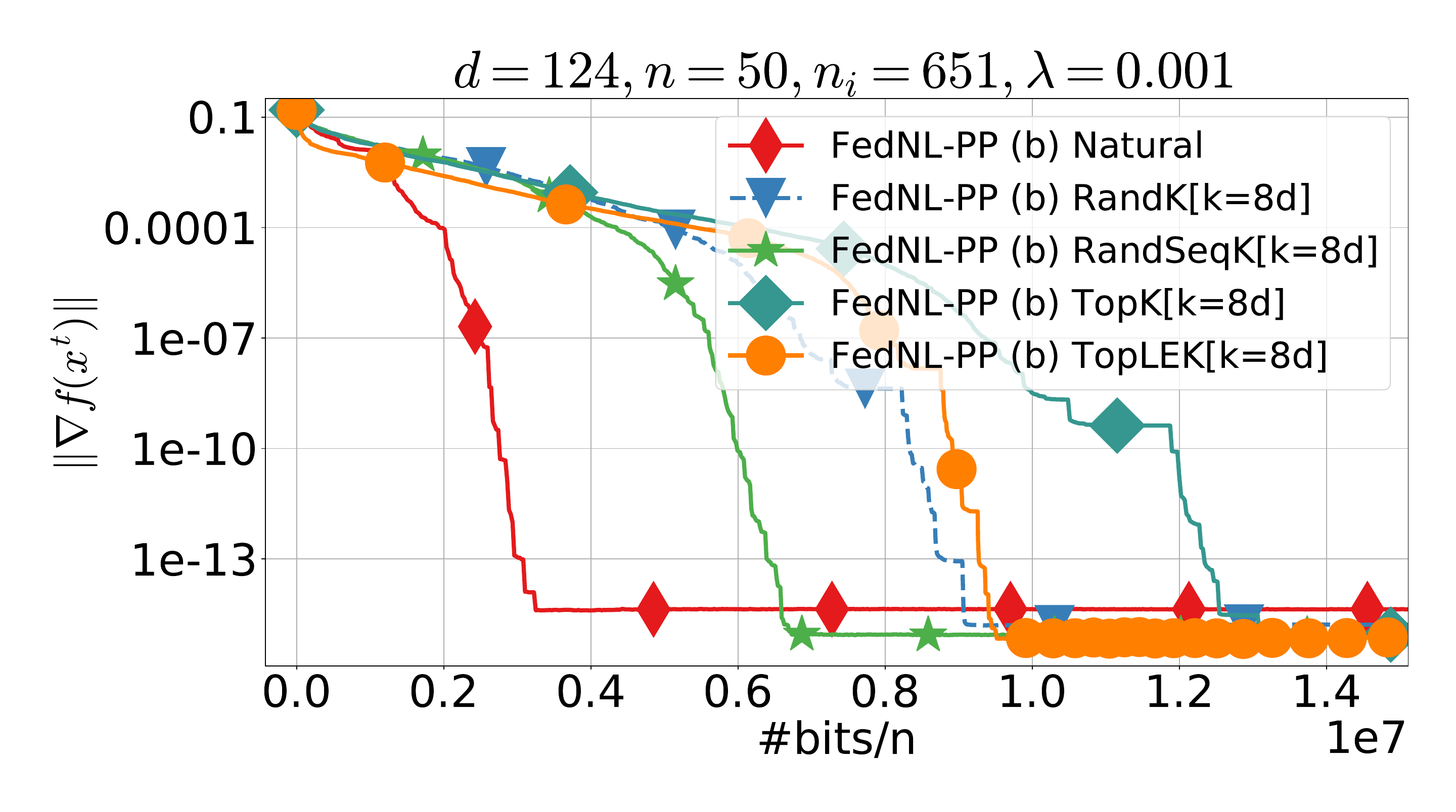}
		\includegraphics[width=0.325\textwidth]{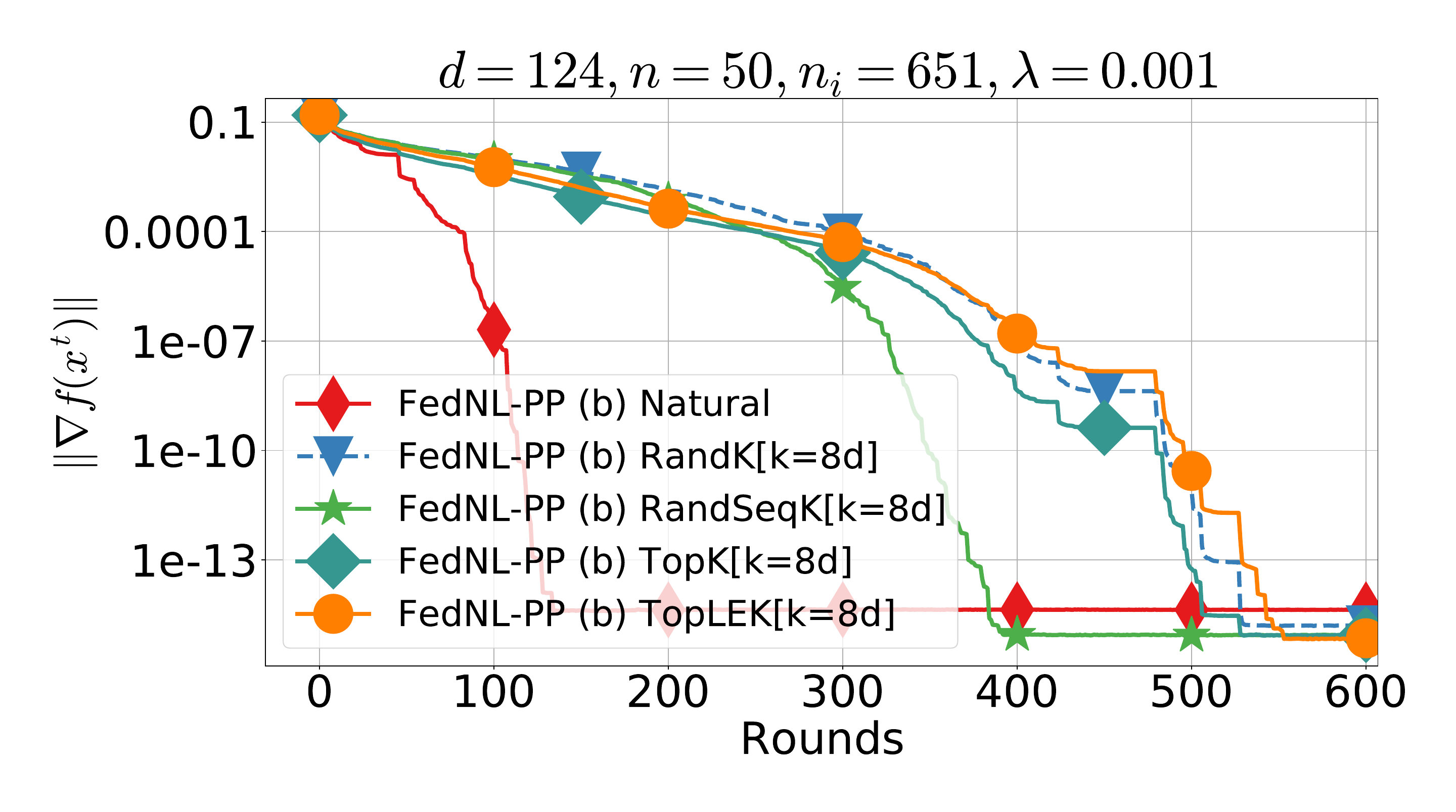}
		
		\caption{\algname{FedNL-PP} at \dataname{A9A} in  multi-node setting, $n=50$, $|S^k|=12$ clients per round, FP64 arithmetic, 1 {CPU} core per node and master, TCP/IPv4. \dataname{A9A} dataset reshuffled u.a.r. and augmented with intercept.}
		\label{fig:fednl-pp-a9a-app}
	\end{figure}

	
	\begin{figure}[h]
		\centering
		\includegraphics[width=0.325\textwidth]{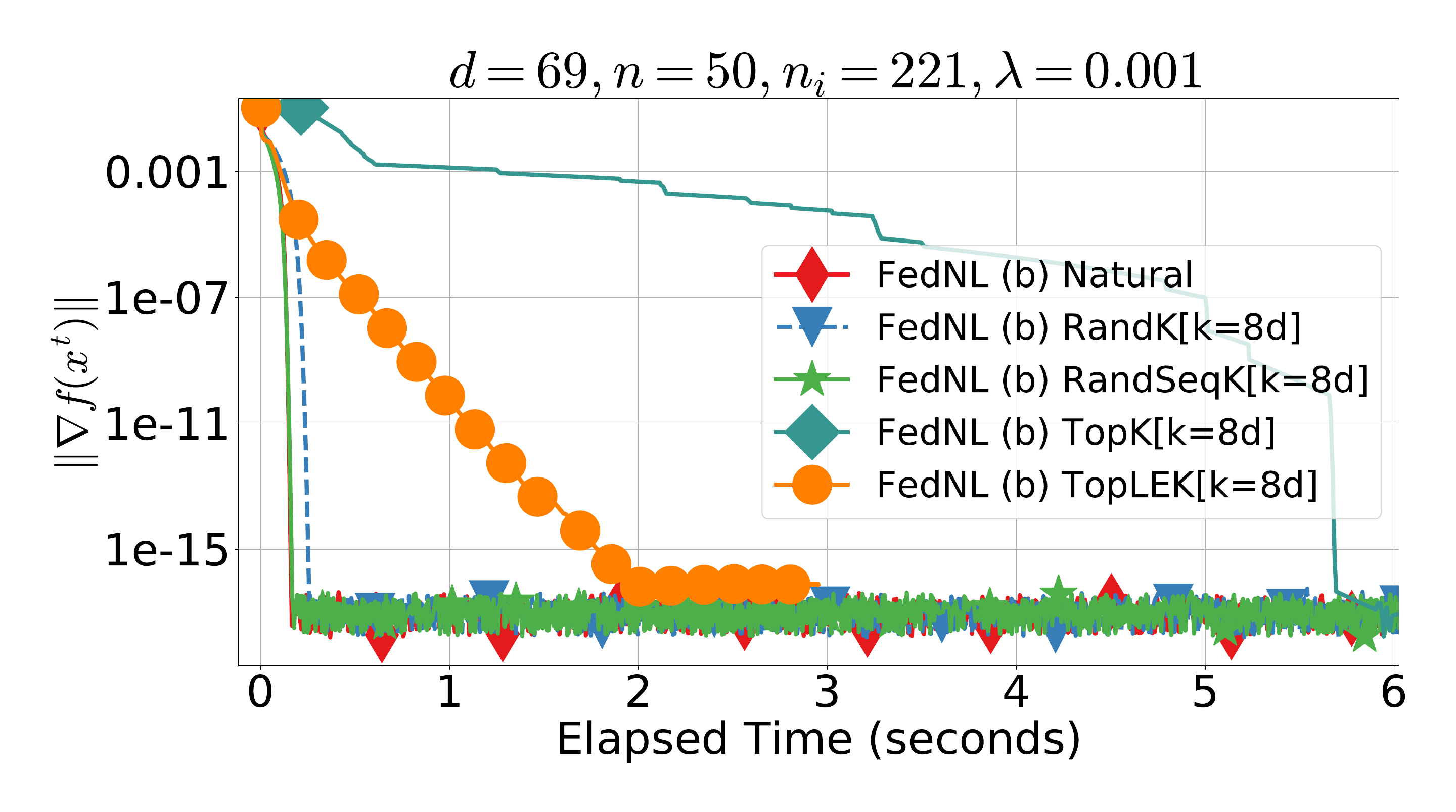}
		\includegraphics[width=0.325\textwidth]{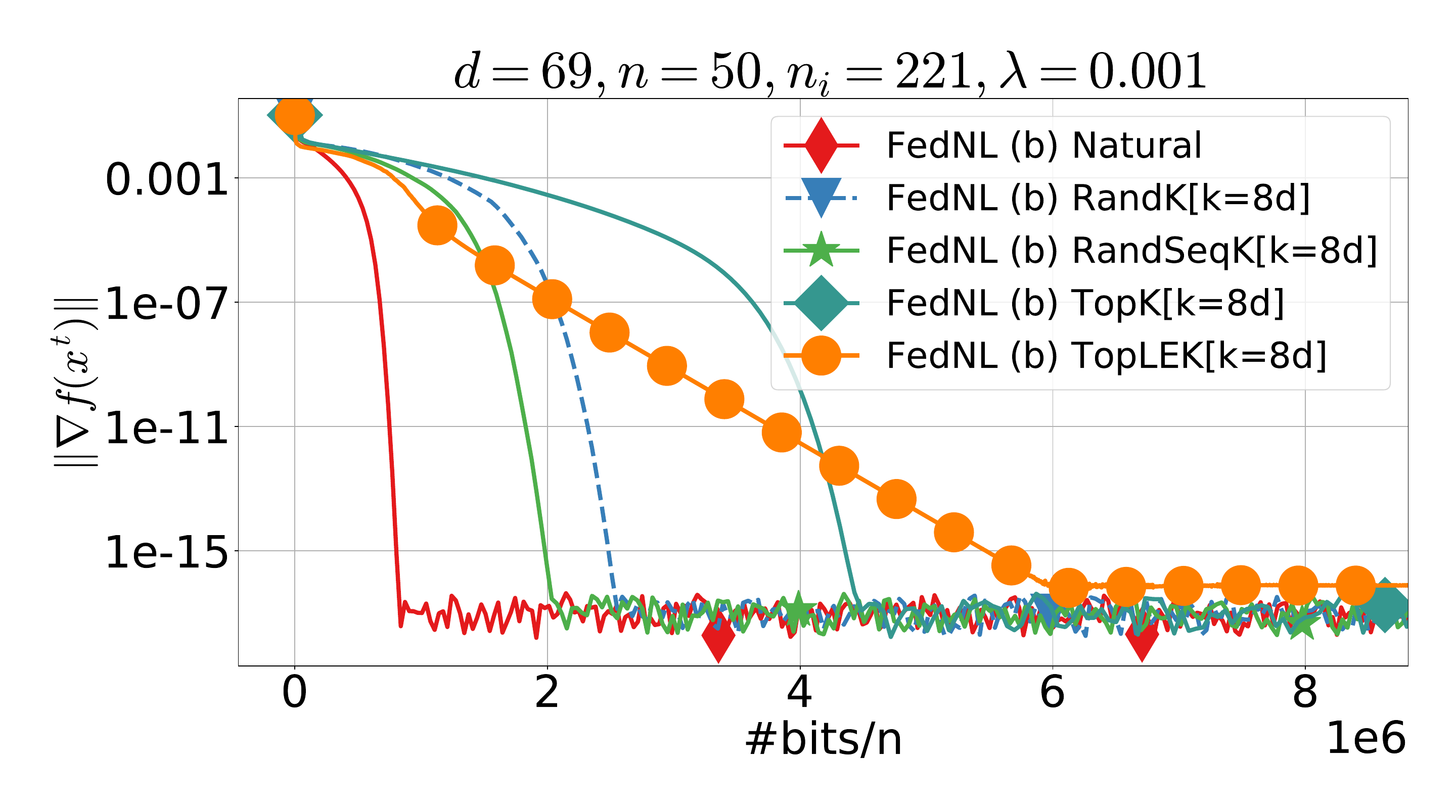}
		\includegraphics[width=0.325\textwidth]{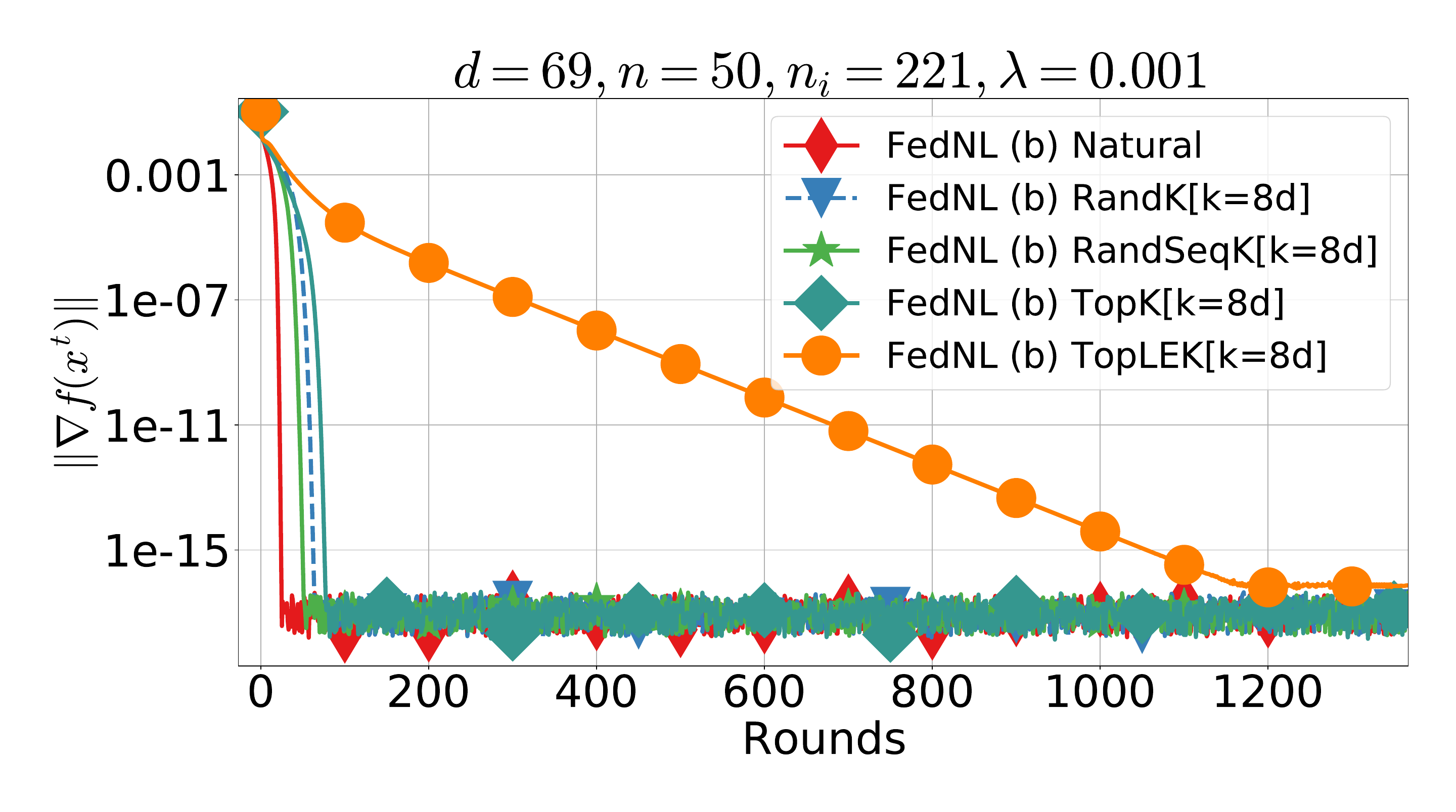}
		
		\caption{\algname{FedNL} in multi-node setting, theoretical step-size, $n=50$, FP64 arithmetic, 1 {CPU} core per node and master, TCP/IPv4, dataset \dataname{PHISHING} reshuffled u.a.r. and augmented with intercept.}
		\label{fig:fednl-phishing-app}
	\end{figure}
	
	\begin{figure}[h]
		\centering
		\includegraphics[width=0.325\textwidth]{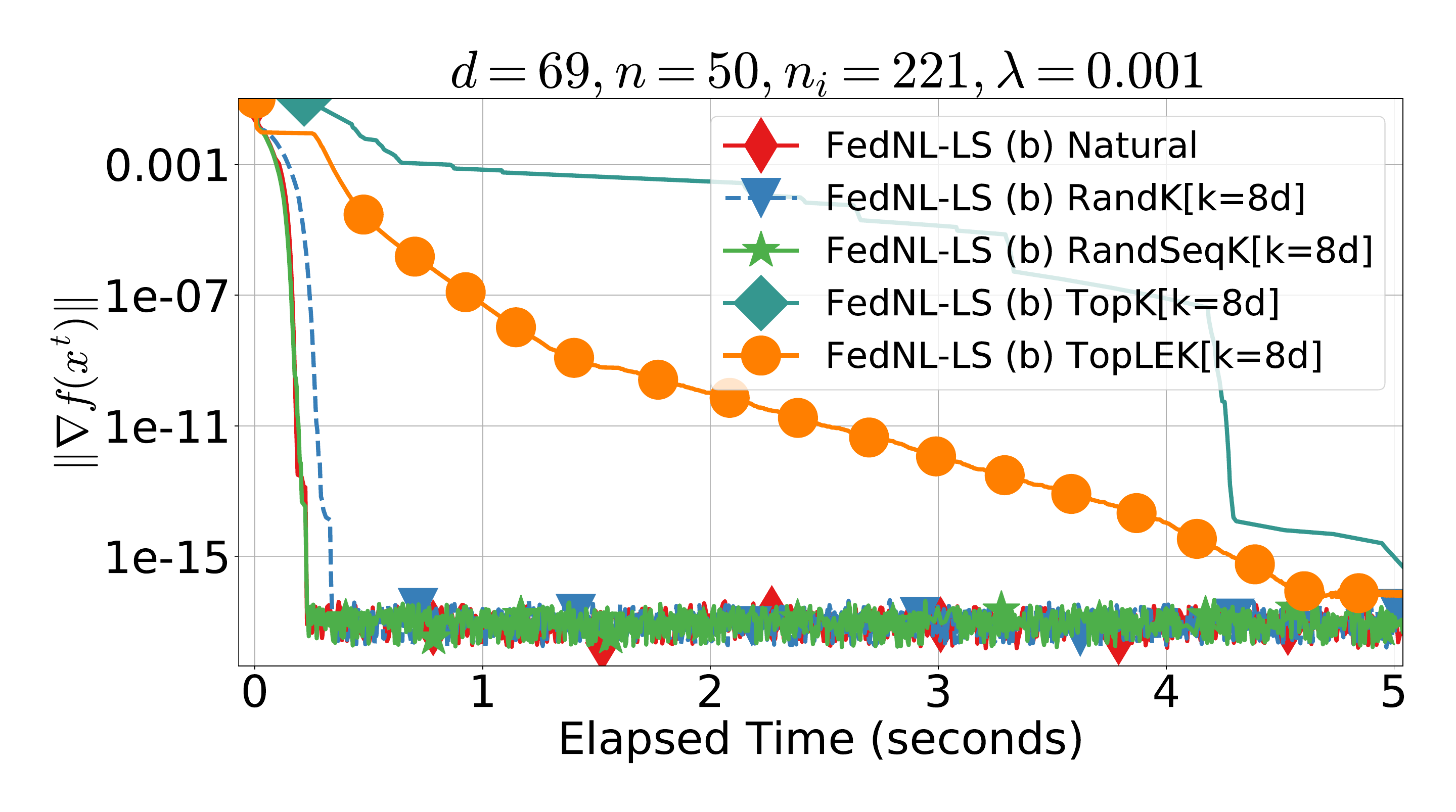}		
		\includegraphics[width=0.325\textwidth]{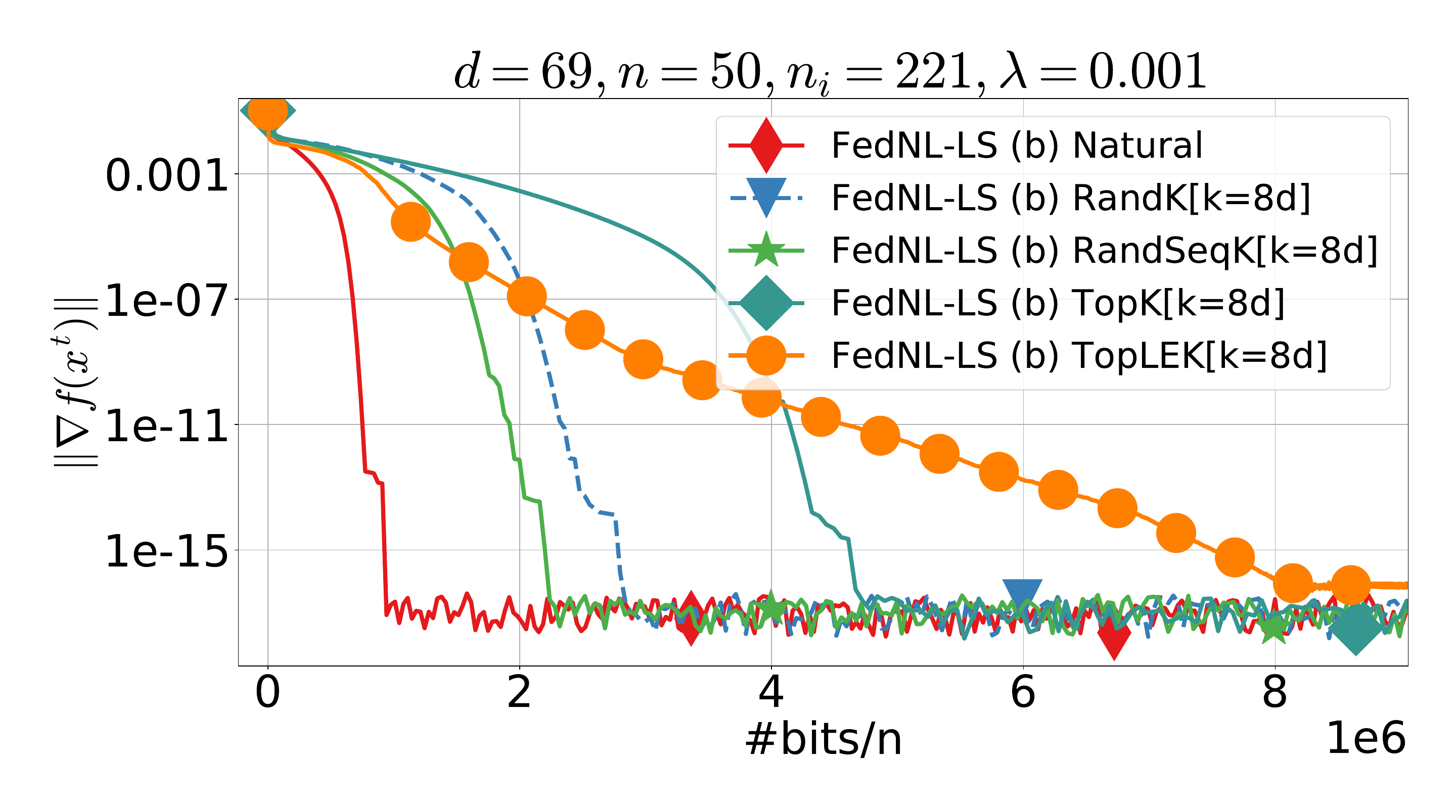}
		\includegraphics[width=0.325\textwidth]{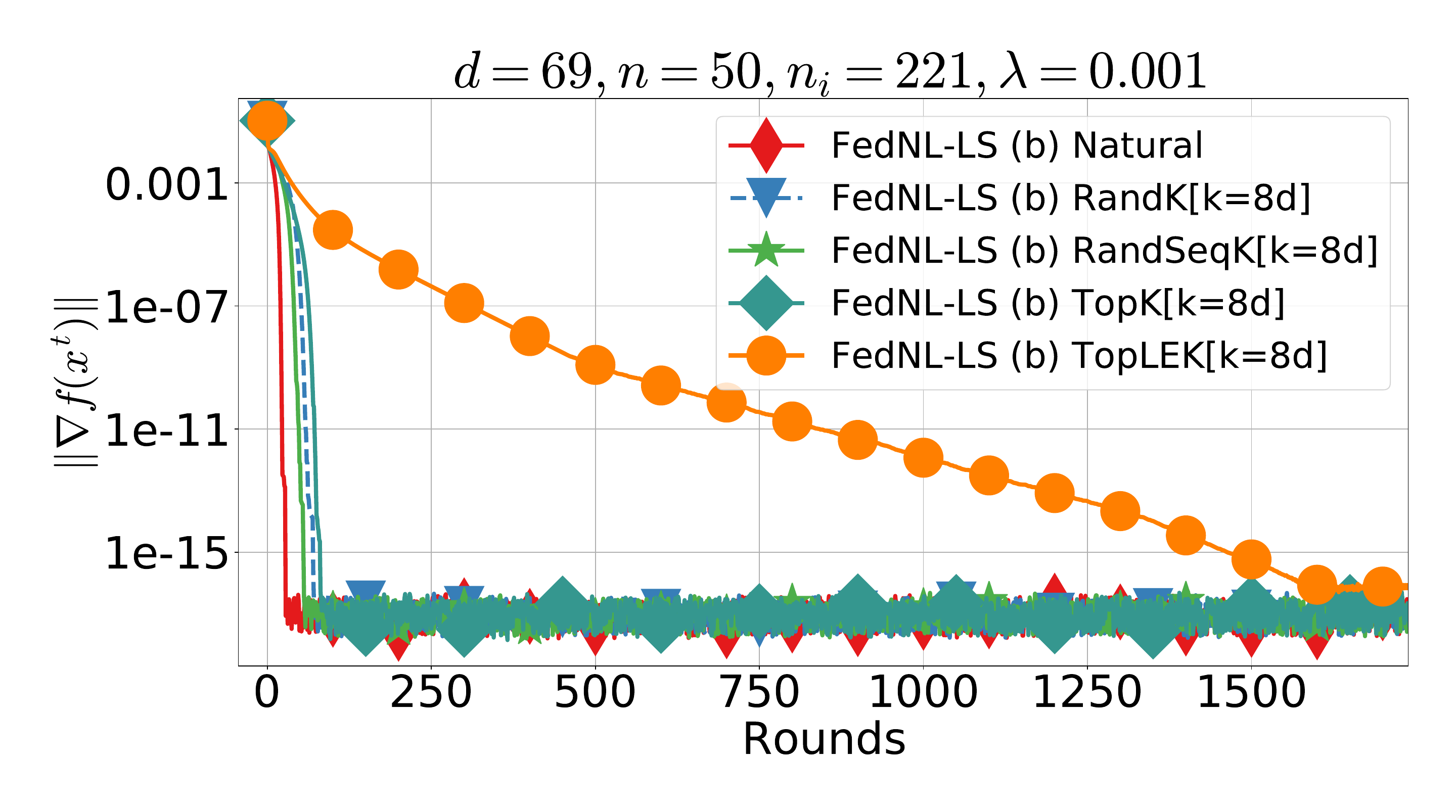}
		
		\caption{ \algname{FedNL-LS} in multi-node setting, $n=50$, FP64 arithmetic, 1 {CPU} core per node and master, TCP/IPv4, dataset \dataname{PHISHING} reshuffled u.a.r. and augmented with intercept. The line search parameters $c=0.49,\gamma=0.5$.}
			
		\label{fig:fednl-ls-phishing-app}
	\end{figure}
	
	\begin{figure}[h]
		\centering
		\includegraphics[width=0.325\textwidth]{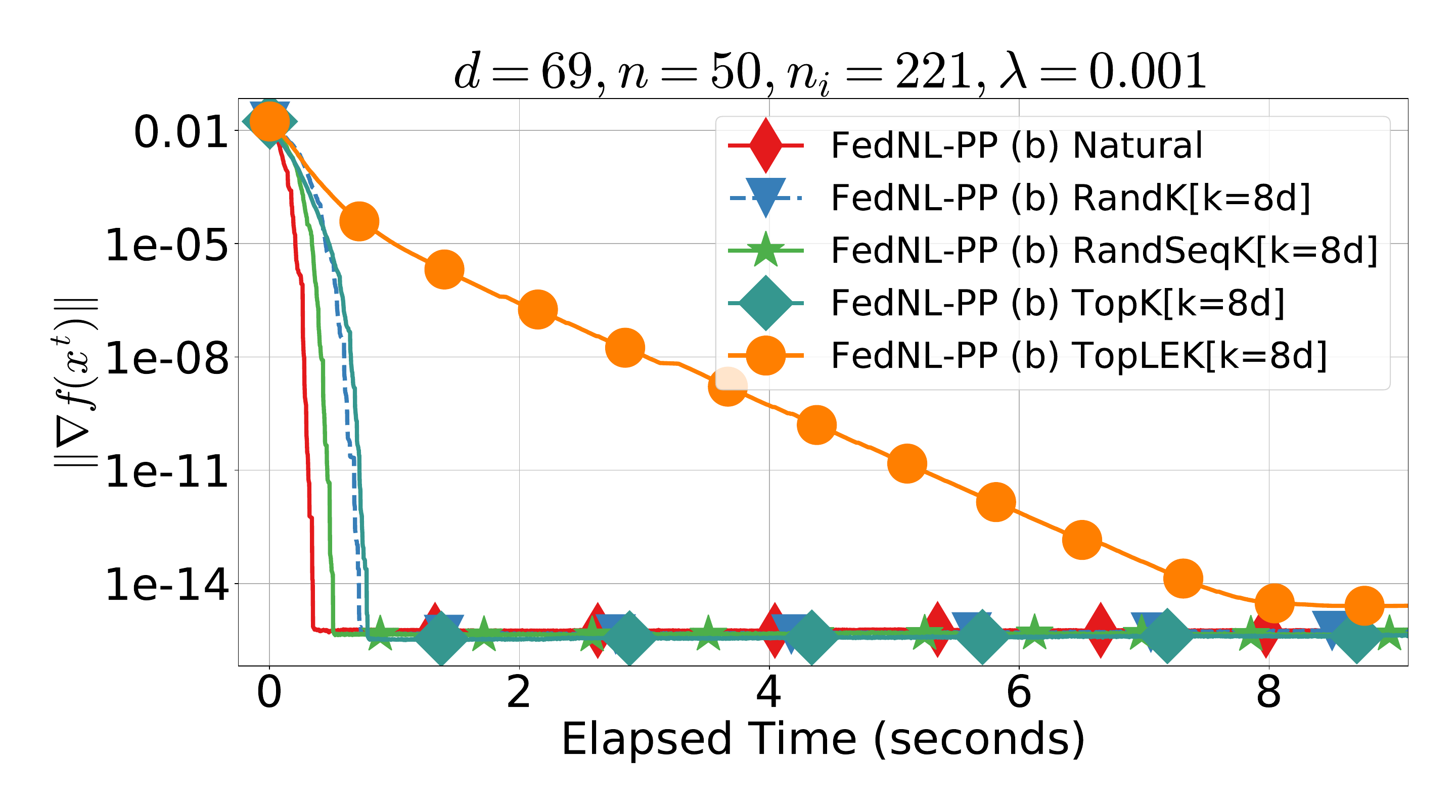}		
		\includegraphics[width=0.325\textwidth]{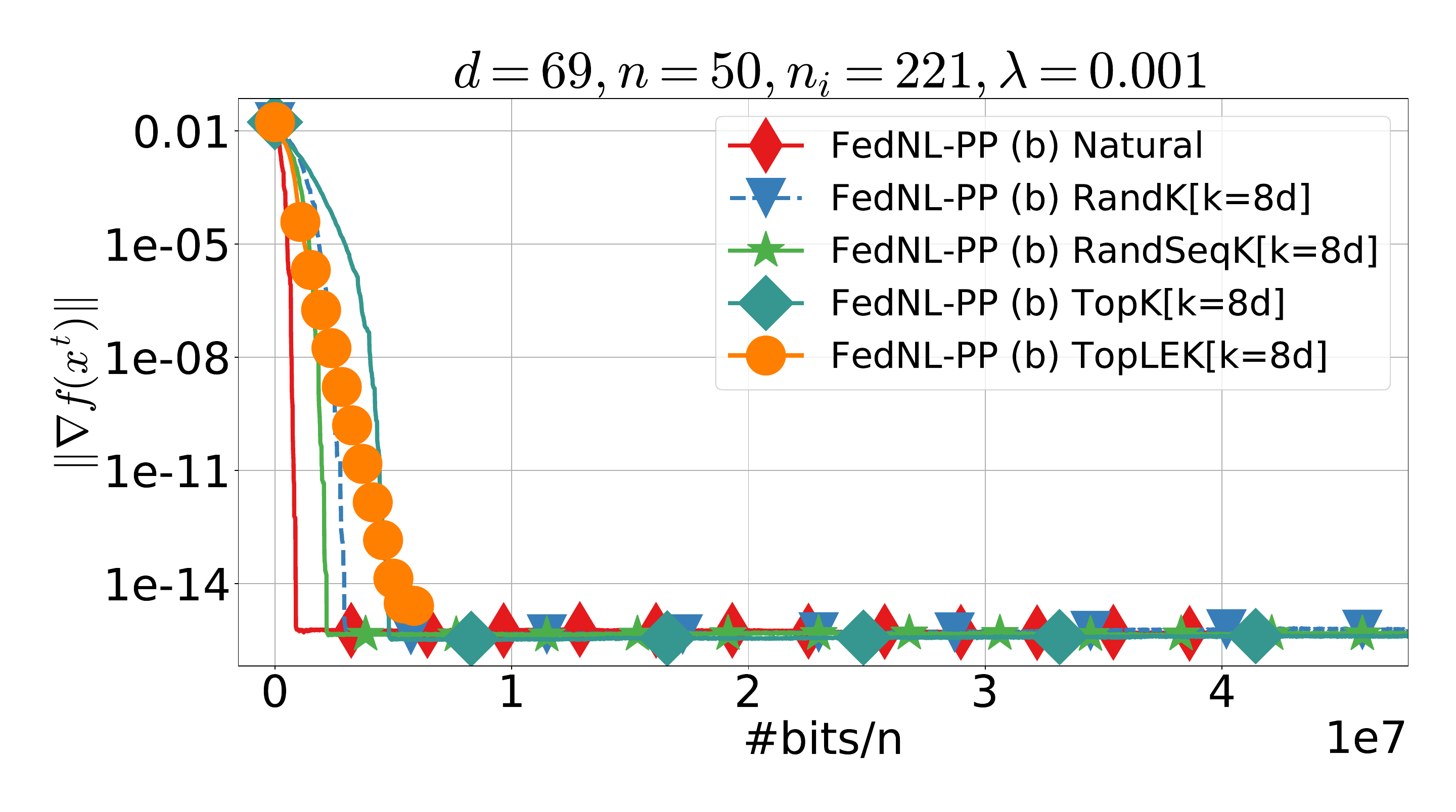}
		\includegraphics[width=0.325\textwidth]{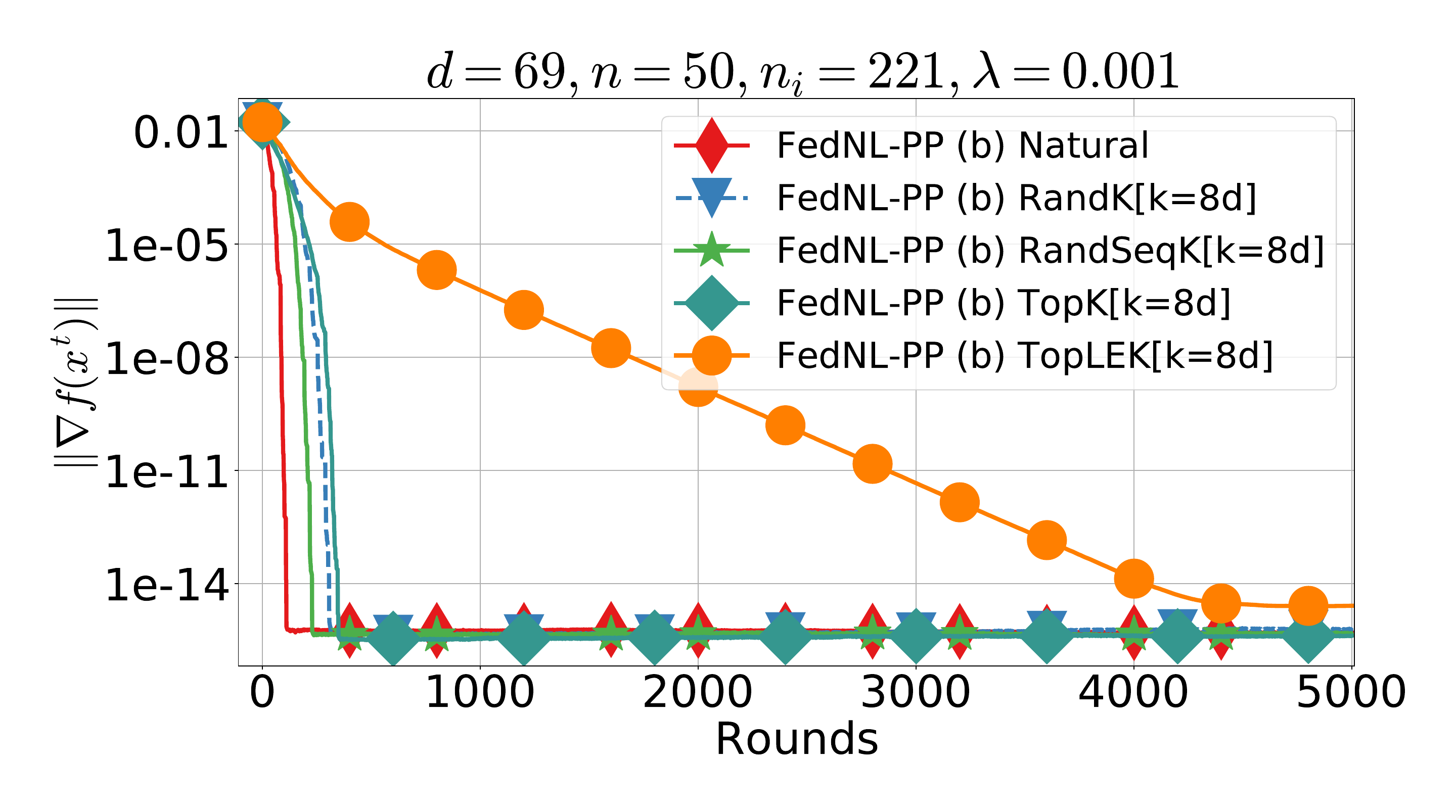} 
		\caption{\algname{FedNL-PP}  in multi-node setting, $n=50$, $|S^k|=12$ clients per round, FP64 arithmetic, 1 {CPU} core per node and master, TCP/IPv4. \dataname{PHISHING} dataset reshuffled u.a.r. and augmented with intercept.}
		\label{fig:fednl-pp-phishing-app}
	\end{figure}

	
	We have carried out experiments with \algname{FedNL}, \algname{FedNL-LS}, and with \algname{FedNL-PP} \algname{FedNL-PP}. The results for \dataname{W8A} are presented in Figures \ref{fig:fednl-w8a-app}, \ref{fig:fednl-ls-w8a-app},  \ref{fig:fednl-pp-w8a-app}, for \dataname{A9A} in Figures  \ref{fig:fednl-a9a-app}, \ref{fig:fednl-ls-a9a-app}, \ref{fig:fednl-pp-a9a-app},  for \dataname{PHISHING} in Figures  \ref{fig:fednl-phishing-app}, \ref{fig:fednl-ls-phishing-app}, \ref{fig:fednl-pp-phishing-app}.
	

	
	In this experiment, we conducted the training of a logistic regression model, formally defined in Eq.\eqref{eq:fi_log_reg_structure}, incorporating $L_2$ regularization on three \dataname{LISVM} \cite{chang2011libsvm} benchmark datasets: \dataname{W8A}, \dataname{A9A}, \dataname{PHISHING}. The training was done within a distributed data center, utilizing the hardware detailed in Appendix~\ref{app:hardware-env-multipled-node}. Fifty clients were connected to a single master node, facilitating communication through a single TCP/IPv4 connection between client and master.
	
	The \algname{FedNL} Algorithm~\ref{alg:FedNL}, particularly in Line 10, necessitated compute reduction for $s^k, l^k$. This reduction was implemented using various communication patterns, conceptualizing client-master communication in a star topology. The master collected information from clients in a centralized manner. Both clients and masters within the experimental setup operated with a single CPU core. Refer to Appendix~\ref{app:hardware-env-multipled-node} for a more comprehensive description of machine capabilities. In this scenario, we disabled the dedicated pool of workers for handling hierarchically gradient reduction and a pool of workers to parallelize the in-place update of the master Hessian.
	
	Firstly, it is observed that the backtracking line search in \algname{FedNL-LS} slowed down the entire training process by a factor of $\times 1.14$, but it ensures global convergence.
	
	The proposed \compname{RandSeqK} outperformed \compname{RandK}, showcasing better convergence as wall-clock time progressed, communication volume increased, and external iterations (rounds) progressed. This is due to two contributing factors. The  \compname{RandSeqK} being cache-aware and consistently reaching the super-linear convergence neighborhood empirically better. The proposed \compname{TopLEK} compressor emerged as the most economical method for sending information from client to master. While it did not enhance \compname{TopK} when measuring progress in terms of communicated bits, it demonstrated improvement when measuring actual time in \dataname{W8A} and \dataname{A9A} datasets. In the \dataname{PHISHING}, it is less effective compared to \compname{TopK} due to an unfortunate situation with catching the neighborhood slower.
	
	Regarding \algname{FedNL-PP}, we caution readers against extrapolating wall-clock time beyond this experiment. The \algname{FedNL-PP} Algorithm~\ref{alg:FedNL-PP} lacks explicit support for the computation of $\nabla f(x^k)$ as part of the training process, and computation of full gradient brought the measured time overhead.
	
	Although our primary goal initially centered on enhancing approaches from the paper \cite{safaryan2021fednl}, scientific curiosity prompted us to explore the application of the \compname{Natural Compressor}. Designed to be unbiased and adapted to a real representation of $\mathbb{R}$ in FP32 and FP64 \cite{IEEE754-2008} with a small $w=1/8$ constant, this compressor was originally proposed within the realm of first-order compressed gradient descent methods. Remarkably, it exhibited favorable behavior for \algname{FedNL} as well. It is important to note that the \compname{Natural Compressor} introduces technical challenges in effective implementation due to operating at the granularity of bits. Despite these, our implementation effectively supports it.
	
	
	\clearpage
	\section{Beyond Training Time: Relaxed Memory and System Requirements}
	
	In this Appendix, we present runtime overhead benchmarks that have been done in Windows OS.
	
	\begin{center}
		\begin{table}[h!]
			\centering
			\begin{threeparttable}
				\caption{Peak Kernel Handles. Single-node Simulation, FP64, Windows 11.}
				\begin{tabular}{|l|c|c|c|c|c|}
					\hline
					Solver & \makecell{W8A $(d=301, n_i=350)$} & \makecell{A9A $(d=124, n_i=229)$} & \makecell{PHISHING $(d=69, n_i=77)$}\\										
					\hline
					\hline
					CLARABEL & 809 & 809 & 809 \\
					\hline
					ECOS & 809 & 809  & 809 \\
					\hline
					ECOS-BB & 811 & 811 & 811   \\
					\hline
					SCS & 811 & 811 & 811   \\
					\hline
					MOSEK & 1087 & 1087 & 1087   \\
					\hline 
					\makecell[l]{\algname{FedNL}/RandK[k=8d]} & \cellcolor{bgcolorwe} 74 & \cellcolor{bgcolorwe} 74 & \cellcolor{bgcolorwe} 74   \\
					\hline
					\makecell[l]{\algname{FedNL}/RandSeqK[k=8d]} & \cellcolor{bgcolorwe} 74 & \cellcolor{bgcolorwe} 74 & \cellcolor{bgcolorwe} 74 \\
					\hline 
					\makecell[l]{\algname{FedNL}/TopK[k=8d]} & \cellcolor{bgcolorwe} 74 & \cellcolor{bgcolorwe} 74 & \cellcolor{bgcolorwe} 74  \\
					\hline 
					\makecell[l]{\algname{FedNL}/TopLEK[k=8d]} & \cellcolor{bgcolorwe} 74 & \cellcolor{bgcolorwe} 74 & \cellcolor{bgcolorwe} 74  \\
					\hline
					\makecell[l]{\algname{FedNL}/Natural} & \cellcolor{bgcolorwe} 74 & \cellcolor{bgcolorwe} 74 & \cellcolor{bgcolorwe} 74   \\
					\hline
					\makecell[l]{\algname{FedNL}/Ident} & \cellcolor{bgcolorwe} 74 & \cellcolor{bgcolorwe} 74 & \cellcolor{bgcolorwe} 74 \\
					\hline	
				\end{tabular}
				\label{tbl:compare-vs-cvxpy-sys-handles}
			\end{threeparttable}
		\end{table}
	\end{center}
	
	\paragraph{Peak kernel handles} Stacking layers of scientific software on top of each other leads to a sufficiently big amount of OS kernel objects\footnote{Examples of kernel objects include files, memory-mapped file views, processes, threads, semaphores, and mutexes. These objects belong to the operating system, but they are created and managed in response to process activity.}. Each kernel handle (in any OS) consumes system resources, such as kernel memory and system table entries. As we see from Table \ref{tbl:compare-vs-cvxpy-sys-handles} \abr{FedNL} simulation imposes a smaller number of kernel handles (responsible for file objects, thread objects, etc.). {Every kernel object occupies memory allocated within the kernel space of the OS and is managed by it. Regardless of whether the memory is paged or pinned, the object consumes resources within the kernel.}
	
	\begin{center}
		\begin{table}[h!]
			\centering
			\begin{threeparttable}
				\caption{Peak Private Bytes. Single-node Simulation, FP64, Windows 11.}
				\begin{tabular}{|l|c|c|c|c|c|}
					\hline
					Solver & \makecell{W8A $(d=301, n_i=350)$} & \makecell{A9A $(d=124, n_i=229)$} & \makecell{PHISHING $(d=69, n_i=77)$} \\
					\hline
					\hline
					CLARABEL & 5 818 936 K & 5 472 376 K & 5 248 824 K \\
					\hline
					ECOS & 5 681 656 K & 5 387 576 K  & 5 192 228 K \\
					\hline
					ECOS-BB & 5 681 936 K & 5 388 092 K & 5 192 048 K   \\
					\hline
					SCS & 5 859 308 K & 5 546 052 K & 5 196 608 K   \\
					\hline
					MOSEK & 6 685 032 K & 5 945 336 K & 5 718 484 K   \\
					\hline 
					\makecell[l]{\algname{FedNL}/RandK[k=8d]} & \cellcolor{bgcolorwe} 770 740 K & \cellcolor{bgcolorwe} 192 296 K & \cellcolor{bgcolorwe} 47 284 K   \\
					\hline
					\makecell[l]{\algname{FedNL}/RandSeqK[k=8d]} & \cellcolor{bgcolorwe} 746 072 K & \cellcolor{bgcolorwe} 191 516 K & \cellcolor{bgcolorwe} 45 236 K \\
					\hline 
					\makecell[l]{\algname{FedNL}/TopK[k=8d]} & \cellcolor{bgcolorwe} 745 072 K & \cellcolor{bgcolorwe} 192 208 K & \cellcolor{bgcolorwe} 45 484 K  \\
					\hline 
					\makecell[l]{\algname{FedNL}/TopLEK[k=8d]} & \cellcolor{bgcolorwe} 745 868 K & \cellcolor{bgcolorwe} 192 908 K & \cellcolor{bgcolorwe} 45 160 K  \\
					\hline
					\makecell[l]{\algname{FedNL}/Natural} & \cellcolor{bgcolorwe} 806 256 K & \cellcolor{bgcolorwe} 199 792 K & \cellcolor{bgcolorwe} 45 568  K   \\
					\hline
					\makecell[l]{\algname{FedNL}/Ident} & \cellcolor{bgcolorwe} 805 996 K & \cellcolor{bgcolorwe} 199 744 K & \cellcolor{bgcolorwe} 46 856 K \\
					\hline	
				\end{tabular}
				\label{tbl:compare-vs-cvxpy-vir-memory}
			\end{threeparttable}
		\end{table}
	\end{center}
	
	\paragraph{Used virtual memory during execution} As we see from Table \ref{tbl:compare-vs-cvxpy-vir-memory} there is a substantial requirement for virtual memory for launching Python scripts. Unfortunately, even a numerical library by itself such as NumPy requires 512 MBytes of virtual address space. The reasons are: (a) NumPy relies on numerical libraries that cumulatively have not negligible size; (b) NumPy initializes big-size internal data structures; (c) The Python ecosystem is completely built on dynamic(shared) libraries (See also Appendix~\ref{app:nopython}).
	
	\begin{center}
		\begin{table}[h!]
			\centering
			\begin{threeparttable}
				\centering
				\caption{Peak Working Set Size (Resident Set Size). Single-node Simulation, FP64, Windows 11.}
				\begin{tabular}{|l|c|c|c|c|c|}
					\hline
					Solver & \makecell{W8A $(d=301, n_i=350)$} & \makecell{A9A $(d=124, n_i=229)$} & \makecell{PHISHING  $(d=69, n_i=77)$} \\
					\hline
					\hline
					CLARABEL & 816 248 K & 522 592 K & 323 364 K \\
					\hline
					ECOS & 668 152 K & 408 440 K  & 256 144 K \\
					\hline
					ECOS-BB & 668 328 K & 409 080 K & 255 916 K   \\
					\hline
					SCS & 813 080 K K & 504 560 K & 281 220 K   \\
					\hline
					MOSEK & 1 216 172 K & 706 616 K & 415 476 K   \\
					\hline 
					\makecell[l]{\algname{FedNL}/RandK[k=8d]} & \cellcolor{bgcolorwe} 750 740 K & \cellcolor{bgcolorwe} 187 888 K & \cellcolor{bgcolorwe} 47 204 K   \\
					\hline
					\makecell[l]{\algname{FedNL}/RandSeqK[k=8d]} & \cellcolor{bgcolorwe} 726 140 K & \cellcolor{bgcolorwe} 187 552 K & \cellcolor{bgcolorwe} 44 168 K \\
					\hline 
					\makecell[l]{\algname{FedNL}/TopK[k=8d]} & \cellcolor{bgcolorwe} 725 512 K & \cellcolor{bgcolorwe} 187 740 K & \cellcolor{bgcolorwe} 44 560 K  \\
					\hline                                    
					\makecell[l]{\algname{FedNL}/TopLEK[k=8d]} & \cellcolor{bgcolorwe} 725 380 K & \cellcolor{bgcolorwe} 188 416 K & \cellcolor{bgcolorwe} 44 260 K  \\
					\hline                                    
					\makecell[l]{\algname{FedNL}/Natural} & \cellcolor{bgcolorwe} 745 532 K & \cellcolor{bgcolorwe} 189 988 K & \cellcolor{bgcolorwe} 45 532 K   \\
					\hline                                    
					\makecell[l]{\algname{FedNL}/Ident} & \cellcolor{bgcolorwe} 785 960 K & \cellcolor{bgcolorwe} 195 892 K & \cellcolor{bgcolorwe} 45 732 K \\
					\hline	
				\end{tabular}
				\label{tbl:compare-vs-cvxpy-ram-memory}
			\end{threeparttable}
		\end{table}
	\end{center}
	
	\paragraph{Peak working set size} In Windows OS, the "peak working set size" refers to the maximum amount of physical memory that a process acquires. We see from Table \ref{tbl:compare-vs-cvxpy-ram-memory} \abr{FedNL} on average exhibits a smaller peak working set by a factor $\times 1.5$ in \dataname{W8A} dataset, and by a factor $\times 6$
	in \dataname{PHISHING} dataset.

	\clearpage
	\section{Hardware Environment for Experiments}
	\label{app:hardware-env}
	
	\subsection{Single-Node Experiment Setup}
	\label{app:hardware-env-single-node}
	
	\begin{enumerate}
		\item \abr{CPU}: Intel(R) Xeon(R) Gold 6246 CPU @ 3.3GHz; Byte Order: Little-Endian
		\item \abr{OS}: Ubuntu 18.04.6 LTS, bionic, x86\_64; Linux kernel version: 5.4.0-150-generic
		\item Physical memory: 251 GB DDR4 Synchronous memory at 2933 MHz
		\item Hard Drive: INTEL SSD SC2 KG03 Disk with a physical sector size of 4096 bytes
		\item Disk write speed: 347 MBytes/s; disk read speed: 2.5 GBytes/s; File System: EXT4
	\end{enumerate}
	
	\subsection{Multi-Node Experiment Setup}
	\label{app:hardware-env-multipled-node}
	
	\begin{enumerate}
		
		\item \abr{CPU}: Intel(R) Xeon(R) Gold 6148 CPU 2.50GHz; Byte Order: Little-Endian
		\item \abr{OS}: Rocky Linux 9.1 (Blue Onyx), x86\_64; Linux kernel version: 5.14.0
		\item Physical memory: 375 GB
		\item Disk write speed: 109 MB/s; Disk read speed: 3.8 GB/s; File System: Network File System v4
		\item Network Interface: MTU Size $1500$ bytes, download 293.18 Mbit/s, upload 476.79 Mbit/s
		\item The experiments in Section \ref{sec:multi-node-cmp-vs-spark} were conducted in a Compute Cluster managed by the Slurm Workload Management system
		\item Masters nodes: $1$; workers nodes: $50$; number of CPU cores/node: $1$
	\end{enumerate}

	\subsection{System Preparation for Reliable Time Measurements and Experimental Reproducibility}
	\label{app:careful-reproduce}
	
	Our experiments were conducted on a system equipped with \textit{Intel Xeon Gold 6246 CPU} \footnote{\href{https://ark.intel.com/content/www/us/en/ark/products/193969/intel-xeon-gold-6246-processor-24-75m-cache-3-30-ghz.html}{Intel Xeon Gold 6246 Processor Technical Specification}}. The CPU has a cache line size of $64$ bytes, $12$ physical cores, and $24$ logical cores, with a core frequency from $3$ GHz to $4.4$ GHz\footnote{The \textit{frequency} is the inverse of the clock \textit{period}, where the clock \textit{period} is the time required to complete one full cycle of the synchronization signal. The typical physical device that generates this signal for electrical circuits is a \textit{crystal oscillator} \cite{harris2015digital}.}. To ensure accurate time measurements we did the following steps:
	
	\begin{enumerate}
		\item \textit{{Disabled Intel Turbo Boost:}} This technology dynamically increases the CPU clock speed above the base frequency based on workload demands. Disabling Turbo Boost ensures a more predictable and consistent level of performance.
		
		\item \textit{{Disabled Simultaneous Multithreading (SMT):}} SMT allows multiple execution threads on a single physical CPU core. We disabled SMT to avoid performance fluctuations.
		
		\item \textit{{Disabled Intel CPU Frequency Scaling (CPU power management):}} This technology allows the operating system to adjust the CPU frequency dynamically.
		
		\item \textit{{Fixed CPU Frequency:}} The CPU frequency during the experiments was set at a stable 3.3 GHz clock rate using the \texttt{cpupower frequency-set} command-line utility in Linux.
		
	\end{enumerate}

	\clearpage
	\section{Software Environment for Experiments}
	\label{app:software-env}
	\subsection{Software Environment for Baselines}
	\label{app:software-python-env}
	
	Version information for used Python libraries: NumPy version: 1.26.2; Pandas version: 2.1.4; scikit-learn version: 1.3.2; Matplotlib version: 3.8.2; Ray version: 1.26.2; Cython version: 3.0.8; CVXPY version: 1.4.1;
	Mosek version: 10.1.21; CVXPY solvers: CBC, CLARABEL, CVXOPT, ECOS, ECOS\_BB, GLOP,	GLPK, GLPK\_MI, GUROBI,	MOSEK, OSQP, PDLP,	SCIP, SCIPY, SCS, XPRESS.
	
	Version information for utilized Python interpreter:
	\begin{itemize}
		\item Python version: 3.9.1 (default, Dec 11 2020); System/OS name: Linux/5.4.0-150-generic
		\item Python interpreter compiled with GCC 7.3.0, CPU architecture: x86\_64
	\end{itemize}
	
	Version information for utilized Apache Spark:
	\begin{itemize}
		\item Apache Spark 3.5.0; PySpark version: 3.5.0
		\item Java Runtime: OpenJDK Runtime Environment (build 11.0.19)
	\end{itemize}
	
	Cluster management system for utilized multi-node settings:
	\begin{itemize}
		\item Experiments from Section \ref{sec:multi-node-cmp-vs-spark} were conducted in a Compute Cluster managed by the Slurm
		\item Slurm version: 23.02.6; Linux OS Kernel: Linux 5.14.0
	\end{itemize}
	
	\subsection{Building Single-Node and Multi-Node FedNL Implementation}
	\label{app:build-fednl-for-exp}

	For wall-clock time measurements, our C++20 \algname{FedNL} implementation was built using the following:
	
	\begin{enumerate}
		\item {Target Architecture:} AMD, Intel x86 64-bit
		\item {CMake Version:} 3.27.0-rc5; {Linux GNU Make Version:} 4.1
		\item {Compiler:} GNU GCC version 11.4.0 with compiling C++20
		\item Optimization flags: \textit{{-O3 -fno-rtti -fno-exceptions -flto -march=native -mtune=native -pthread}}
		\item {Utilized float point format}: FP64. All computations involving matrix-matrix, matrix-vector, and vector-scalar operations are performed in FP64 (double precision)
		\item {CPU Extension:} The compute-intensive sections of the code leverage SIMD \abr{CPU} extension
	\end{enumerate}

	\clearpage
	\section{System Requirements for Unlocking FedNL}
	\label{app:system-requirements-for-ufednl}
	
	\subsection{Unlocking FedNL: Supported Operating Systems and Compilers}
	\label{app:supported-os-and-compilers}
	
	Our solution is designed to run on various desktop operating systems, including:
	
	\begin{enumerate}
		\item Microsoft Windows (Windows 10, Windows 11, and higher)
		\item Linux Operating Systems (Ubuntu 18.04, 22.04, and higher)
		\item macOS Monterey (12.6.1 and higher)
	\end{enumerate}
	
	The supported toolchains for building our code are (i) Visual Studio/MSVC (minimum: VS2022, MSVC 19.30); (ii) GNU GCC/G++ (minimum: 11.4); (iii) LLVM CLANG (minimum: 10.0). 
	\clr{If you want to build a project with CUDA Support please use CUDA Toolkit compatible with 12.4.}
	
	Our code adheres to the C++ 2020 standard\footnote{
		\href{https://www.iso.org/standard/79358.html}{ISO/IEC 14882:2020 C++ standard.}}. To facilitate the build process, we provide a helper script \texttt{project\_scripts.py} written in Python 3.9.4. This script helps with the following tasks:
	
	\begin{enumerate}
		\item Generate release/debug project files via CMake. Please use CMake (minimum: 3.12)
		\item Launch the building process from the command line and parallelize it
		\item Speed up the build with Ninja (GNU Make alternative) and CMake Unity/Jumbo build
		\item Enable the usage of precompiled headers to expedite the build
		\item Enable LLVM optimization compiler remarks for study
		\item Launch unit tests, GNU code coverage, and documentation generation via Doxygen
	\end{enumerate}	
	
	\subsection{Unlocking FedNL: Supported and Tested CPU Architectures}
	\label{app:cpus}
	
	The code can be compiled to support the following processor supplementary instruction sets:
	
	\begin{enumerate}
		\item \textit{{SSE2 (128-bit vector registers)}} – supported by Intel (Pentium 4)
		\item \textit{{AVX2 (256-bit vector registers)}} – supported by Intel (Sandy Bridge), AMD (Bulldozer)
		\item \textit{{AVX-512 (512-bit vector registers)}} – supported by Intel (Knights Landing), AMD (Zen 4)
		\item \textit{{ARM Neon (128-bit vector registers)}} - supported by many ARMv7 processors				
		\item \textit{{No special instructions}} - explicitly turn off usage of any computing  vectorization
	\end{enumerate}

	\paragraph{Processors with Single Instruction Multiple Data (SIMD) facility Used for resting}
	
	\begin{enumerate}
		\item {Intel CPUs with x86\_64 architecture}: Core i7-10875H, Xeon Gold 6246, Xeon Platinum 8272CL, Core i7-8700B; byte order: Little-Endian
		\item {CPUs with {ARM AArch64} architecture}: ARMv8 Neoverse N1 R3P1 Google; byte order: Little-Endian
	\end{enumerate}
	
	The x86\_64 Instruction Set Architecture finds widespread use in desktops, laptops, and servers. Processors from vendors like Intel and AMD, such as Intel Core and AMD Ryzen series, are commonly integrated into systems utilizing the x86\_64 architecture. The ARM AArch64 Instruction Set Architecture (also denoted as A64) is typically utilized in mobile and embedded devices. 
	
	\clr{
		Practical assembly language differences between ARM and x86\_64 in our opinion include the following: (i) ARM's dedicated register for storing return addresses (link register $x30$); (ii) built-in conditional execution within instructions; (iii) the ability of arithmetic instructions to avoid modifying CPU flags; (iv) memory load/store instructions that can incorporate indexing directly; and (v) a dedicated zero register in the ARM ISA.
		
		The primary reason for the growing adoption of the ARM ISA is that the x86\_64 architecture, being a complex instruction set architecture is more challenging to implement. The inclusion of complex instructions often results in simple operations being implementable but significantly less energy-efficient.
	}

	\clr{Both x86 and ARM are prominent and widely used processor architectures in today's computing industry and our implementation supports both of them. We have also tested our implementation on several other CPU architectures. A list of tested processor architectures is below.}

	\clr
	{
		\paragraph{List of tested processor architectures for implementation}
		We tested our implementation across the following CPU architectures on Linux/Ubuntu OS:
		\begin{enumerate}
			\item AMD64/x86\_64: 64-bit logical address; byte order: Little-Endian
			\item ARM v7: 32-bit logical address; byte order: Little-Endian
			\item ARM64 v8: 64-bit logical address; byte order: Little-Endian
			\item PowerPC PPC64LE: 64-bit logical address; byte order: Little-Endian
			\item RISC-V 64: 64-bit logical address; byte order: Little-Endian
			\item IBM Z Series Architecture (S390X): 64-bit logical address; byte order: Big-Endian
		\end{enumerate}
		
		We tested our implementation on MacBooks with the following CPU architectures:
		
		\begin{enumerate}
			\item ARM64 v8 (Apple M1): 64-bit logical address; byte order: Little-Endian
			\item x86\_64 (Intel Core-i7 1068NG7): 64-bit logical address; byte order: Little-Endian
		\end{enumerate}
		
		This demonstrates that our implementation is fully cross-platform, designed to function across multiple operating systems, CPU architectures, integer word sizes, and endianness.
	}
	
	\subsection{Unlocking FedNL: Supported and Tested GPU Architectures}
	\label{app:gpus}
		
	Our current GPU support is limited (see Appendix~\ref{app:limitations}). We primarily focused on the minimal support of creating oracles for optimization problems using NVIDIA CUDA\footnote{\href{https://developer.nvidia.com/cuda-toolkit}{https://developer.nvidia.com/cuda-toolkit} - NVIDIA CUDA Toolkit provides an environment for creating high-performance, GPU-accelerated applications for NVIDIA GPUs.}, providing minimal support for expressing computation in dense and non-structure linear algebra primitives. The used version of the NVIDIA driver is 560.35.03. 
	
	Functionality has been tested on: 

	\begin{enumerate} 
		\item \textit{Rocky Linux 9.1, x86-64, Kernel: Linux 5.14.0-162.23.1.el9}
		\item \textit{Windows 11, x86-64, Versio 23H2, OS Build: 22631.4169}
	\end{enumerate}
	
	With GPUs featuring the following microarchitectures:
	
	\begin{enumerate}
		\item NVIDIA Volta microarchitecture, CUDA Compute Capability: 7.0
		\item NVIDIA Turing microarchitecture, CUDA Compute Capability: 7.5
		\item NVIDIA Ampere microarchitecture, CUDA Compute Capability: 8.0, 8.6
		\item NVIDIA Ada Lovelace microarchitecture, CUDA Compute Capability: 8.9
		\item NVIDIA Hopper microarchitecture, CUDA Compute Capability: 9.0
	\end{enumerate}
	
	In addition, we provide several executable applications for system introspection to detect OpenCL and CUDA-compatible devices and their characteristics (see Table \ref{tab:auxiliary-binaries} in Appendix~\ref{app:futresearch}).
	
	\clearpage
	\section{Backgrounds}
	\label{app:backgrounds}
	
	\subsection{Memory Latency Variance in Computing Systems}
	\label{app:memory-hierachy-latencies}
	
	We present a minimum background that is enough to motivate  \compname{RandSeqK} in Appendix~\ref{app:seqk}. For more about Compute Architecture see \cite{hennessy2011computer}. Modern computing systems are equipped with various storage devices. The following table, adapted from \cite{gregg2014systems}, illustrates typical latency scales for them. More hardware-focused comparisons can be found in digital circuits design literature \cite{harris2015digital}.
		
	\begin{table}[h!]
		\centering
		\caption{Memory Latency Comparison in Computing Devices.}
		\begin{tabular}{|c|c|c|}
			\hline
			Device and Memory Level & Approximate Access Latency (ns) & Scale \\
			\hline
			\hline
			CPU cycle & 0.3 & x1 \\
			\hline
			\clrshort{CPU register (SRAM)} & 0.3 & x1 \\
			\hline
			L1 cache (SRAM) & 0.9 & x3 \\
			\hline
			\clrshort{Floating Point addition, subtraction, and multiplication} & 1.2 & x4
			\\
			\hline
			L2 cache (SRAM) & 3 & x10 \\
			\hline
			L3 cache (SRAM) & 10 & x33 \\
			\hline			
			Main memory or Physical Memory (DRAM) & 100 & x330 \\
			\hline
			\clrshort{The OS System Call: Transitioning from user to kernel space} & 300 & x1000 \\
			\hline
			Solid-State Disk (SSD) & 10 000 & x33 000 \\
			\hline
			Rotational Hard Disk Drive (HDD) & 10 000 000 & x33 000 000 \\
			\hline
		\end{tabular}
		\label{tbl:latencies-for-memory}
	\end{table}
	
	\clr
	{
	In modern CPUs, the functional units responsible for floating-point addition, subtraction, and multiplication typically exhibit a latency of 4 clock cycles \cite{fog_instruction_tables}. While Big-O notation abstracts away many performance details, real-world execution is heavily influenced by data access times, especially when retrieval from disk or I/O is involved. As shown in Table~\ref{tbl:latencies-for-memory}, performance can vary significantly depending on the primary location of the data.	
	}

	\paragraph{A Hard Disk Drive (HDD)} HDD comprises one or more rigid disks or platters, with a head that moves to the correct location on the disk to read or write data magnetically as the disk rotates. Due to its partial mechanical nature, an HDD exhibits relatively slow access speeds. In contrast, Solid State Drives (SSDs) use flash memory, and these types of storage devices have an improved access latency compared to HDD. Only  HDD and SSD from Table~\ref{tbl:latencies-for-memory} store data persistently. 
	
	\paragraph{The OS System Call.} An OS system call is initiated by a process's thread to request the kernel to perform privileged instructions or system tasks, such as interacting with peripherals through device drivers. The transition from user space to kernel space for a specific thread is known as a \textit{mode switch}. In modern OS architectures, threads typically have both a user-level stack and a kernel stack, with the latter used to store temporary data during driver routine execution in kernel mode. While the execution time of the system calls itself can be not a major concern (see Table~\ref{tbl:latencies-for-memory}), the transition can trigger thread rescheduling, leading to cold and capacity cache misses in the CPU. One way to reduce the number of system calls is to utilize advanced OS primitives available in user space, such as memory mapping for file read operations when possible (see Section \ref{app:mmap}).

	\paragraph{Dynamic Random Access Memory (DRAM)} Dynamic Random Access Memory serves as a memory for non-persistent storage. It is structured as a series of memory chips and physically stores data typically as capacitor charge. To safeguard against data corruption, additional logic may store bits for error correction. Accessing DRAM is facilitated through the Memory Controller, which manages access to the primary Memory Chips when the data is not in the cache.
			
	\paragraph{The CPU Memory Cache} The CPU memory cache reduces latency in accessing DRAM by temporarily storing frequently accessed data. It is typically implemented using Static Random-Access Memory (SRAM), which relies on semiconductors for its electrical circuits. In SRAM, each bit of data is typically stored in cross-coupled inverters \cite{harris2015digital}, typically using six transistors. While both SRAM and DRAM lose their stored data when the supply voltage is off, they differ in the materials used and their operational mechanisms. If the requested data is not found in the cache or DRAM, the processor retrieves it from a backup stored in virtual memory on disk.
	
	In general, low-level cache optimization at the software level is challenging due to the complexity of modern CPU architectures, where cache management is often opaque. Two contemporary strategies for optimizing algorithms are designing \textit{cache-aware} algorithms, which consider cache behavior and structure, and \textit{cache-oblivious} algorithms, which adapt to available multilevel cache structure and available cache size in a specific moment implicitly \cite{demaine2002cache}.
	
	\paragraph{Connection between DRAM and CPU Caches} The DRAM Memory Controller is responsible for handling memory transactions between caches and DRAM. When interacting with caches, data is processed in blocks with a fixed size known as \textit{Cache Lines}. Cache line sizes can vary across microarchitectures, but the typical size is $64$ bytes in most x86\_64 and AArch64 architectures.
	
	\subsection{A Background on Network Communication}
	\label{app:networks-background}	
	
	In the realm of the Internet, edge devices transmit packets through routers and switches. On the sender side, the long message is fragmented into smaller units called \abr{packets}. These packets traverse \textit{intermediate routers}, which, in turn, process them, route and enqueue to the appropriate outgoing link, and decide if a packet should be discarded.
	
	When \textit{clients} and \textit{master} are linked via multiple routers, the delays from all intermediate routers cumulatively contribute to a \textit{single packet} delay. Sending a tiny packet from the client to master and back leads to a time metric known as Round Trip Time (\abr{RTT}), analogous to the Latency in \textit{Compute Architecture}. When multiple packets traverse the communication path the rate at which information can be sent may become more crucial. In \textit{Compute Architecture} and \textit{Network Communication}, if several intermediate devices form a pipeline, then the speed of complete operation per time is determined by the slowest component in the pipe. The technical term denoting the speed is called \textit{bandwidth}, and in networks, it is measured in bits/second. If the link for message transfer is shared, the effective bandwidth is divided among clients. The derived quantity that describes the actual bandwidth is denoted as \abr{throughput}. The \textit{{bottleneck link}} is a link with the minimum throughput for communication. If the throughput is $R$ bits/s, and transferred message is $L$ bits, and the sender waits for acknowledgments, then the time for transferring an $L$-bit message can be estimated as:
	\[
	t_{\mathring{delay}} = RTT + \frac{L}{R_{\mathrm{bottleneck}}}.
	\]
	
	This is a classical formula available in literature \cite{kurose2005computer}, but it is important to note its limitations:
	
	\begin{enumerate}
		\item It does not account for packet loss delays
		\item The formula assumes that $RTT$ and $R_{\mathrm{bottleneck}}$ are constants, which may not be true
	\end{enumerate}
	
	\paragraph{Communication network protocols stack} The \textit{Network Communication} also serves as a bridge, linking the \textit{Electrical Engineering} with \textit{Applications} through a stack of protocols consists of:
	\begin{enumerate}
		\item \textit{Application Layer} is designed to facilitate the easy development (HTTP [RFC 7230])
		
		\item \textit{Transport Layer} is designed to hide defects of the underlying protocols (TCP [RFC 793])
		
		\item The\textit{ Network Layer} addresses routing, forwarding, and congestion control  ({IPv4} [RFC 791])
		
		\item The \textit{Link Layer} connects IP protocol with the lower levels via splitting data into frames, carry error control and medium access (CSMA/CD [IEEE 802.3], CSMA/CA [IEEE 802.11]) 
		
		\item The \textit{Physical Layer} takes charge of the actual physics of the utilized medium.  (Bluetooth 802.15.1, Family of Wi-Fi IEEE 802.11, Family of Ethernet IEEE 802.3)
	\end{enumerate}
	
	This comprehensive framework driven by IEEE and RFC (Request for Comments) standards ensures the seamless integration and functionality of diverse elements within the network ecosystem.
	
	\paragraph{The size of transferred messages in TCP/IP} The theoretical maximum size of an \abr{IPv4} packet is $65535$ bytes, inclusive of a $20$-byte header. When using \abr{TCP/IP}, effective communication faces limitations if the message size exceeds the allowable Maximum Transfer Unit (\abr{MTU}) along the communication link. If an \abr{IP} packet surpasses the \abr{MTU}, it undergoes fragmentation, breaking into units with \abr{MTU} size. These fragments are later reassembled recreating the original packet, but \abr{IP} fragmentation is generally considered undesirable. IP protocols define minimum \abr{MTU} sizes - $576$ bytes for \abr{IPv4} and $1500$ bytes for IPv6. The TCP header is $20$ bytes long, the \abr{IPv4} header typically spans $20$ bytes, and the \abr{IPv6} header is $40$ bytes. When operating at the \abr{TCP} level to avoid fragmentation, the payload should be less than the \abr{MTU} size minus \abr{IP} and \abr{TCP} header sizes. When Nagle Algorithm \cite{nagle1984congestion} is turned off, the \abr{TCP/IP} stack inside the \abr{OS} will not delay sending small packages. And sending should be done with knowledge about this effect. For more see \cite{kurose2005computer}.

	\paragraph{Hardware support for TCP/IP in the end systems.}
	
	Modern communication devices supporting TCP/IP are physically implemented as \textit{network interface cards} (NICs). These devices provide one or more ports to connect to an external communication medium which in practice can be optical, electrical, or wireless. Internally, a NIC contains a component known as a \textit{network controller}, which is driven by its microprocessor. The network controller facilitates the efficient transfer of packets between the NIC's \textit{ports} and the Input/Output subsystem of the computing system. During increasing amounts of sent information, both the network port and hardware interface connection for the Input/Output subsystem can be a bottleneck.
	
	\paragraph{Operating support for TCP/IP in the end systems.}
		
	User-level applications typically access the network interface card (NIC) using the TCP/IP protocol through programmable endpoints, known as sockets, via the Berkeley socket API. This API, which has been in use for decades, is now ported to virtually every operating system (OS). Essentially this API provides a mechanism to transfer data between the application and OS driver buffers, as well as the OS kernel buffers. After this the NIC reads and sends raw data from the OS kernel buffer, utilizing the Direct Memory Access (DMA) controller for packet transmission. When an incoming packet arrives at the NIC, the OS is generally notified through an asynchronous interrupt service request (IRQ). Some OS and NIC configurations slightly modify this standard behavior. For example, in Linux, if an interrupt coalescing mode is used, interrupts are not raised after arriving at each packet. Instead, an interrupt is raised either after a timeout or after a specified number of packets have been received.
	
	TCP (Transmission Control Protocol) is a connection-oriented protocol that provides a reliable data transfer service on top of the unreliable IP service. However, it is well-known that TCP includes several unspecified characteristics, the management of which falls under the responsibility of the OS. Some of these characteristics include:
	
	\begin{itemize}
		\item The first step in a connection-oriented protocol is establishing a TCP connection through a three-way handshake. The time to establish the initial connection is referred to as connection latency. However, TCP at the protocol level does not specify timeouts for connection establishment.
		\item During message transfer, TCP facilitates the exchange of data between nodes in a communication graph. However, TCP at the protocol level does not specify exactly how the OS should handle out-of-order segments.
		\item For the purpose of closing the connection one side initiates the connection closure by sending a packet with the FIN flag. The other side acknowledges this packet and sends a symmetric FIN packet in return. After this exchange, the OS enters the TIME\_WAIT state for the specific connection(socket), which can last for a significant period (e.g., 1 minute), depending on the OS configuration. The resources associated with the connection can only be released after this period. This timeout is not specified by the TCP protocol as well.
	\end{itemize}
	
	It is also important to note that certain operating systems provide alternative ways for interacting with the NIC to use TCP/IP. For example, the Data Plane Development Kit (DPDK)\footnote{The Data Plane Development Kit. Open source project managed by the Linux Foundation: \href{https://www.dpdk.org/}{https://www.dpdk.org/}} and the eXpress Data Path (XDP) \footnote{eXpress Data Path(XDP). A high performance, programmable network data path in the Linux kernel: \href{https://prototype-kernel.readthedocs.io/en/latest/networking/XDP/introduction.html}{https://prototype-kernel.readthedocs.io/}} in POSIX OS and Windows OS environments enable tighter coupling between user-space logic and the NIC, at the cost of added complexity in user-space application development. These interfaces require the user to implement part of the TCP/IP logic within the application itself. Although relatively new, these interfaces open up additional opportunities for optimizing network communication.

	\clearpage
	\clr{
		\section{{Broader Impact}}
		\label{app:discussions}

		\subsection{Towards More Realistic Scientific Investigations}
		\label{app:scientific-experiments}
		\paragraph{Importance of Distinguishing Design and Runtime.}
		According to Amdahl's Law, overall performance is constrained by the weakest component. In complex, multi-layered software stacks, identifying bottlenecks becomes increasingly difficult, as multiple layers and their interactions can affect performance. The C++ language offers several distinct advantages: (i) its design facilitates systematic and modular thinking without introducing runtime overhead; (ii) user-defined types are as efficient as built-in types; and (iii) unused language features or libraries incur no time or memory cost. These properties made C++ an essential choice for our work, providing the fine-grained control required for in-depth exploration, while maintainability-related constructs are optimized away at compile time.
		
		\paragraph{Practical performance measurement for research ideas.}
		
		Our implementation allows researchers to measure execution time in seconds, a capability often lacking in prototypical setups. Even theoretically optimal algorithms may unperformed in practical benchmarks that measure time and memory due to large hidden implementation constants in layered designs. Our methodology enables modifications without compromising performance.
}

\clr{
\subsection{The Role of Theory in Practical Implementation}
\label{app:theory-role}

\paragraph{Convergence Guarantees}
The convergence guarantees of the \algname{FedNL} family ensure that these algorithms not only converge but also effectively address the problems outlined in Eq. \ref{eq:main}, under Assumptions \ref{asm:1} and \ref{asm:2}. Our objective was to implement this algorithm family effectively within a Federated Learning (FL) setting, which is characterized by a less controlled training environment. The theory provides strong guarantees for the convergence of the training.

\paragraph{Problem Independence in the Runtime of FedNL}
The theoretical framework of the \algname{FedNL} family \cite{safaryan2021fednl} has the advantageous property that its runtime does not explicitly depend on the characteristics of the local loss function $f_i(x)$. Thus, the algorithm can be applied to any problem in class, provided that the underlying numerical algorithms are stable.

\paragraph{Zero Heuristics}
The algorithm's \algname{FedNL} runtime does not require any prior estimations of parameters exactly or heuristically which provides essentially zero theoretical and practical gap. If this gap is not zero, in practice, heuristics are used to eliminate it. In this regard, several considerations must be noted: (i) their development of heuristics (or approximation) requires significant researcher time; (ii) they may unintentionally be problem-instance specific; (iii) as the number of heuristics increases, navigating their space can become exponentially complex; (iv) transferring heuristics from the design phase (with a human in the loop) to actual runtime (without human involvement) poses challenges.

\paragraph{Leveraging FedNL Theory for Autonomous Implementation}
Improving core applications on edge devices requires acknowledging that mobile applications and operating systems typically cannot afford human intervention during runtime. We believe our implementation can potentially help with a rich class of applications in edge devices. In classical centralized ML, training algorithm requirements can be relaxed as long as convergence is achieved; the resulting trained model (obtained offline or during design time) can be applied in client applications with robust inference algorithms (in runtime). A key distinction in FL is that it involves not only inference on edge devices but also training itself.

\paragraph{Constructive Theoretical Requirements for Compressors}
Our discovery of two new compressors supports the point that the compressors definition from the original \algname{FedNL} paper \cite{safaryan2021fednl} facilitates the development of novel compressors.

}

\clr{
\subsection{Generality and Extensibility of the Proposed FL Implementation}
\label{app:extensibility-of-impl}

\paragraph{Selection of logistic regression as a benchmark.} We have chosen logistic regression as a benchmark for three main reasons. First, we used the same datasets and problem formulation as in the original reference FedNL implementation, because our improvements were built upon the original work. Second, solving logistic regression in a manner that is (i) applicable in edge environments for FL; (ii) requires no fine-tuning of the training process; (iii) effectively executed on edge devices is virtually absent as we already described in Section~\ref{sec:existing-fl-sota-system}.

\paragraph{Generality of the Theory and Our Improvements to Other Problem Classes.}

The theoretical framework of \algname{FedNL} is general and extends beyond solving the specific class of logistic regression problems. The algorithm is provably capable of solving any problem in the form of Eq.~\ref{eq:main} under Assumptions \ref{asm:1} and \ref{asm:2}. During improving FedNL, only steps 17, 21, 50, and 58 from Appendix~\ref{app:history-of-improvements} were directly aimed at enhancing the Hessian Oracle for this model. Other steps contributed to performance enhancements through orthogonal improvements, independent of the specific problem class.

\paragraph{Extensibility of our Implementation for future research.}

In addition to guiding how to use our solution (Appendix~\ref{app:usability}), we also detail how to incorporate new objectives and make system modifications for future research. To achieve this, we provide a robust C++ framework for implementing custom oracles, including essential tools such as linear algebra operations and linear solvers. Researchers can leverage our compute primitives, which are optimized for a wide range of CPU architectures (with and without special SIMD instructions) and NVIDIA GPU architectures via expressing computation directly in  CUDA. In addition, we provide methods for numerically verifying the correctness of computed quantities such as $f_i(x)$, $\nabla f_i(x)$, and $\nabla^2 f_i(x)$. For details see Appendix~\ref{app:futresearch}.

}

	\clearpage
	\section{Technical Discussions}
	\label{app:technical}

	\subsection{Why We Selected TCP/IP for FedNL Implementation}
	\label{app:networks-why-tcp}
	
	In our \abr{FedNL} implementation, clients establish connections with the master using either \abr{TCP/IPv4} or \abr{TCP/IPv6} and use this level of protocols. This level of abstraction allows us to distance ourselves from the physical principles governing information transfer, providing a high-level interface without sacrificing finer granularity optimizations. Although the UDP protocol induces less overhead, it lacks essential features such as flow control, congestion control, delivery guarantees, and ordering guarantees.
	
	The HTTP  application protocol is employed as transport layers in various communication and remote procedure call protocols such as gRPC, XML-RPC, or REST. Classical HTTP/1 introduces overheads, such as the \textit{protocol request text, request header, blank lines}, preceding the actual transferred message. If the communication session is relatively long, special treatment is required for maintaining a persistent connection. We decided \textit{not to rely on HTTP} and upper-level application protocols. Even \abr{gRPC} employs a persistent \abr{TCP} connection, we deemed \abr{gRPC} unnecessary for our purposes. In our view, \abr{gRPC} should be employed for its primary purpose only. Next, gRPC/REST/XML-RPC relies on \abr{HTTP}, and \abr{HTTP} relies on \abr{TCP/IP}. We do not see a reason why we should work at such a high level. Any unnecessary abstractions that can not be eliminated during design time take resources and are not free.
	
	
	\clr{
	\subsection{More about Memory-Mapped Files}
	\label{app:mmap}

	Details about this mechanism can be found in specialized OS literature, such as \cite{kerrisk2010linux}, though it is seldom widely discussed. Counterintuitively, the most efficient way to read data from a file system is not through standard read operations provided by the OS. While files on disk adhere to the file system structure, conventional file-reading methods often involve an intermediate buffer managed by the OS I/O dispatcher and drivers. A memory-mapped file is an OS primitive available in modern systems such as Windows, Linux, and macOS, allowing files to be mapped directly into a process's virtual address space. 
	
	\paragraph{This mechanism offers several key advantages:}
	
	\begin{enumerate}
		\item \textit{Direct Access:} Read and write operations bypass system calls and the I/O dispatcher, with the CPU and OS handling page faults through built-in interrupt mechanisms.
		
		\item \textit{Efficient Memory Usage:} Modern OS can use swap files to emulate larger physical memory. When utilizing memory-mapped files, even with insufficient OS memory, the file itself serves as backup storage, preventing additional allocation in the swap.

		\item \textit{Flexible Memory Management:} If physical pages in DRAM must be evicted, the OS can release physical pages directly without writing data back to disk. In contrast, standard buffers require data to be written back, consuming additional resources.
		
		\item \textit{Importance of Implementation in OS:} A well-implemented memory mapping mechanism by authors of OS is crucial. It is used for loading native applications and libraries and forms the basis for inter-process communication methods. Therefore, it's very likely that this mechanism is extremely well-designed in the target OS.
		
	\end{enumerate}
	
	}

\clr{
	\clearpage
	\subsection{Performance Limitations of the Python Ecosystem}
	\label{app:nopython}

	\paragraph{Inheriting Downsides of Interpretation Agnostic to Implementation}

	Interpreters execute scripts directly without preprocessing, while compiled languages involve multiple stages before generating executable binaries, including code generation and code optimizations. According to \cite{wexelblat2014history}, compiler research began around 1950s with John Backus's Speedcoding project on the IBM 704 to address rising software costs in society. This project led to the creation of FORTRAN, a foundation for many later languages, including C++ \cite{stroustrup1994design}. Though compiled languages like C++ are often seen as low-level nowadays, raising the abstraction level can risk losing control over algorithm performance and its interaction with hardware and OS, potentially degrading performance despite user expertise\footnote{This assumes the algorithm is fixed. Adapting algorithms to architectural features is a viable research direction as discussed in Section~\ref{sec:extra-motivation}.}. As discussed in Appendix~\ref{app:history-of-improvements}, programming language choice yielded a $\times 20$ speedup, but $\times 50$ additional speedup comes from the ability to implement needed computational and mathematical optimizations directly on the hardware.

	\paragraph{The Suboptimality of Design with Using Dynamic Libraries}

	The Python interpreter leverages in addition to its core to the Python modules and extension modules to expand its capabilities. Extension modules for Python are dynamic (or shared) libraries provided by the operating system. While they may appear to be a viable option, but for time-critical applications, heavy reliance on shared libraries can be a poor design choice per se. For performance-critical applications, static linking of isolated modules is often a far better alternative. This is because modern compilation tools can access the entire codebase during final program construction, enabling more compiler optimization techniques. Another performance drawback of dynamic libraries is the issue that arises when two or more libraries cannot be loaded at the same virtual address inside the executable process. To address this, approaches such as Code Relocation are utilized in the Windows OS, and Position Independent Code (PIC) is used in macOS and POSIX operating systems. Both of them lead to performance problems. Finally, the actual application start time degrades as the number of loaded dynamic libraries increases. The overall situation can become quite problematic if these libraries (located in SSH/HDD for which access is slow based on Table~\ref{tbl:latencies-for-memory}), in a recursive manner, import additional libraries.
}

\clr
{
	\paragraph{Unimposing of System Thinking}
	Python's modular approach promotes the reusability of solutions constructed from distinct blocks. Sometimes, it is what is needed to focus in case of complex situations. However, this block-based thinking can contradict a system-level perspective, which emphasizes a holistic view of solutions rather than concentrating on individual components. Compiled languages offer a robust set of instrumentation tools that support a holistic and system-level mentality to isolate the problems. Such style teaches that there is a need to profile and look into the entire computation stack.
}

\clr{
\paragraph{Lack of Compiler Level Optimization for Users.} Compilers is a well-established field of Computer Science. In case use compilers it provides ways to utilize techniques such as code inlining, global optimization, converting arithmetic operations to more efficient bit shifts, optimizing memory layout, eliminating dead code, removing loop-invariant computations, loop fusion, utilizing fast local stack storage, and profile-guided optimization. These challenging optimizations still require collaboration between the algorithm designer and the compiler. Implementing them in interpreted environments is even more challenging.
}

\clr{

\paragraph{Challenges in Developing Multi-Threaded Implementations.} 

Creating reliable multi-threaded implementation involves several challenges, including managing memory barriers (or memory fences) to enforce the correct order of CPU instructions, ensuring proper synchronization, preventing data races, and handling atomic operations. Implementing effective multi-threading in interpreted languages is particularly complex. For example, while systems like Ray \cite{moritz2018ray} provide distributed computing for Python, they rely on multi-processing, rather than multi-threading, for scalability, even within a single node.

}

\clr{	
\paragraph{Challenges with Inefficient Memory Access.} As user-defined applications in the interpreter scripting language grow, the Python script execution logic may become deeply intertwined with the interpreter's operations from a CPU perspective. This intertwining can result in the interpreter consuming valuable CPU resources. For example, even when the objective is to run a training algorithm, the CPU caches may become polluted with the interpreter's logic, leaving less room for the user’s algorithm to benefit from caching.
}	

\clr{
	\paragraph{Inability to Leverage Special CPU Registers and CPU Instructions Directly}
	
	At a fundamental level, all executable binaries, regardless of their file format, are ultimately executed using the CPU's native instruction set, which encodes operations into digital machine language. Python cannot directly interact with this hardware level, which limits performance in computation-heavy scenarios. Modern CPUs are equipped with special instructions and registers in their instruction set architectures (ISA) designed to optimize computation. Examples of such special instructions for 64-bit processors include SSE2 (16 XMM registers, each $4 \times 32$-bit), AVX2 (16 YMM registers, each $8 \times 32$-bit), AVX-512 (32 ZMM registers, each $16 \times 32$-bit), and ARM AArch64 Neon (32 V registers, each $4 \times 32$-bit). These special instructions allow highly efficient elementwise operations on small dense vectors. Additionally, special-purpose instructions such as Fused Multiply-Add (FMA) enhance performance by combining multiple operations into one, dramatically improving performance in compute-intensive tasks. Interpreted languages like Python cannot directly leverage these capabilities due to their reliance on abstraction layers.
}

\clr
{	
	\paragraph{Ineffective Use of CPU Registers}
	As shown in Table \ref{tbl:latencies-for-memory}, the electrical circuits representing registers are the fastest data storage available within the CPU. Communication with the CPU from any software occurs through instructions that adhere to a specific Instruction Set Architecture (ISA). The ISA defines memory architecture, data types, input/output interfaces, and registers. The number of registers in ISA is limited and typically ranges from $32$ to $64$. This discrepancy between high-end processor vendors, who can allocate more registers in silicon and practical applications has been resolved as follows. The $32-64$ registers are the Front-End registers specified by the ISA, while at the microarchitecture level, the actual registers are stored in a common area known as the Register File, which can contain significantly more but is often limited by $1000$ registers\footnote{Microarchitecture details are typically subject to non-disclosure agreements.}. The Compiler community does not lack awareness of this discrepancy, but they do not have enough tools to operate. In reality, once the available registers are exhausted, the Compiler's code emission algorithms must resort to \textit{register spilling}, backing up registers in memory. A more effective approach to leverage CPU pipelines (and implicitly the Register File) is to express Instruction Level Parallelism (ILP) directly within the code, making the code simple and predictable enough for modern CPU's Instruction Decode and Operation Issue blocks. Interpreted languages introduce numerous wrappers that ruin the ability to use ILP and Register File via ILP.

    \paragraph{The Hidden Challenges of Relying on Vectorization in Scripting Languages.} In Python, vectorization often refers to processing large data blocks in parallel, typically achieved through external libraries. While this approach can significantly enhance performance, it poses challenges in certain scenarios: (i) when you need to implement such functionality yourself, relying on Python may be a strategically poor choice; (ii) when specific problems cannot be effectively represented or solved using existing library interfaces; and (iii) when the overhead of dispatching operations to external libraries exceeds the potential performance gains.
}

	\subsection{Carried Steps to Ensure Usability of Implementation}
	\label{app:usability}
	\clr{\paragraph{Overview.}} Multiple optimization levels increased efficiency and performance. However, we strived to ensure that the presented work and future work building on top of the presenting implementation are practically feasible. To attain this we provide different means:

	\begin{enumerate}
		\item Compiled automatic code documentation in Doxygen.
		\item Means to build projects and launch unit tests in macOS, Linux, and Windows
		\item Utils for observing the system configuration
		\item Generating synthetic optimization problems for logistic regression
		\item Code Ecosystem for systems, networking, data parsing, and mathematical routines
		\item Primitive to work with (various forms of) dense matrices and vectors
		\item Wrappers to work with CPU SIMD to eliminate the need for low-level compiler intrinsics
		\item Means for sanity checks for gradient and Hessian oracles with finite differences approach
		\item Classical iterative and dense-direct solvers for solving linear systems
		\item Means for launch automatic builds, launching unit tests, launch  code coverage
	\end{enumerate}

\clr{
	\paragraph{Step-by-Step Guidelines.}
	We provide step-by-step guidelines accessible through standalone Markdown files available in the code repository. These guidelines are also included in the compiled automatic code documentation, offering coherent instructions for using our work:
	
	\begin{enumerate} 
		\item \textit{Minimal Environment:} Setting up the core build and runtime environment across Windows, Linux, and macOS, including configuration for alternative runtimes. 
		\item \textit{Extra Tools:} Guidelines for setting up additional tools. 
		\item \textit{Local Build:} Instructions on building the project locally (macOS, Linux, Windows). 
		\item \textit{Docker Build:} Steps to build the project in a Docker container, ideal for testing on different CPU architectures. 
		\item \textit{Experiments:} Instructions for running the experiments described in the  paper. 
		\item \textit{GitHub Integration:} Utilizing GitHub Actions for continuous integration. 
		\item \textit{Project Structure:} An overview of the project’s structure and its key components. 
		\item \textit{Advanced Configurations:}  Configuring the build beyond the default settings.
	\end{enumerate}
}

\clr{

\paragraph{Addressing Slow Compilation Time.}

One of the primary reasons for the limited popularity of C++ in ML is the long compile time. To mitigate this issue, we provide the following:

\begin{enumerate} 
	\item Exclude specific components from the build to reduce compile time: 
	\begin{enumerate} 
		\item Utility programs such as dataset generators, system viewers, and GPU viewers. 
		\item Unit tests, which cover nearly all functionality. 
		\item CUDA support, NVIDIA's C/C++ extension for writing logic on GPUs. 
		\item OpenCL support, an API for GPUs across various vendors. 
		\item Generate bindings for other languages (e.g., Python) via Simplified Wrapper and Interface Generator(SWIG) \cite{beazley1995simplified}. This process can be time-consuming. 
		\item Build shared libraries for clients and servers, including executable applications.
	\end{enumerate} 
	
	\item The project supports invoked build processes, running unit tests, and launching binaries from the command line via created \texttt{project\_script.py}. This allows for offloading building and testing to another machine. 
	
	\item For local builds, \texttt{project\_script.py} facilitates the use of the Ninja build system (faster than GNU Make on Linux/macOS) and allows configuration of the number of CPU cores used during compilation. 
	
	\item Support for Unity (Jumbo) Build, a compile optimization technique that combines multiple source files into a single large file. 
	
	\item Precompiled headers to speed up compilation times. 
	\item Custom C and C++ compilers can be specified using the CXX and CC environment variables, which are supported by common build systems like CMake, GNU Make, and Microsoft Visual Studio. 
	
	\item CPU Instruction Set Architecture-specific optimizations are automatically detected but can also be configured manually. We offer user-friendly options in a tool-independent manner to adjust build settings, primarily affecting performance.
	 
	\item Extra instrumentation in the build process can be easily enabled or disabled.

\end{enumerate}
}	
\clr{
	
\paragraph{Handling Complexity in Large Projects.}

As a project grows in size, making non-trivial modifications becomes more challenging. To address this, we provide the following:

\begin{enumerate} 
	
	\item The code is organized as a collection of static libraries for easier management. 
	
	\item We ensure near-complete test coverage, verified with the GNU Coverage tool. 
	
	\item Comprehensive Doxygen documentation provides detailed code documentation. 
	
	\item If your IDE has limited navigation features, we include diagrams in the compiled HTML documentation to visualize the project structure. 
	
	\item To facilitate learning, the documentation and the code are integrated into the compiled documentation. 
	\item The project can be built using command-line utilities and exported to build and debugged into popular Integrated Development Environments such as Microsoft Visual Studio, QtCreator, and CLion. 
\end{enumerate}
}	

\clr{
\paragraph{Tools for Code Quality.}

To ensure code quality, we employed the following tools:

\begin{enumerate}

	\item The compilation process ensures that there are no compile-time or linkage errors. Unlike scripting languages, where errors can be encountered at runtime, C++ requires all compile errors to be resolved before producing a valid program. In contrast, Python parses functions only at the moment of direct invocation.
	
	\item We address compiler warnings across multiple C++ implementations. 
	
	\item Unit tests to enhance confidence in the correctness.
	
	\item Valgrind (memcheck) to detect memory leaks
	
	\item Valgrind (callgrind), Intel VTune, Linux Perf Tool to detect memory cache problems
		
	\item We used CppCheck and PVS-Studio for static code analysis to enhance quality.
	
	\item We mainly used the functionality of nvprof - command line profiler from NVIDIA CUDA Toolkit to improve CUDA support implementation.
	
	\item Algorithmic convergence is verified against theoretical expectations. 
\end{enumerate}
}

\clr{
	\paragraph{Integration into Applications.}
	
	Our Federated Learning (FL) training runtime can be integrated with existing applications in several ways:
	
	\begin{enumerate} 
		
		\item \textit{Standalone Executable:} The local client and server are native applications for your target OS, with no dependencies on third-party libraries at runtime.
		
		\item \textit{Dynamic Libraries:} The project can be built as distributed dynamic libraries (\texttt{.dll} for Windows, \texttt{.so} for Linux, and \texttt{.dylib} for macOS), with separate libraries for client and server, each having a single entry point. 
		
		\item \textit{Language Bindings via SWIG:} If your project uses a language supported by the Simplified Wrapper and Interface Generator (SWIG), such as Python, Perl, Java, or C$\#$, you can automatically generate interfaces to invoke the training process. We provide an example of code generation for Python. 
		
	\end{enumerate}
}

\clr{
	\clearpage
	\subsection{Technical Overview of Project Architecture}	
	\label{app:futresearch}
}
\begin{table}[h]
	\centering
	\begin{tabular}{|p{0.28\textwidth}|p{0.54\textwidth}|p{0.12\textwidth}|} 
		\hline
		\rowcolor{gray!30}
		\textcolor{black}{\textbf{Component Name}} & \textcolor{black}{\textbf{Goal}} & \textcolor{black}{\textbf{Type}} \\ \hline
		CMakeLists.txt & Main root CMake project configuration file. & Config \\ \hline
		project\_scripts.py & Script to help launch builds, tests, docs, and etc. & Script \\ \hline
		copylocal & Low-level utilities to work with bytes and bits and copy them effectively. & Static Library \\ \hline
		cmdline & C++ cross-platform implementation of useful command line parsing mechanisms. & Static Library \\ \hline
		fs & Wrappers to work with string conversion routines, operate with files and filenames, and memory map in a system-independent way. Well optimized. & Static Library \\ \hline
		linalg\_linsolvers & Collection of dense and iterative specialized linear solvers in CPU. & Static Library \\ \hline
		math\_routines & Special math routines - matrix sparsification, convex optimization. Also includes data structures and algorithms: sorting, tries, heaps, and indexed heaps. & Static Library \\ \hline
		numerics & Number differentiation to evaluate derivatives, gradients, and Hessians numerically. & Static Library \\ \hline
		timers & Various timers - systems, from C++ runtime (CPUs clocks to seconds), low-level wrappers to calculate clocks for x86 and AArch64. & Static Library \\ \hline
		digest & Digest to check bit-bit equivalence CRC-32-IEEE 802.3, MD5 RFC1321 Algorithms (system part). & Static Library \\ \hline
        network & TCP/IPv4, TCP/IPv5, UDP/IPv4, UPP/IPv6 wrappers for Windows and POSIX API (system part). & Static Library \\ \hline
		system & Memory pools, OS Memory Allocators, Low-Level operations on Float/Double scalars. & Static Library \\ \hline
		random & Calculate central statistics (mean, variance), uniform pseudo-random generators, R.V. generators, and shuffling with early stopping. & Static Library \\ \hline
		linalg\_vectors & Dense vectors and light vector views with carefully designed API, including vectors with different underlying storages and custom implementation (with SIMD). & Static Library \\ \hline
		linalg\_matrices & Dense matrix implementation for BLAS operations, Cholesky, and QR factorization. & Static Library \\ \hline
		linalg\_linsolvers & Several linear systems solvers: Jacobi, Gauss-Seidel, Conjugate-Gradient, Gauss-Elimination, backward and forward substitution (with SIMD). & Static Library \\ \hline
		gpu\_compute\_support & The project partially supports CUDA GPU computation. It contains dense vector, dense matrix operations, GPU memory management, and wrappers over low-level GPU code invocation. & Static Library \\ \hline
		numerics & Tools for numerical differentiation to validate the correctness of Hessian and gradient computations. & Static Library \\ \hline
		optimization\_problems & Implements optimization problems, including logistic regression and quadratic minimization. & Static Library \\ \hline

	\end{tabular}
	\caption{\clrshort{Components of the Self-Contained Runtime}}
	\label{tab:ufednl-components}
\end{table}

		\clr{ 
			\paragraph{The Core Components.}
			The core low-level components are detailed in Table~\ref{tab:ufednl-components}. We offer out-of-the-box implementations for logistic regression and Symmetric Quadratic Objectives. From Table~\ref{tab:ufednl-components} it can be observed that we provide reach expressability of various optimization problems on modern CPUs beyond logistic regression. In terms of computation, although we support GPU computation, our implementation is pretty restricted to NVIDIA GPUs, for which we have developed functionality for dense vector and matrix operations. 
			
			Currently, we do not support OpenCL or Apple Metal API. Additionally, our compressors operate on the CPU, with master aggregation also occurring on the CPU. Exploring robust implementations for modern GPUs, which follow a different computational model, remains a valid, practical research direction.
		}
	
	\clr{
		\paragraph{Auxiliary Binary Tools.}
		
		To facilitate the ongoing efforts of this project, we believe that the specifically designed binary tools listed in Table~\ref{tab:auxiliary-binaries} will  assist future researchers in advancing this work.

		\paragraph{Main Executable Applications.}
		The obtained binary application for single node simulation are listed in Tables~\ref{tab:single-node-runners}. The obtained binary application for multiple-node simulation are listed in Tables~\ref{tab:multiple-node-training}. To obtain extra executable binaries you should specify:
	\begin{enumerate}
		\item \texttt{DOPT\_SWIG\_INTERFACE\_GENERATOR} - Enables the use of SWIG to generate Python API wrappers. The generated libraries will have the prefix \texttt{"python\_"}.
		\item \texttt{DOPT\_BUILD\_SHARED\_LIBRARIES} - Instructs the build system to create shared libraries, which will be prefixed with \texttt{"shared\_"}.
	\end{enumerate}		
			
	}
		\begin{table}[h]
			\centering
			\begin{tabular}{|p{0.28\textwidth}|p{0.54\textwidth}|p{0.12\textwidth}|} 
				\hline
				\rowcolor{bgcolorwe}
				
				\textbf{Utility Name}            & \textbf{Goal}                                                                                                                                                                             & \textbf{Type}   \\ \hline
				bin\_tests                       & Google Unit tests. Total number of tests is 102, but each test is pretty big. Tests cover all projects.                       & Executable \\ \hline
				utils/bin\_host\_view             & Binary application to check used compiler name, flags, version, linker flags, information about OS, DRAM, and CPU extensions.                                                        & Executable \\ \hline
				bin\_opencl\_view           & Observe platforms for OpenCL, devices available that support computation via OpenCL in the platform, its type, and available extensions if you will provide \texttt{show-extensions}.        & Executable \\ \hline
				bin\_cuda\_view             & Observe NVIDIA CUDA compatible devices, show peak computation and memory throughput limits, available DRAM, and features. Run simple tests to verify. Flags: \texttt{verbose}, \texttt{benchmark}. & Executable \\ \hline
				bin\_opt\_problem\_generator & Optional synthetics optimization problem generator. Can be used for debugging purposes.                                                                                              & Executable \\ \hline
				bin\_split                 & Binary program to take one dataset in text, optionally reshuffle, add intercept, obtain information about several clients, and split it into several files.                          & Executable \\ \hline
			\end{tabular}
			\caption{\clrshort{Auxiliary Help Binaries}}
			\label{tab:auxiliary-binaries}
		\end{table}

	\begin{table}[h]
		\centering
		\begin{tabular}{|p{0.42\textwidth}|p{0.4\textwidth}|p{0.12\textwidth}|} 
			\hline
			\rowcolor{gray!30}
			\textbf{Project} & \textbf{Goal} & \textbf{Type} \\ 
			\hline
			[bin|shared|python]\_fednl\_local & Local simulation on the machine for FedNL and FedNL-LS with mentioned compressors. & Executable \\ 
			\hline
			[bin|shared|python]\_fednl\_local\_pp & Local simulation on the machine for FedNL and FedNL-LS with compressors for Partial Participation. & Executable \\ 
			\hline
		\end{tabular}
		\caption{\clrshort{Single-Node Multi-Core Runners}}
		\label{tab:single-node-runners}
	\end{table}

	\begin{table}[h]
		\centering
		\begin{tabular}{|p{0.42\textwidth}|p{0.40\textwidth}|p{0.12\textwidth}|} 
			\hline
			\rowcolor{gray!30}
			\textbf{Project} & \textbf{Goal} & \textbf{Type} \\ 
			\hline
			[bin|shared|python]\_fednl\_distr\_client & Client application to participate in FedNL and FedNL-LS algorithms with various compressors. & Executable \\ 
			\hline
			[bin|shared|python]\_fednl\_distr\_client\_pp & Client application for partial participation in FedNL and FedNL-LS algorithms. & Executable \\ 
			\hline
			[bin|shared|python]\_fednl\_distr\_master & Server application to control FedNL and FedNL-LS training. & Executable \\ 
			\hline
			[bin|shared|python]\_fednl\_distr\_client\_pp & Server application for partial participation in FedNL and FedNL-LS training. & Executable \\ 
			\hline
		\end{tabular}
		\caption{\clrshort{Multiple-Node FL Training Runners}}
		\label{tab:multiple-node-training}
	\end{table}

	\clearpage
	\section{Acknowledgements}

	The research reported in this publication was supported by funding from King Abdullah University of Science and Technology (KAUST): i) KAUST Baseline Research Scheme, ii) Center of Excellence for Generative AI, under award number 5940, iii) SDAIA-KAUST Center of Excellence in Artificial Intelligence and Data Science.

	\clearpage
	\section{Limitations}
	\label{app:limitations}
	
	\paragraph{Lack of Mobile OS Support} Our tested operating systems encompass macOS, Linux, and Windows, but exclude mobile platforms, even though our implementation supports ARM CPU with (and without) ARM Neon extensions
	\clrshort{, and in general our work was tested in a rich set of configurations (see Appendix ~\ref{app:supported-os-and-compilers}, \ref{app:cpus}).}
	
	\paragraph{Consumed Memory for Dataset}
	
	The main issue that motivated our work was the extremely long duration of executing \algname{FedNL}. The memory aspect of dataset storage has not been considered in our work. Essentially the dataset during training is read from preprocessed in-memory storage, rather than from disk during training. 
	
	\paragraph{Not investigate Mathematical Aspects} Interesting mathematical aspects are still unexplored for \algname{FedNL} including a more robust globalization strategy, addressing storage concerns for Hessian shifts $H_i^k$ in clients and $H^k$ in master, or integrating iterative inexact linear solvers.
	
	\clr{
		\paragraph{Limited GPU Support}
		
		GPU acceleration can significantly enhance performance, provided the overhead of executing code on the GPU and transferring input/output data is minimal or can be hidden. Our current support for GPU utilization is limited, focusing primarily on help for forming oracles using NVIDIA CUDA or exploiting linear algebra primitives. Expanding GPU support in the future presents both interesting opportunities and challenges:
		
		\begin{enumerate}
			\item There is no unified language across major GPU vendors, which complicates the development of subtle, vendor-agnostic implementations.
			\item Making the \algname{FedNL} implementation more GPU-friendly is a complex task due to the intricate rules governing computation and memory transfers that are exposed to algorithm designers.
			\item GPU support is currently absent not only in the reference \algname{FedNL} implementation but also in widely used libraries such as \texttt{SkLearn}, \texttt{MOSEK}, \texttt{Spark}, and \texttt{CVXPY}.
		\end{enumerate}
	}		

	\clearpage
	\section{Conclusions}
	

	Our work represents a significant contribution to advancing the field of FL by addressing a crucial gap between theoretical advancements in \algname{FedNL} optimization algorithm family and their practical implementation. Drawing inspiration from \cite{safaryan2021fednl}, our work is rooted in cutting-edge optimization theory, demonstrating the way of implementing theory into resource-constrained FL settings. Our work serves as a guiding beacon for researchers navigating the intricate path of translating theoretical algorithms into impactful implementations across diverse domains of Machine Learning. Our work emphasizes the multifaceted considerations involved in aiming to improve the actual wall clock time.

	Also, our work challenges the predominant Python-centric design philosophy in Machine Learning. It underscores the significance of considering alternative languages when prioritizing computational and memory efficiency.
	
\end{document}